\documentclass[iicol]{sn-jnl}% Math and Physical Sciences Numbered Reference Style 
\usepackage{graphicx}%
\usepackage{multirow}%
\usepackage{amsmath,amssymb,amsfonts}%
\usepackage{amsthm}%
\usepackage{mathrsfs}%
\usepackage[title]{appendix}%
\usepackage{xcolor}%
\usepackage{textcomp}%
\usepackage{manyfoot}%
\usepackage{booktabs}%
\usepackage{algorithm}%
\usepackage{algorithmicx}%
\usepackage{overpic}%
\usepackage{algpseudocode}%
\usepackage{listings}%
\usepackage{cite}%
\usepackage{bm}%
\usepackage{setspace}
\usepackage{subcaption}
\usepackage{adjustbox}
\usepackage{enumerate}
\usepackage{marvosym}
\theoremstyle{thmstyleone}%
\theoremstyle{thmstyletwo}%

\theoremstyle{thmstylethree}%

\begin{document}

% \title[Article Title]{Reblurring-guided image defocus deblurring with Misaligned Training Pairs }
\title{Reblurring-Guided Single Image Defocus Deblurring: A Learning Framework with Misaligned Training Pairs}

\author[1]{\fnm{Dongwei} \sur{Ren}\textsuperscript{\Letter}}
\equalcont{These authors contributed equally to this work.}

\author[2]{\fnm{Xinya} 
\sur{Shu}}%\email{iauthor@gmail.com}
\equalcont{These authors contributed equally to this work.}

\author[2]{\fnm{Yu} \sur{Li}}%\email{liyuhit@hotmail.com}

\author[2]{\fnm{Xiaohe} \sur{Wu}}%\email{rendongweihit@gmail.com}

\author[3]{\fnm{Jin} \sur{Li}}%\email{lijin@tju.edu

\author[2]{\fnm{Wangmeng} \sur{Zuo}}%\email{cswmzuo@gmail.com}

\affil[1]{\orgdiv{College of Intelligence and Computing, Tianjin University}}

\affil[2]{\orgdiv{Faculty of Computing, Harbin Institute of Technology}}

\affil[3]{\orgdiv{School of Electrical and Information Engineering, Tianjin University}}

\abstract{
	For single image defocus deblurring, acquiring well-aligned training pairs (or training triplets), \emph{i.e.}, a defocus blurry image, an all-in-focus sharp image (and a defocus blur map), is a challenging task for developing effective deblurring models. 
	Existing image defocus deblurring methods typically rely on training data collected by specialized imaging equipment, with the assumption that these pairs or triplets are perfectly aligned. 
	However, in practical scenarios involving the collection of real-world data, direct acquisition of training triplets is infeasible, and training pairs inevitably encounter spatial misalignment issues.
	In this work, we introduce a reblurring-guided learning framework for single image defocus deblurring, enabling the learning of a deblurring network even with misaligned training pairs.
	By reconstructing spatially variant isotropic blur kernels, our reblurring module ensures spatial consistency between the deblurred image, the reblurred image and the input blurry image, thereby addressing the misalignment issue while effectively extracting sharp textures from the all-in-focus sharp image. 
	Moreover, spatially variant blur can be derived from the reblurring module, and serve as pseudo supervision for defocus blur map during training, interestingly transforming training pairs into training triplets.
	To leverage this pseudo supervision, we propose a lightweight defocus blur estimator coupled with a fusion block, which enhances deblurring performance through seamless integration with state-of-the-art deblurring networks. 
	Additionally, we have collected a new dataset for single image defocus deblurring (SDD) with typical misalignments, which not only validates our proposed method but also serves as a benchmark for future research.
	The effectiveness of our method is validated by notable improvements in both quantitative metrics and visual quality across several datasets with real-world defocus blurry images, including DPDD, RealDOF, DED, and our SDD.
	The source code and dataset are available at~\url{https://github.com/ssscrystal/Reblurring-guided-JDRL}.}

\maketitle

\begin{figure*}
	\centering
	\begin{minipage}[b]{0.95\linewidth}
		\centering
		\centerline{
			\includegraphics[width =\linewidth]{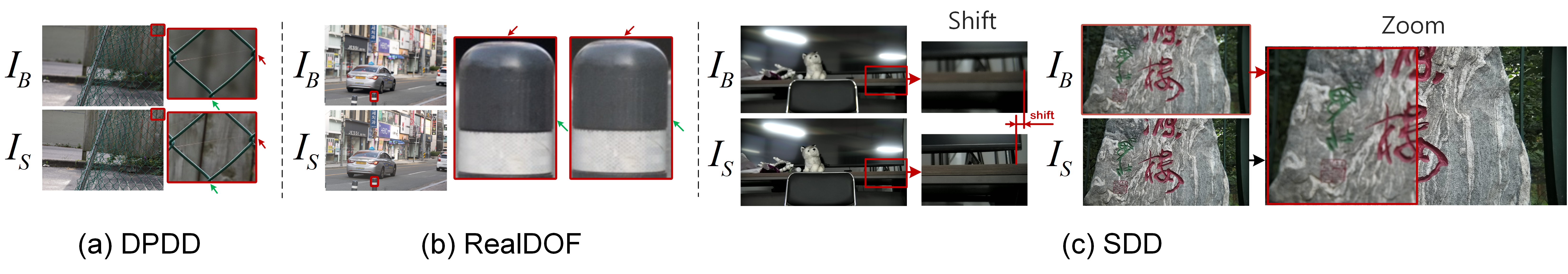}}
	\end{minipage}
	\caption{\small Misalignment issues in the pairs of ground-truth sharp image $\bm{I}_S$ and defocus blurry image $\bm{I}_B$ in DPDD \cite{abuolaim2020defocus}, RealDOF \cite{lee2021iterative} and our SDD datasets. Although ground-truth and blurry image pairs in these datasets are designed to be aligned, spatial misalignment still exists.}
	\label{intro}
\end{figure*}

\section{Introduction}
\label{sec1}
Defocus blur typically occurs when scene objects fall outside the camera's Depth of Field (DOF). This phenomenon can create a visually pleasing effect in certain photographic contexts. However, defocus blur often compromises the clarity of image details, thereby negatively impacting image quality and hindering research on high-level tasks such as object detection \cite{zhao2019object, zou2023object, papageorgiou1998general} and segmentation \cite{li2014secrets, carreira2010constrained, hariharan2015hypercolumns}.
To address these challenges, image defocus deblurring is required to handle various and complex blurred areas produced during the photographing process.
Traditionally, the image defocus deblurring process adopts a two-step approach. 
It initially computes a defocus blur map \cite{goldstein2012blur,Park_2017_CVPR,Shi_2015_CVPR}, which delineates the amount of blur per pixel within a defocused blurry image. 
This map is then used to perform non-blind deconvolution~\cite{krishnan2009fast,fish1995blind} on the image.
The effectiveness of this strategy heavily depends on the precision of the defocus blur map. 
However, this approach often overlooks the nonlinear dynamics of real-world blurring, and tends to rely on simplistic blur models such as disk or Gaussian kernels.
Recently, dual-pixel cameras have been employed to address defocus blur \cite{abuolaim2020defocus, abuolaim2021learning} through the utilization of two-view images. However, it is worth noting that the majority of consumer cameras still produce single images for user observation. Therefore, this paper primarily focuses on the domain of single image defocus deblurring.

The emergence of deep learning techniques, particularly convolutional neural networks based models \cite{abuolaim2021learning,abuolaim2020defocus} and Transformer based models \cite{zamir2022restormer, wang2022uformer,zhang2024unified,li2023efficient}, has significantly propelled the field forward by providing solutions to address defocus blur.
These models demonstrate effectiveness through the acquisition of intricate mappings from extensive training data. 
However, the effectiveness of these learning-based approaches is intricately related to the quality and alignment of training samples. 
Often, these deblurring methods depended on precisely aligned pairs of images. In DED dataset \cite{ma2021defocus}, well-aligned defocus blurry images and their corresponding all-in-focus counterparts can be obtained using a light field camera like Lytro \cite{ng2005light}. From such pairs, a defocus blur map can be estimated, resulting in training datasets comprising aligned triplets: a defocus blurry image, an all-in-focus ground-truth image, and a defocus blur map. The spatial alignment plays a pivotal role in effectively training and validating deep learning models for image defocus deblurring task.

However, for other consumer cameras, \emph{e.g}., digital single lens reflex or smartphone cameras, the availability of training triplets is impractical. 
In widely used datasets for image defocus deblurring, such as DPDD \cite{abuolaim2020defocus} and RealDOF \cite{lee2021iterative} datasets, meticulous control is exercised over the capturing camera to ensure consistent acquisition of training pairs comprising a defocus blurry image and a ground-truth sharp image. 
Nevertheless, as illustrated in Fig.~\ref{intro}, inevitable misalignments still occur. 
We also note that different imaging sensors introduce various types of defocus blur, posing challenges for learned deblurring models based on specific cameras when handling cases involving other sensor types.
Therefore, effective deployment on target devices requires learning a device-specific image defocus deblurring model while relaxing the requirement for perfect alignment in training pairs.

In this work, we propose a novel learning framework for single image defocus deblurring (Fig. \ref{fig pipeline}), specifically focusing on effectively learning a single image defocus deblurring model from misaligned training pairs that can be easily obtained for a given imaging camera. 
% In particular, we introduce a reblurring-based learning framework, as shown in Fig. \ref{fig:reblurnet}, to address the challenge of misaligned training pairs by incorporating a new reblurring module. 
%We first introduce a baseline image defocus deblurring model that utilizes a lightweight blur prediction subnetwork to provide degradation prior and guide the deblurring process. 
% {\color{red}We first enhance the defocus deblurring network with a lightweight blur map estimator for degradation prior estimation and a fusion block for effective prior integration, guiding the deblurring process.}
% %
% Then, we propose a reblurring-based learning framework, as shown in Fig. \ref{fig:reblurnet}, to address the challenge of misaligned training pairs. 
% To ensure the consistent spatial alignment between the deblurred image, the reblurred image and the input blurry image, our reblurring module consists of a kernel prediction network and weight prediction network to reconstruct spatially variant isotropic blur kernels.  
% Moreover, pseudo defocus blur map can be derived from the reblurring module, and is then utilized to guide the training of defocus blur map estimator, thereby expanding the training pairs to training triplets. 
% % Besides, we employ a bidirectional optical flow-based deformation technique to handle misalignment issues. 
%

We first develop a reblurring-based learning framework
% (Fig. \ref{fig:reblurnet}), 
to address the challenge of misaligned training pairs. 
To ensure the consistent spatial alignment between the deblurred image, the reblurred image and the input blurry image, our reblurring module consists of a Kernel Prediction Network (KPN) and a Weight Prediction Network (WPN) to reconstruct spatially variant isotropic blur kernels. 
% (Fig. \ref{fig:isotropic}).  
Moreover, spatially variant blur can be derived from the reblurring module, and serve as pseudo supervision for defocus blur map during training, interestingly expanding the training pairs to training triplets. 
To employ the pseudo supervision, we design a lightweight defocus blur estimator coupled
with a fusion block, which enhances deblurring performance through seamless integration with state-of-the-art deblurring networks.

Furthermore, we introduce a new dataset named SDD for single image defocus deblurring. The image pairs are captured using a HUAWEI X2381-VG camera, which can be adjusted to capture pairs of blurry and sharp images by manipulating the camera motor or aperture size. 
Despite our efforts to maintain alignment during the collection process of the SDD dataset, some misalignment persists due to variations in consumer-grade cameras and collection settings. The misalignment primarily manifests in two forms: zoom misalignment and shift misalignment, as illustrated in Fig.~\ref{intro}. Importantly, the degree of misalignment within the SDD dataset tends to be more severe compared to that observed in the DPDD~\cite{abuolaim2020defocus} dataset, making it be a testbed for evaluating our reblurring-guided image defocus deblurring techniques and serve as a benchmark for future research in this field.

This work is previously presented as a conference paper with oral presentation \cite{li2023learning}, upon which this manuscript has made three major improvements: 
(\emph{i}) We have enhanced the learning framework by deriving pseudo defocus blur maps from the reblurring module and constructing triplets in training datasets. 
(\emph{ii}) Correspondingly, we have improved the deblurring network architecture by incorporating a lightweight defocus blur map estimator coupled with a fusion block. This design not only seamlessly integrates with existing deblurring models but also significantly enhances deblurring performance.
(\emph{iii}) To provide a comprehensive comparison, we have included state-of-the-art methods based on convolutional neural networks (CNN) and Transformer for evaluation, including UformerT \cite{wang2022uformer}, Restormer \cite{zamir2022restormer}, DID-ANet \cite{ma2021defocus} and Loformer \cite{mao2024loformer}.
Furthermore, we have incorporated the DED dataset \cite{ma2021defocus} to evaluate competing methods.

In summary, the contributions of this paper can be summarized as 
\begin{enumerate}
	\item[-]
	A novel reblurring-guided learning framework is proposed for image defocus deblurring that effectively exploits misaligned training pairs for learning deblurring models. 
	
	% \item[-] 
	% A baseline defocus deblurring network based on training triples, where pseudo defocus map supervision can be derived from our reblurring module,  

	\item[-] 
	A lightweight blur map estimator and a fusion block are designed to integrate with state-of-the-art deblurring networks, where the pseudo supervision on spatially variant blur map can be derived from our reblurring module.
	% A baseline defocus deblurring network that integrates a defocus blur map estimator coupled with a fusion block to enable the utilization of predicted degradation prior for enhancement of deblurring performance, 
	
	% \item[-]
	% The pseudo ground truth defocus map can be derived from our reblurring module, serving as supervision for optimizing the defocus blur map estimator, 
	% } 

% {\color{red}\item[-]
	% A fusion block for integrating blurry images and estimated defocus blur maps, which is more adaptable in misaligned training tasks,}

\item[-]
A new dataset named SDD comprising high-resolution image pairs with diverse contents is introduced for evaluating image defocus deblurring models and facilitating future research in this field. 
On benchmark datasets including DPDD \cite{abuolaim2020defocus}, RealDOF \cite{lee2021iterative}, DED \cite{ma2021defocus} and our SDD, our method is significantly superior to existing methods. 

\end{enumerate}

The remainder of this paper is organized as follows: Section \ref{sec2} provides a comprehensive review of the relevant literature, Section \ref{sec3} introduces our proposed method along with the new dataset, Section \ref{sec4} experimentally validates the effectiveness of the proposed approach, and finally Section \ref{sec5} concludes this paper by summarizing key findings.

\section{Related Work}\label{sec2}
In this section, we briefly review relevant works including methods and datasets for defocus deblurring, and reblurring strategies. 

\subsection{Defocus Deblurring Methods}
\label{subsec2}
Traditional defocus deblurring approaches typically focus on estimating a defocus map \cite{goldstein2012blur,xu2010two} by leveraging predefined models. 
These methods then apply non-blind deconvolution techniques \cite{krishnan2009fast,fish1995blind} to restore a sharp image. 
However, their performance is often limited by the inherent constraints of these blur models.
Recent approaches predominantly utilize deep learning to overcome the limitations of the traditional methods by restoring the image directly from the blurred image. 
Abuolaim et al. \cite{abuolaim2020defocus} pioneered the first end-to-end learning-based method DPDNet.
This approach achieved significantly better results than traditional two-stage methods, establishing a new benchmark in the field.
Subsequently, Lee et al. \cite{lee2021iterative} designed a network featuring an iterative filter adaptive module to address spatially varying defocus blur.
Son et al. \cite{son2021single} proposed a kernel prediction adaptive convolution technique that further refines the capacity to address complex defocus patterns.
Despite these advancements, it is crucial to note that the effectiveness of these deep learning-based defcus deblurring methods heavily depends on the quality of the training data, which is primarily derived from the DPDD~\cite{abuolaim2020defocus} dataset.
The dependence on high-quality, well-aligned training pairs has been a persistent challenge, motivating our research to focus on single image defocus deblurring with misaligned training data. 
Although some weakly supervised or unsupervised methods\cite{zhao2023attacking,chen2024unsupervised,tang2023uncertainty} can partially address the reliance on large-scale data, these methods often do not perform as well as supervised approaches.
Recently, All-in-One restoration methods have emerged, aiming to tackle multiple complex and unknown image degradations with a unified model. 
%
% Potlapalli el al. used 
%
Park et al.\cite{park2023all} proposed an adaptive discriminative filter-based model to restore images with unknown degradations.
Some advanced approaches\cite{potlapalli2024promptir,ai2024multimodal,lin2024improving,yang2024ldp,pang2024hir} utilized large language model or diffusion model to effectively restore images from various types and levels of degradation, such as PromptIR \cite{potlapalli2024promptir}.
Ai et al.\cite{ai2024multimodal} harnessed Stable Diffusion priors to further enhance the restoration process which has also shown promising results in the field of defocus deblurring. 
While these approaches have demonstrated commendable performance in defocus deblurring, they heavily rely on the capabilities of large models and sometimes require fine-tuning, which is a relatively labor-intensive process.
Specifically targeting defocus blur, several methods\cite{quan2023single,quan2023neumann} that employ specially designed defocus blur kernels have been proposed. 
These methods address the blur issue with greater precision.
We propose a novel framework designed to overcome the limitations of existing datasets by accommodating misaligned training pairs, thus broadening the applicability of defocus deblurring techniques in real-world scenarios.

\subsection{Defocus Deblurring Datasets}
\label{subsec2}
The availability of high-quality datasets plays a pivotal role in training and evaluating defocus deblurring algorithms.
While several datasets have been proposed for single image deblurring, most synthetic datasets\cite{sun2013edge,nah2019ntire} used for network training primarily focus on specific types of defocus blur, such as Gaussian or disk blur.
Consequently, they often overlook other forms of blur that are commonly encountered in real-world scenarios. 
To address this gap, various datasets have been introduced, such as the DPDD\cite{abuolaim2020defocus} dataset, which was collected using remote control mechanisms to acquire aligned data pairs, exhibiting varying degrees of defocus blur. 
Lee et al.\cite{lee2021iterative} developed the RealDOF dataset using a dual-camera system which is capable of capturing both blurry and sharp images concurrently, offering a novel approach to dataset creation in this field.
However, one common challenge with these datasets is the requirement for precisely aligned image pairs, which limits their applicability in real-world scenarios. 
Ma et al.\cite{ma2021defocus} proposed DED dataset, which used a Lytro Illum light field camera\cite{ng2005light} to collect a dataset with strictly aligned ground truth and input images. 
However, such cameras are not commonly used in everyday life. 
Additionally, the images in the DED dataset\cite{ma2021defocus} have relatively lower resolution compared to existing datasets.
In our research, instead of striving for the construction of a perfectly aligned dataset, we focus on incorporating and addressing misalignments within our network.

\begin{figure*}[!t]
\centering
\setlength{\abovecaptionskip}{-20pt}  
\setlength{\belowcaptionskip}{0pt}  
\hspace*{-0.8cm}  
\includegraphics[width=1.1\linewidth]{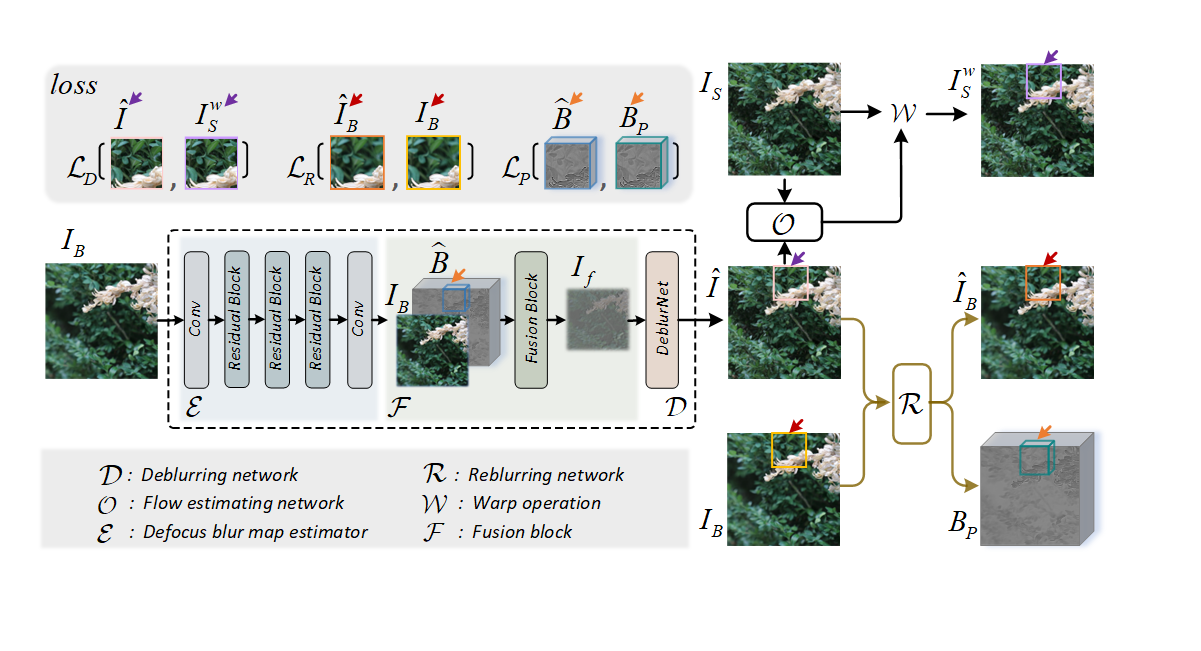}
\vspace{-2em}
\caption{Overview of our reblurring-guided learning framework for image defocus deblurring. 
	It consists of deblurring module and reblurring module. 
	In deblurring module, our introduced blur estimator $\mathcal{E}$ and fusion block $\mathcal{F}$ provide spatially variant degradation priors for enhancing deblurring performance, and can be seamlessly integrated with existing deblurring network $\mathcal{D}$. 
	To tackle spatial misalignment of training pairs, optical flow-based deformation $\mathcal{W}$ is adopted to accommodate misalignment, while reblurring network $\mathcal{R}$ ensures spatial consistency of the deblurred image, the reblurred image and the input blurry image. Pseudo supervision $\bm{B}_P$ of defocus blur map $\hat{\bm{B}}$ can be derived from reblurring network.}
\label{fig pipeline}
\end{figure*}

\subsection{Reblurring Process}
\label{subsec3}
Compared with the end-to-end defocus deblurring methods, reblurring process is less explored in learning-based approaches, yet it holds significant promise for enhancing image restoration tasks, including defocus deblurring. 
Zhang et al.\cite{zhang2020deblurring} introduced a reblurring network designed to generate additional blurry training images using GAN model, while Chen et al.\cite{chen2018reblur2deblur} enhanced the video deblurring process by utilizing separately operated reblurring and deblurring networks.
Furthermore, Lee et al. \cite{lee2021iterative} introduced an additional network that inverts predicted deblurring filters to reblurring filters, and reblurred an all-in-focus image.
These studies show that the reblurring process can beneficially contribute to the deblurring process.
Our reblur module leverages the concept of spatially variant reblurring in defocus deblurring field, where the reblurring process can adapt to different regions of the image, thus better simulating real-world blur phenomena.
Specifically, the reblur module predicts a set of concentric isotropic blur kernels along with corresponding weight maps, which can more accurately simulate the spatial variations of blur found in real-world scenes.
Acknowledging the standalone utility of the reblurring process in previous studies, our research posits that reblurring can also offer valuable prior cues to the deblurring network. 
This integration not only enriches the deblurring process but also utilizes the reblurring stage as a means to provide the deblurring network with insights that guide more effective image restoration.
By jointly training the JDRL network, we aim to recover sharp images using misaligned training data pairs.

\section{Proposed Method}\label{sec3}
Recent advancements in image defocus deblurring have primarily focused on the development of learning-based models using training pairs, denoted as $\{\bm{I}_B^n, \bm{I}_S^n\}_{n=1}^N$, where $\bm{I}_B$ represents a defocus blurry image and $\bm{I}_S$ denotes a ground-truth sharp image. 
However, even with careful alignment during data collection and post-processing techniques, spatial misalignment remains an inevitable issue as depicted in Fig. \ref{intro}.
Furthermore, when considering practical applications with new sensors, the severity of spatial misalignment issues may increase compared to those encountered in DPDD\cite{abuolaim2020defocus}. 
Additionally, it is important to note that different imaging sensors exhibit distinct patterns of defocus blur which limits the generalization ability of learned deblurring models to real-world cases captured by other devices. 
Therefore, for effective deployment on target devices, it is necessary to learn a image defocus deblurring model specifically tailored for each device while relaxing the requirement for perfect alignment in training pairs.

For instance,  we acquire training pairs using a HUAWEI camera, and the models trained on DPDD dataset and DED dataset have limited performance. 
If we employ a pixel-wise loss function, such as Chamober loss \cite{charbonnier1994two}, to train a UNet \cite{ronneberger2015u} based on the misaligned dataset, it may introduce deformation artifacts in the deblurred results, as depicted in Fig. \ref{fig:mis_comp}, where distortions like those highlighted by yellow boxes significantly impact image restoration quality. 
To sum up, learning deblurring models based on misaligned training pairs is both challenging and meaningful.  
In the following, we first provide an overview of our proposed reblurring-based learning framework, which effectively leverages misaligned training pairs to learn image defocus deblurring models. 
Subsequently, we offer detailed explanations on the reblurring module with derivation of pseudo defocus blur map, the deblurring model equipped with a defocus blur map estimator and a fusion block, and finally introduce our newly established dataset.
%Subsequently, we offer detailed explanations on the baseline deblurring model, reblurring process, derivation of pseudo defocus blur map, and finally introduce our newly established dataset.

\subsection{The Overall Framework}\label{subsec3}

\begin{figure}[!t]
\small
\centering
\setlength{\abovecaptionskip}{5pt} 
\setlength{\belowcaptionskip}{0pt}
\setlength{\tabcolsep}{18pt}
% \begin{tabular}{c}	
	\includegraphics[width=0.5\textwidth]{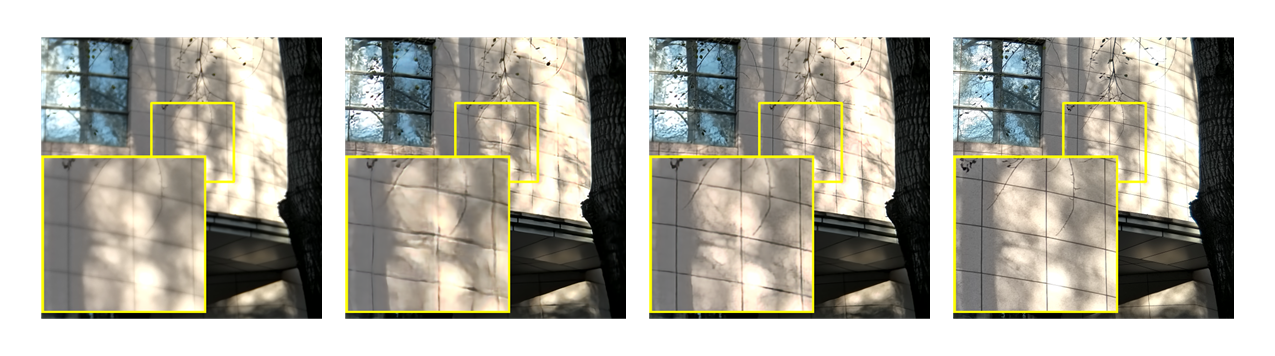}
	% \end{tabular}

\begin{tabular}{cccc}	
	
	\small{Input}&
	\small{UNet}&
	\small{UNet$^\dagger$}&
	\small{GT}
	\\
\end{tabular}
\caption{The direct utilization of pixel-wise loss function for training a UNet model based on misaligned training pairs often leads to deformation artifacts, such as distorted brick lines. 
	These artifacts can be effectively alleviated by our UNet$^\dagger$, where our reblurring-based learning framework Eq. \eqref{eq:jdrl} is adopted for training.
}
\label{fig:mis_comp}
\end{figure}

\begin{figure*}[!t]
\centering\begin{overpic}[width=0.96\textwidth]{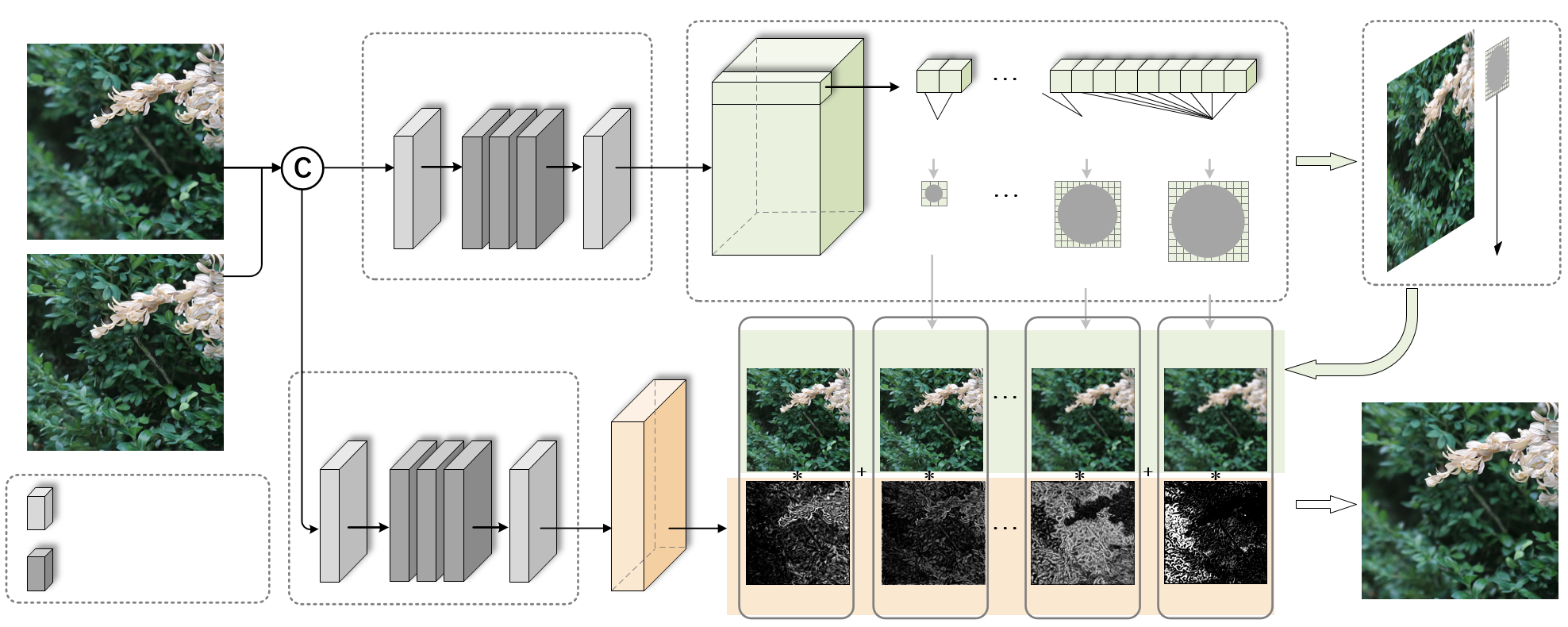}
	\put(-1.2,30){\small $\bm{I}_B$}
	\put(-1.2,17){\small $\hat{\bm{I}}$}
	\put(4.5,7){\tiny Convolution}
	\put(4.5,3){\tiny Residual block}
	\put(92,15){\small $\hat{\bm{I}}_B$}
	\put(50,17){\scriptsize$\hat{\bm{I}}_B^1$}
	\put(58,17){\scriptsize$\hat{\bm{I}}_B^2$}
	\put(67.2,17){\scriptsize$\hat{\bm{I}}_B^{m\!-\!1}$}
	\put(76,17){\scriptsize$\hat{\bm{I}}_B^m$}
	\put(49.3,1){\scriptsize$\bm{W}_1$}
	\put(57.8,1){\scriptsize$\bm{W}_2$}
	\put(66,1){\scriptsize$\bm{W}_{m\!-\!1}$}
	\put(75.5,1){\scriptsize$\bm{W}_m$}
	\put(58.8,30.3){\scriptsize$\mathit{s}_{_\mathrm{2}}$}
	\put(67,30.3){\scriptsize$\mathit{s}_{m\!-\!1}$}
	\put(75.5,30.3){\scriptsize$\mathit{s}_m$}
	\put(58.5,24){\scriptsize$k_2$}
	\put(67.2,22){\scriptsize$k_{m\!-\!1}$}
	\put(75.5,21.3){\scriptsize$k_m$}
	\put(96.6,35){\rotatebox{270}{\tiny Sliding over $\hat{\bm{I}}$}}
	\put(88.8,35.6){\small $\hat{\bm{I}}$}
	\put(29.5,35){\footnotesize $\mathcal{R}_{kpn}$}
	\put(25,13.5){\footnotesize $\mathcal{R}_{wpn}$}
	\put(38.7,0.3){\scriptsize $\bm{W}$}
	\put(48,21.8){\scriptsize $\bm{S}$}
\end{overpic}

\caption{The structure of reblurring network $\mathcal{R}$. There are two branches in $\mathcal{R}$: Kernel Prediction Network $\mathcal{R}_{kpn}$ predicts isotropic defocus blur kernels that are then used to generate blurred images with different blur levels, and Weight Prediction Network $\mathcal{R}_{wpn}$ predicts weight maps for integrating reblurred images.}
\label{fig:reblurnet}
\end{figure*}

Given a training set with $N$ pairs $\{\bm{I}_B^n, \bm{I}_S^n\}_{n=1}^N$, where the blurry image $\bm{I}_B$ is not perfectly aligned with the ground-truth sharp image $\bm{I}_S$, our objective is to learn a deblurring model that effectively addresses misalignment issues while minimizing deformation distortions. In this work, we propose a novel reblurring-guided learning framework, wherein the misalignment problem can be resolved through the integration of a reblurring module. 
%
% As illustrated in Fig. \ref{fig pipeline}, our proposed framework comprises two main components: a deblurring module, and a reblurring module. The deblurring module utilizes a dedicated network $\mathcal{F}$ to process the input blurry image $\bm{I}_B$ and generate an estimated deblurred image $\hat{\bm{I}}$.
%

As illustrated in Fig. \ref{fig pipeline}, our proposed framework comprises two main components: a deblurring network, and a reblurring network. The deblurring netwok utilizes a dedicated network $\mathcal{D}$ to process the input blurry image $\bm{I}_B$ and generate an estimated deblurred image $\hat{\bm{I}}$.

To address misalignment between the ground-truth sharp image $\bm{I}_S$ and its corresponding estimate $\hat{\bm I}$, we introduce an optical flow-based deformation strategy integrated in our approach instead of relying solely on direct pixel-wise loss.
In this manner, the deblurred result $\hat{\bm{I}}$ can adaptively learn sharp textures from the sharp image $\bm{I}_S$, thereby liberating it from the constraints of pixel-level precision. To address potential artifacts caused by optical flow deformation, we introduce a calibration mask and cycle deformation, which are further elaborated in Section \ref{subsec2}. Subsequently, the reblurring module ensures spatial consistency between $\hat{\bm{I}}$ and $\bm{I}_B$. For this purpose, we propose a reblurring network $\mathcal{R}$ tasked with generating a reblurred image  $\hat{\bm{I}}_B$ that closely approximates $\bm{I}_B$. Meanwhile, to ensure spatial coherence between $\hat{\bm{I}}$ and $\bm{I}_B$, the reblurring network $\mathcal{R}$ can also predict the isotropic blur kernels in polar coordinates.

The training loss for learning the parameters in deblurring network $\mathcal{D}$ and reblurring network $\mathcal{R}$ can be formally expressed as 
\begin{equation}\label{eq:jdrl-simple}
\begin{split}
	\mathcal{L} = &\mathcal{L}_{D}(\hat{\bm{I}}, \bm{I}_{S}) + \alpha \mathcal{L}_{R}(\hat{\bm{I}}_B, \bm{I}_B),  
\end{split}
\end{equation}
where $\mathcal{L}_D$ (Eq. \eqref{eq:deblurloss}) is deblurring loss, $\mathcal{L}_R$ (Eq. \eqref{eq:L_R})  is reblurring loss, and $\alpha$ is a trade-off parameter. 
Benefiting from the reblurring-guided training strategy, a Unet can be trained to be free from deformation artifacts, as shown in Fig. \ref{fig:mis_comp}.

In this work, we further suggest that incorporating a defocus blur map can enhance the deblurring performance of deblurring networks. Specifically, an existing deblurring network $\mathcal{D}$ can be incorporated with a defocus blur map estimator $\mathcal{E}$ coupled with a fusion block $\mathcal{F}$ to enable the utilization of predicted degradation prior for enhancement of deblurring performance. The estimator $\mathcal{E}$ predicts the defocus blur map $\hat{\bm{B}}$, while $\mathcal{F}$ incorporates the degradation prior to enhance deblurring performance using a deformable attention mechanism.
% However, the ground-truth defocus blur maps are inherently unattainable, resulting in the lack of reliable supervision for training the estimation network. Therefore, our reblurring module not only ensures the consistent spatial alignment，but also provides the pseudo defocus blur map $\bm{B}_{P}$ to guide the training of the estimator $\mathcal{E}$. In the following
%
The utilization of degradation-related priors as input has been validated in various tasks ~\cite{ma2021defocus,ye2023accurate}, as estimating the degradation is comparatively easier than performing deblurring itself. To enable the training of our baseline deblurring model, we need to prepare training triples, denoted as $\{\bm{I}_B^n, \bm{I}_S^n, \bm{B}^n\}_{n=1}^N$, where $\bm{B}$ represents the ground-truth spatially variant defocus blur map. However, obtaining such ground-truth maps is impractical due to difficulties in capturing spatially variant blur accurately.
While previous work ~\cite{ma2021defocus} estimated defocus blur maps from light field data captured by a Lytro camera, this approach is not suitable for popular consumer cameras. 
% and limits generalization across different devices. 
Fortunately, our proposed reblurring module provides a means to obtain pseudo ground-truth defocus maps $\bm{B}_P$ easily for each input blurry image and serves as supervision for the defocus blur map estimator $\mathcal{E}$ within our deblurring model.
Therefore, based on the training triplets $\{\bm{I}_B^n, \bm{I}_S^n, \bm{B}_P^n\}_{n=1}^N$, the final training loss can be defined as 
\begin{equation}\label{eq:jdrl}
\begin{split}
	\mathcal{L} \!=\! &\mathcal{L}_{D}(\hat{\bm{I}}, \bm{I}_{S}) \!+\! \alpha \mathcal{L}_{R}(\hat{\bm{I}}_B, \bm{I}_B) \!+\! \beta \mathcal{L}_{P}(\hat{\bm{B}}, \bm{B}_{P}^{}),  
\end{split}
\end{equation}
where $\hat{\bm{B}} = \mathcal{E}({\bm{I}_B})$ denotes the estimated defocus blur map, $\mathcal{L}_P$ (Eq. \eqref{eq:L_p}) is a loss function for defocus blur map, and $\beta$ is a positive trade-off parameter. 

In this way, we have developed a robust framework that enables the deblurring network to effectively leverage spatially adaptive sharp textures from the ground-truth sharp image while maintaining spatial consistency with the blurry input. During inference, our baseline deblurring model can generate latent sharp images while discarding the reblurring module.
%{\color{blue} In the following, we will introduce the spatially invariant reblurring module, and the deblurring module equipped with defocus blur map estimator and fusion block in details.}

\begin{figure*}[htbp]
\centering\begin{overpic}[width=0.95\textwidth]{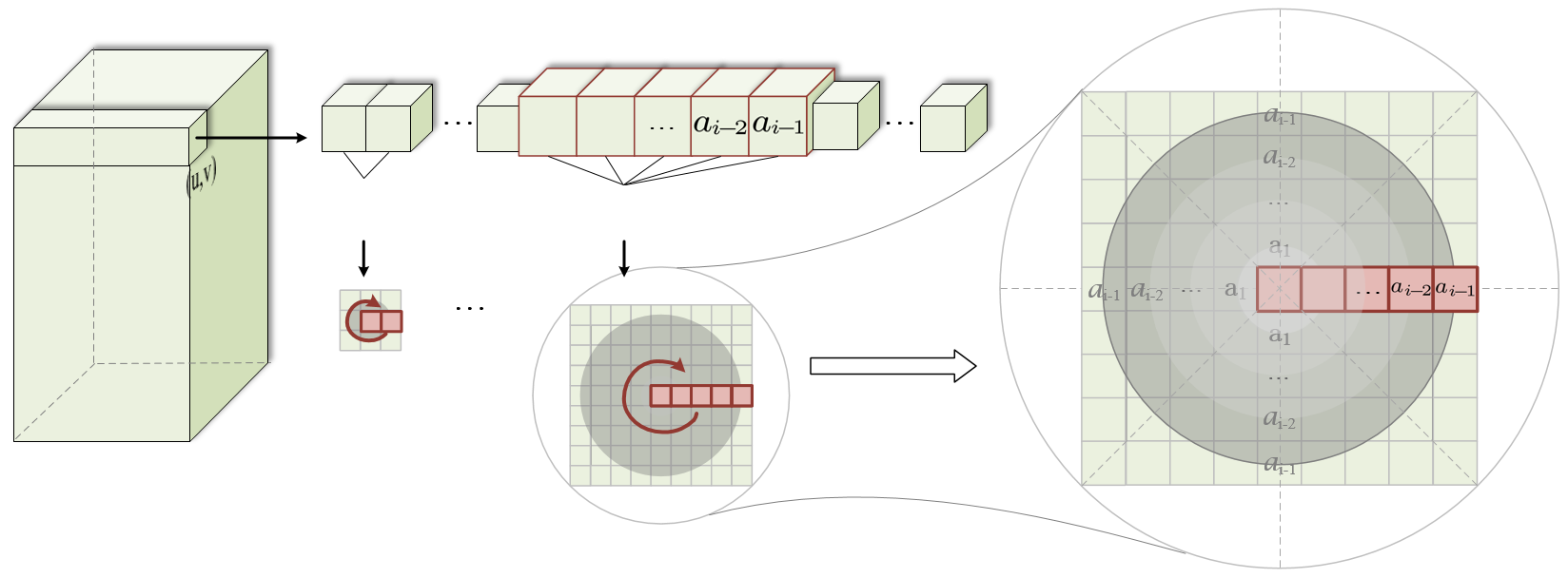}
	\put(4.9,5){\small $\bm{S}$}
	\put(16.5,21.7){\scriptsize $H$}	
	\put(14,9.8){\scriptsize $W$}
	\put(9,34){\scriptsize $M$}	
	\put(34,28.5){\scriptsize $a_0$}
	\put(37.8,28.5){\scriptsize $a_1$}
	\put(80.6,17.8){\scriptsize $a_0$}
	\put(83.2,17.8){\scriptsize $a_1$}
	\put(88,25){\scriptsize $\rho\!=\!{i\!\!-\!\!1}$}
	\put(97,17.8){\scriptsize {\textcolor[RGB]{127,127,127}{$0^\circ$}}}
	\put(80,34){\scriptsize {\textcolor[RGB]{127,127,127}{$90^\circ$}}}
	\put(64,17.8){\scriptsize {\textcolor[RGB]{127,127,127}{$180^\circ$}}}
	\put(80,1){\scriptsize {\textcolor[RGB]{127,127,127}{$270^\circ$}}}
	\put(52,15){\scriptsize zoom in}	
	\put(21.5,23){\scriptsize $\mathit{s}_{\mathrm{2}}^\mathit{u,v}$}	
	\put(39,22.5){\scriptsize $\mathit{s}_{\mathrm{i}}^\mathit{u,v}$}	
	\put(22,11.5){\scriptsize $\mathit{k}_2^\mathit{u,v}$}	
	\put(40.5,1.6){\scriptsize $\mathit{k}_{\mathrm{i}}^\mathit{u,v}$}
	
\end{overpic}
\caption{Illustration of how isotropic blur kernels are generated. 
	For a feature vector located at coordinate $(u,v)$, it is first split into a set of kernel seeds $\{s_i^{u,v}\}$ and then converted to blur kernels $\{k_i^{u,v}\}$. 
	Referring to Eq. \eqref{eq:kernel generation}, the value of each element in $k_i^{u,v}$ is interpolated in polar coordinates by considering the distance between this element and the center of $k_i^{u,v}$.}
\label{fig:isotropic}
\end{figure*}

\subsection{Reblurring Module} 
\label{subsec3}
%

% In the preceding section, we outlined the overall framework of the reblurring-guided learning approach. 
% In the deblurring module, we develop design a lightweight defocus map estimator $\mathcal{E}$ to predict the defocus blur map $\hat{\bm{B}}$ and a fusion block $\mathcal{F}$ to integrate blurry image $\bm{I}_{B}$ and $\hat{\bm{B}}$, both of which can be seamlessly integrated into various existing deblurring models $\mathcal{D}$, enabling the incorporation of degradation priors to enhance the processing of blurry image.
% %
% To ensure spatial consistency between the deblurred image $\hat{\bm{I}}$ and the input blurry image $\bm{I}_{B}$, we introduce the spatially invariant reblurring module $\mathcal{R}$, which comprises a Kernel Prediction Network (${\mathcal{R}_{kpn}}$) and a Weight Prediction Network (${\mathcal{R}_{wpn}}$), along with its associated training loss $\mathcal{L}_{R}$. 
% %
% By utilizing the outputs of KPN and WPN, we further generate the pseudo ground-truth defocus blur map $\bm{B}_{p}$, which serves as supervision for training the defocus blur map estimator $\mathcal{E}$ within the deblurring module. 
% In the preceding section, we outlined the overall framework of the reblurring-guided learning approach. 
In this section, We first introduce our reblurring network along with its loss function $\mathcal{L}_R$, and then present the acquisition of pseudo defocus blur map $\bm{B}_P$.

%

% Following the deblurring process, the reblurring module is employed to enhance the spatial consistency of the deblurred image. The objective of the reblurring module $\mathcal{R}$ is to maintain the spatial coherence between $\hat{\bm{I}}$ and $\bm{I}_B$. A spatially invariant reblurring network $\mathcal{R}$ is utilized to reintroduce blurriness into $\hat{\bm{I}}$, guided by predicted isotropic blurring kernels. Subsequently, pixel-wise loss is applied to supervise $\hat{\bm{I}}_B$ with respect to its correspondence with $\bm{I}_B$.
%

\subsubsection{Reblurring Network}
As illustrated in Fig.~\ref{fig:reblurnet}, reblurring network $\mathcal{R}$ consists of a Kernel Prediction Network $\mathcal{R}_{kpn}$ and a Weight Prediction Network $\mathcal{R}_{wpn}$. 

\textbf{Kernel Prediction Network} $\bm{\mathcal{R}_{kpn}}$:
The objective of $\mathcal{R}_{kpn}$ is to predict the blur kernel for each pixel. Defocus blur typically arises from a circular aperture, resulting in symmetric and uniformly distributed blur spots in all directions. When defocusing occurs, the optical characteristics of the lens usually cause light to scatter uniformly, indicating that the defocus blur kernel can be considered isotropic in nature. Initially, ${\mathcal{R}_{kpn}}$ predicts kernel seeds. The functionality of ${\mathcal{R}_{kpn}}$ can be described as
\begin{equation}
\bm{S} = \mathcal{R}_{kpn}(\hat{\bm{I}}, \bm{I}_B),
\end{equation}
where the input consists of the concatenation of $\hat{\bm{I}}$ and $\bm{I}_B$, both having dimensions $H\times W \times 3$, while the output, denoted as $\bm{S}$, is a feature volume with dimensions $H \times W \times M$.
The feature vector of size $ 1 \times 1 \times M$ corresponding to each position $(u,v)$ is partitioned into a set of kernel seeds $\{s_i^{u,v} \ |\  i=2,3,...,m,\ M=\sum_{i=2}^{m}i\}$ for further processing. These seeds are utilized to generate a collection of isotropic kernels. Each kernel $k_i^{u,v}$ represents a single-channel map with dimensions $(2i-1)\times(2i-1)$. The process of generating the kernels is illustrated in Fig.~\ref{fig:isotropic}.
Let $s_i^{u,v}=[a_0, a_1, ..., a_{i-1}]^{\rm T}$ denote the kernel seed of position $(u,v)$, we explain in detail how the corresponding kernel $k_i^{u,v}$ is generated. 

For a single element of $k_i^{u,v}$, its value is determined by its distance from the center of $k_i^{u,v}$ using interpolation in polar coordinates. Specifically, we first represent the elements of $k_i^{u,v}$ with the form $(\rho,\theta)$ in polar coordinates. 
To formalize this process, we represent the elements of $k_i^{u,v}$ as $(\rho,\theta)$ in polar coordinates

% \begin{spacing}{1}
\begin{equation}\label{eq:kernel generation}
	\begin{split}
		k_i^{u,v}(\rho,\theta) \!=\! 
		\left\{
		\begin{aligned}
			& a_{\rho}\ ,\ {\rm if} \ \rho\le i-1 \ {\rm and}\ \rho\  {\rm is} \ {\rm integer}\\
			& 0\ ,\ {\rm if}\ \rho>i-1\\
			& \dfrac{\rho \!-\! \lceil\rho\rceil}{\lfloor\rho\rfloor \!-\! \lceil\rho\rceil}a_{\lfloor\rho\rfloor} \!+\! \dfrac{\rho \!-\! \lfloor\rho\rfloor}{\lceil\rho\rceil \!-\! \lfloor\rho\rfloor}a_{\lceil\rho\rceil},\ {\rm else}
		\end{aligned}
		\right.
	\end{split}
\end{equation}
% \end{spacing}
where $\lfloor\rfloor$ and $\lceil\rceil$ denote the floor and ceiling operations. 
The calculated kernel values are then normalized using a \texttt{Softmax} function, ensuring that $\sum_{\rho,\theta} k_i^{u,v}(\rho,\theta) = 1$ and $k_i^{u,v}(\rho,\theta)>0$. Through these operations, we obtain $m-1$ pixel-wise isotropic defocus blur kernels of sizes $\{3\times 3, 5\times 5, ..., (2m-1)\times (2m-1)\}$ to account for different positions of $\hat{\bm{I}}$. Subsequently, $m-1$ blurry images $\{\hat{\bm{I}}_B^2, ..., \hat{\bm{I}}_B^{m}\}$ at various levels of blur can be generated by convolving $\hat{\bm{I}}$ with the corresponding blur kernels. It is important to note that the zero-blur-level image $\hat{\bm{I}}_B^1$ corresponds to the non-blurred image $\hat{\bm{I}}$.
\textbf{Weight Prediction Network }$\bm{\mathcal{R}_{wpn}}$:
The generation of weight maps to integrate the estimated blurry images is crucial due to the spatial variability of blur across a blurry image. The operation $\mathcal{R}_{wpn}$ can be mathematically expressed as
\begin{equation}
\bm{W} = \mathcal{R}_{wpn}(\hat{\bm{I}}, \bm{I}_B),
\end{equation}
where the feature volume $\bm{W}$ is of size $ H\times W \times m$ and is subsequently divided into $m$ weight maps, each with a single channel ($\bm{W}_1, \bm{W}_2, ..., \bm{W}_m$). These weight maps are then normalized using the \texttt{Softmax}  function to ensure that the values at each position $(u,v)$ sum up to 1, \emph{i.e}., $\sum_i \bm{W}_i^{u,v} = 1$, where $\bm{W}_i^{u,v}\ge0$.
\subsubsection{Reblurring Loss $\mathcal{L}_R$}
The reblurred image can be reconstructed as
\begin{equation}\label{eq:reblur}
\hat{\bm{I}}_B = \sum_{i=1}^{m}\bm{W}_i*\hat{\bm{I}}_B^i.
\end{equation}
The reblurring loss is specified as
% \begin{equation} \label{eq:L_R}
% {\mathcal{L}_{R}}=\sqrt{\parallel \color{red}{(}\bm{I}_B - \hat{\bm{I}}_B) \parallel^{2} + \varepsilon^2},
% \end{equation}
\begin{equation} \label{eq:L_R}
{\mathcal{L}_{R}}=\sqrt{\parallel \color{black}{\bm{I}_B - \hat{\bm{I}}_B} \parallel^{2} + \varepsilon^2},
\end{equation}
where $\varepsilon=1\times 10^{-3}$ is set in experiments. 
It is observed that the spatial consistency among $\hat{\bm{I}}$, $\hat{\bm{I}}_B$, and ${\bm{I}}_B$ is guaranteed due to the isotropic characteristics of the predicted blur kernels. In terms of network architectures, simple networks consisting of fundamental convolutional layers and residual blocks are utilized for both $\mathcal{R}_{kpn}$ and $\mathcal{R}_{wpn}$, as depicted in Fig. \ref {fig:reblurnet}.

%\subsubsection{Defocus Map Loss $\mathcal{L}_P$ with Pseudo Supervision}
The final issue lies in the acquisition of pseudo defocus map supervision $\bm{B}_P$. As shown in Eq. \eqref{eq:reblur}, a reblurred image can be considered as a weighted summation of blurred images generated by isotropic blur kernels. Consequently, the blur amount measurement at a pixel coordinate in the blurred image can be obtained through the utilization of weighted kernel seeds. Therefore, obtaining $\bm{B}_P$ with size $H\times W \times M$ involves assigning weights $\bm{W}$ with size $H \times W \times m$ to the kernel seeds $\bm{S}$ with size $H \times W \times M$. 

\subsubsection{{Derivation of Pseudo Defocus Map}}
\label{subsec:pseudo}
\begin{figure}[!t]
\small
\centering
\vspace{-3em}
\setlength{\abovecaptionskip}{0pt} 
\setlength{\belowcaptionskip}{0pt}
\includegraphics[width=0.5\textwidth]{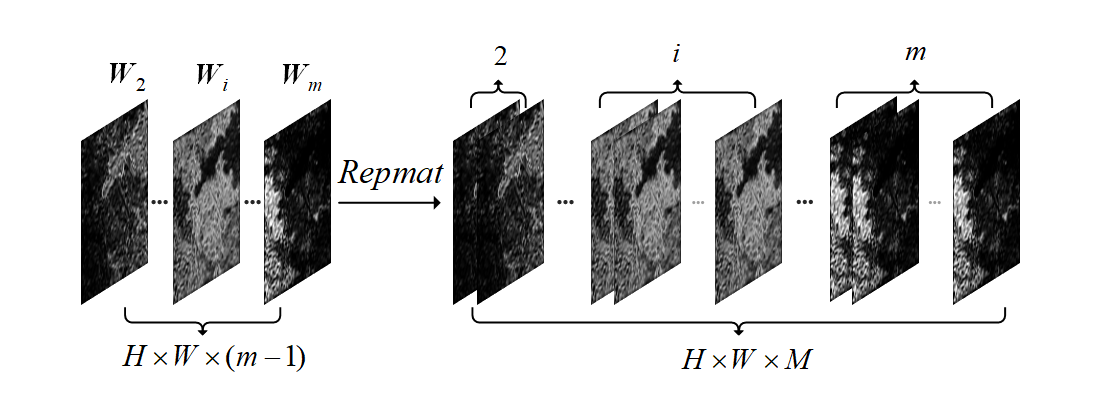}
\vspace{-2em}
\caption{{Visualization of \textit{Repmat}: The matrix $\bm{W}_i$ is replicated $i$ times, enabling feasible element-wise multiplication with $\bm{S} \in \mathbb{R}^{H \times W \times M}$, where $M = \sum_{i=2}^{m} i$. Note that $\bm{W}_1$, which corresponds to the zero-blur-level image $\hat{\bm{I}}_B^1$, does not contribute to the blur amount, and thus is excluded before \textit{Repmat} operation.}}
\label{fig:repmat}
\end{figure}

The final issue lies in the acquisition of pseudo defocus map supervision $\bm{B}_P$. As shown in Eq. \eqref{eq:reblur}, a reblurred image can be considered as a weighted summation of blurred images generated by isotropic blur kernels. Consequently, the blur amount measurement at a pixel coordinate in the blurred image can be obtained through the utilization of weighted kernel seeds. Therefore, obtaining $\bm{B}_P$ with size $H\times W \times M$ involves assigning weights $\bm{W}$ with size $H \times W \times m$ to the kernel seeds $\bm{S}$ with size $H \times W \times M$. 
We note that $i$-th kernel seed at a pixel coordinate has $i$ values, and the summation $M = \sum_{i=2}^{m}i$ satisfies. 
To ensure dimension compatibility, each channel $\bm{W}_i$ with size $H\times W\times 1$ needs to be expanded to match the dimension $H\times W\times i$ of the corresponding $i$-th kernel seed in $\bm {S}$. Formally, the pseudo defocus map $\bm {B}_{P}$ can be obtained by  
\begin{equation}
\bm {B}_{P}= \textit {Repmat}(\bm {W}) \odot \bm {S},
\end{equation}
where $\odot$ is element-wise multiplication, and $\textit {Repmat}$ duplicates matrix $\bm {W}_i$ for $i$ times to ensure that kernel seeds $\bm{s}_i$ and their corresponding weighting matrices ${\bm{{W}}}_i$ have matching dimensions. {This process is shown in Fig.~\ref{fig:repmat}}.  
Subsequently, pseudo defocus map $\bm {B}_{P}$ will be utilized to train the deblurring module as presented in Sec. \ref{subsubsec:deblurring_loss}, supervising the estimation of defocus blur map.
%
%
% Finally, during training of the prediction network, pseudo defocus map ${\bm{{B}}_{P}}$ serves as supervision
% \begin{equation}
% {\mathcal{L}_{P}}=\sqrt{\parallel \bm{B}_P - \hat{\bm{B}} \parallel^{2} + \varepsilon^2}.
% \end{equation}

\subsection{Deblurring Module}\label{subsec2}

% In the deblurring module, we introduce a baseline deblurring model $\mathcal{D}$ along with its training loss $\mathcal{L}_D$. Typically, it is equipped with a lightweight defocus blur map estimator $\mathcal{E}$ and a fusion block $\mathcal{F}$ for integrating the blurry image and estimated defocus blur map. By leveraging this loss function $\mathcal{L}_D$, the deblurred output $\hat{\bm{I}}$ obtained from $\mathcal{F}$ can dynamically acquire high-resolution textures from the sharp image $\bm{I}_S$, thereby transcending the limitations imposed by pixel-level precision. And the introduced $\mathcal{E}$ and $\mathcal{F}$ can be seamlessly incorporated into existing deblurring networks $\mathcal{D}$ without altering their original architecture.
We present the deblurring network, and the training loss functions $\mathcal{L}_D$ and $\mathcal{L}_P$.

%

%
% TODO: caption是不是太长了，放到正文中？

\subsubsection{Deblurring Model}
Upon existing deblurring network $\mathcal{D}$, \emph{e.g.}, the CNN-based or Transformer-based deblurring networks, we introduce a blur map estimator $\mathcal{E}$ and a fusion block $\mathcal{F}$, by which the estimated blur map $\hat{\bm{B}}=\mathcal{E}(\bm{I}_B)$ and fused image $\bm{I}_f=\mathcal{F}(\bm{I}_B, \hat{\bm{B}})$.  
Taking the blurry image $\bm{I}_B$ as input, the deblurring module can be formulated as 
\begin{equation}
\hat{\bm{I}} = \mathcal{D}(\bm{I}_f) = \mathcal{D}\left(\mathcal{F}\left(\bm{I}_B, \mathcal{E}({\bm{I}_B})\right)\right).
\end{equation}
Since we do not modify the architecture of $\mathcal{D}$, we in the following only focus on blur map estimator $\mathcal{E}$ and fussion block $\mathcal{F}$. 
%
%Our baseline deblurring model $\mathcal{D}$ takes the blurry image $\bm{I}_B$ as input and generates an intermediate defocus blur map $\hat{\bm{B}}$ and a latent sharp image $\hat{\bm{I}}$. The estimated defocus blur map, denoted as $\hat{\bm{B}}$, serves as a degradation prior to enhance the deblurring performance within the deblurring network. Specifically, our baseline model $\mathcal{F}$ consists of two subnetworks: defocus blur map estimator $\mathcal{E}$ and deblurring sub-network $\mathcal{F}_D$, which collectively produce the final deblurred output
%\begin{equation}
%\hat{\bm{I}} = \mathcal{F}_D(\bm{I}_B, \mathcal{F}_B({\bm{I}_B})).
%\end{equation}

\textbf{Defocus Blur Map Estimator $\mathcal{E}$:}
The defocus blur map estimator is designed to predict the defocus blur map, denoted as $\hat{\bm{B}} = \mathcal{E}(\bm{I}_{B})$. This lightweight network consists of 2 convolutional layers and 3 residual layers. However, since no direct supervision is available for the estimated defocus blur map $\hat{\bm{B}}$, it may result in significant deviation from the true defocus blur map. To address this issue, we utilize the pseudo ground-truth defocus maps generated by our reblurring module, as described in Sec. \ref{subsec:pseudo}, and introduce a dedicated loss term $\mathcal{L}_{P}$ to optimize the estimator.

\begin{figure}[!t]
\small
\centering
\vspace{-3em}
\setlength{\abovecaptionskip}{0pt} 
\setlength{\belowcaptionskip}{0pt}
\includegraphics[width=0.5\textwidth]{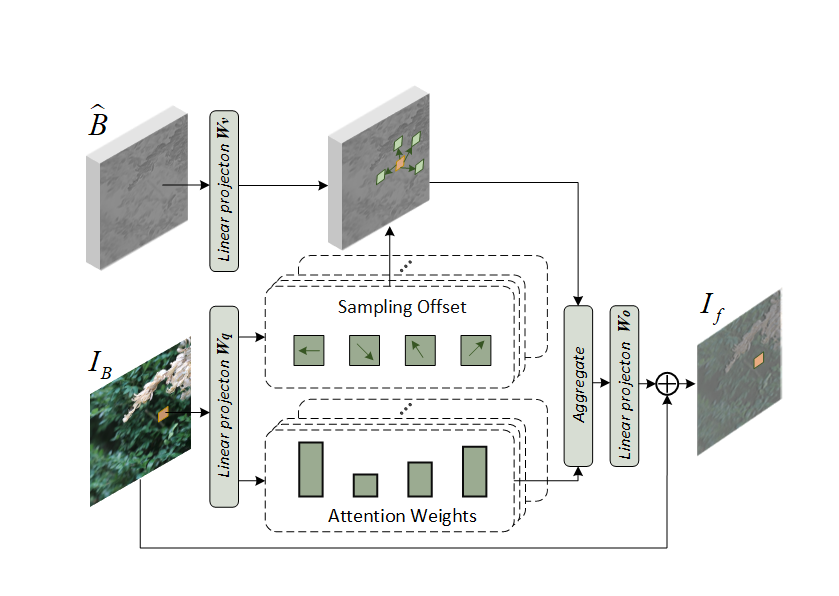}
\vspace{-2em}
\caption{Overview of our fusion block for integrating defocus map and input blurry image, which incorporates a deformable attention mechanism that better aligns the blur prior with the image content. The fused image $\bm{I}_f$ is of size $H\times W \times 3$ that can be fed to existing deblurring networks without architecture modification. }
\label{fig:fusion}
\end{figure}

\textbf{Fusion Block $\mathcal{F}$:} 
Empirically, we observe that the defocus blur map and blurry image also have misalignment issues, especially for early training epochs. Therefore, we explore the fusion strategy by employing deformable attention mechanism as illustrated in Fig.~\ref{fig:fusion} to effectively incorporate the degradation-related prior for image deblurring.
The deformable attention mechanism allows the model to focus on relevant spatial regions of the pseudo defocus blur maps while dynamically adjusting the attention weights based on the local features of the blurred image.

{Given the estimated defocus blur map $\hat{\bm{B}} \in \mathbb{R}^{H \times W \times M}$ and the blurry input image $\bm{I}_B \in \mathbb{R}^{H \times W \times 3}$, we introduce a deformable cross-attention mechanism to adaptively fuse information across different feature representations. For each pixel coordinate $(x, y)$ in the spatial domain, the fusion process of the deformable cross-attention is formulated as follows:
\begin{equation}\label{eq:deformable}
	% \begin{split}
		% \bm{I}_f(x,y) = \bm{I}_{B}(x,y) + 
		% \bm{W}_o \sum_{n=1}^{N_p} \left( \bm{W}_{q(n)} \bm{I}_{B}(x,y) \right) \cdot \left( \bm{W}_{v(n)} \hat{\bm{B}}{\left(x + \Delta x_n,\ y + \Delta y_n\right)} \right),    
		% \end{split}
	\begin{split}
		\bm{I}_f(x,y) = & \bm{I}_{B}(x,y) \!+\! 
		\bm{W}_o \sum_{n=1}^{N_p} \left( \bm{W}_{q(n)} \bm{I}_{B}(x,y) \right) \odot \\
		& \left( \bm{W}_{v(n)} \hat{\bm{B}}{\left(x \!+\! \Delta x_n,\ y \!+\! \Delta y_n\right)} \right),
	\end{split}
\end{equation}
where $\odot$ is element-wise multiplication, and $N_p$ is the number of deformable sampling points, which controls the spatial scope of attention. 
Specifically, $\bm{I}_{B}(x,y) \in \mathbb{R}^{3 \times 1}$ denotes the feature vector of the input image $\bm{I}_B$ at position $(x, y)$, while $\hat{\bm{B}}{(x, y)} \in \mathbb{R}^{M \times 1}$ represents the corresponding feature from the blur map $\hat{\bm{B}}$ sampled at position $(x, y)$. 
The learnable weights $\bm{W}_{q(n)} \in \mathbb{R}^{M \times 3}$ and $\bm{W}_{v(n)} \in \mathbb{R}^{M \times M}$ are used to project features in blurry image and defocus blur map, respectively. 
% the $n$-th components of the weight matrices $W_q \in \mathbb{R}^{N_p \times 3}$ and $W_v \in \mathbb{R}^{N_p \times M \times M}$, respectively. 
$\Delta x_n $ and $\Delta y_n $ are the spatial offsets. $\bm{W}_o \in \mathbb{R}^{3 \times M}$ is the output projection matrix that maps the aggregated features back to the original image dimension. }
The fused image $\bm{I}_f \in \mathbb{R}^{H\times W \times 3}$ can then be fed to existing deblurring network without modifying on architecture. 
This strategy overcomes the limitations of traditional fixed attention mechanisms by enabling more flexible integration of information from defocus blur maps, leading to better alignment of the blur prior with the image content.

The architecture of $\mathcal{F}$ employs 5 attention heads, with each head using 4 sampling points. Each attention head
independently computes its attention weights and feature fusion process, and the outputs from all heads are
subsequently merged that enables the model to focus on relevant spatial regions of the pseudo defocus blur maps.
%The processed defocus map and blurred image are then fed into $\mathcal{F}_D$ composed of transformer blocks for deblurring.
%
%The deblurring $\mathcal{F}_D$ has a 4-level symmetric encoder-decoder as \cite{zamir2022restormer}.
%
The parameters of $\mathcal{D}$, $\mathcal{E}$ and $\mathcal{F}$ can be learned by optimizing a single deblurring loss $\mathcal{L}_D$. 
Moreover, to supervise the estimated defocus blur map $\hat{\bm{B}}$, we also introduce the dedicated loss $\mathcal{L}_P$.
%However, this approach does not provide supervision for the estimated defocus blur map $\hat{\bm{B}}$, which may result in significant deviation from the true defocus blur map. To address this issue, we introduce an additional loss term $\mathcal{L}_P$ based on the pseudo defocus blur map $\bm{B}_P$. %, as explained in Section \ref{subsec:pseudo}.
% In this section, we mainly focus on the deblurring loss function $\mathcal{L}_D$ and the $\mathcal{L}_P$.
%

\begin{figure}[!t]
\centering
\setlength{\abovecaptionskip}{3pt} 
\setlength{\belowcaptionskip}{0pt}
\begin{overpic}[width=0.48\textwidth]{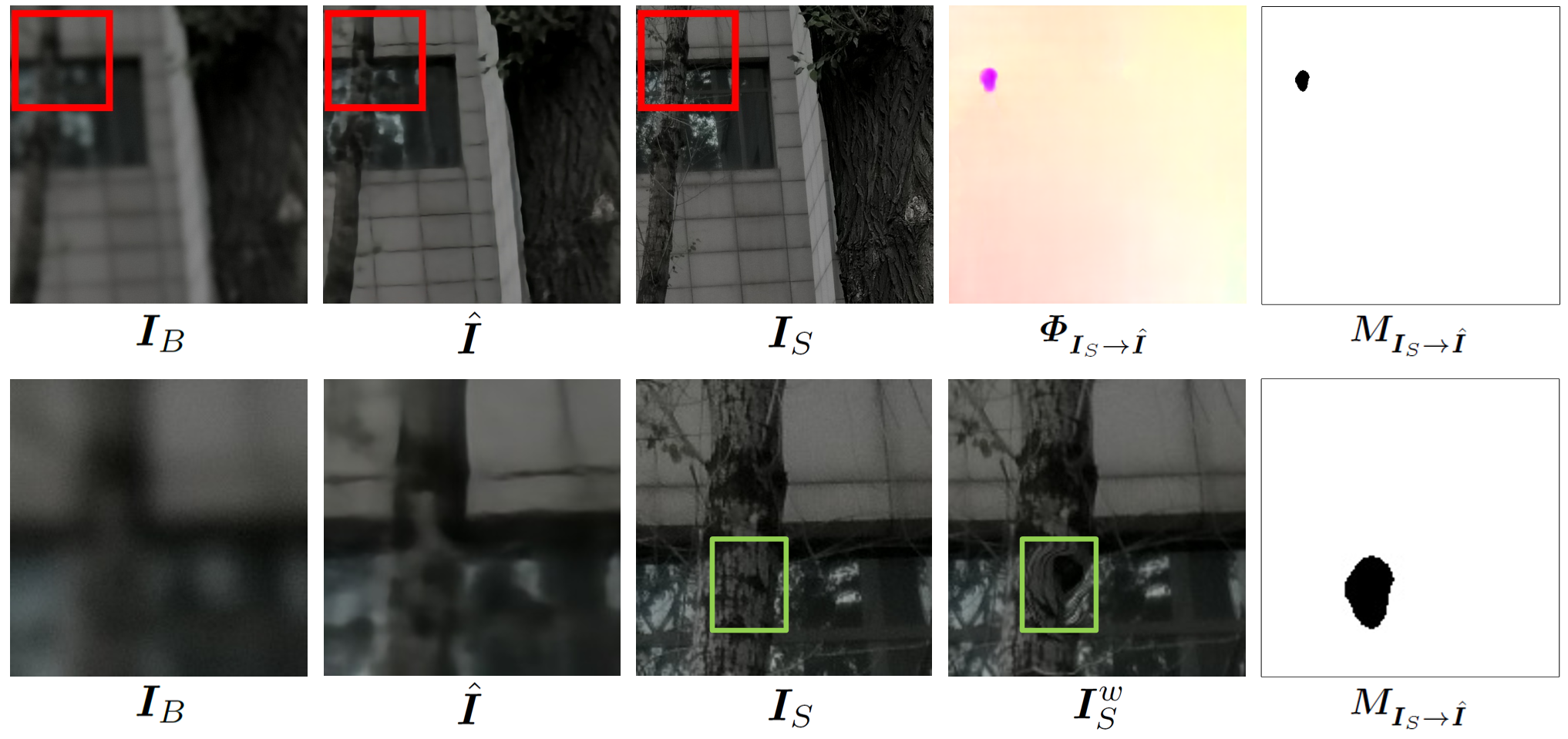}
\end{overpic}
\caption{\small Errors in optical flow estimation can incorrectly deform the sharp ground-truth image, \emph{e.g}., in areas marked by green rectangles. Calibration masks can help by filtering out this adverse region.
}
\label{fig:mask}
\end{figure}

\subsubsection{Deblurring Losses $\mathcal{L}_D$ and $\mathcal{L}_{P}$}
\label{subsubsec:deblurring_loss}
To account for the spatial misalignment inherent in the training pairs, we incorporate an optical flow-based deformation. This approach allows the framework to accommodate potential misalignment between $\bm{I}_S$ and $\hat{\bm{I}}$. Specifically, we employ an optical flow estimation network $\mathcal{F}_{flow}$ \cite{sun2018pwc} to estimate the optical flow $\bm{\Phi}_{\bm{I}_{S}\rightarrow\hat{\bm{I}}}$ from $\bm{I}_S$ to $\hat{\bm{I}}$
\begin{equation}\label{eq:flow}
\bm{\Phi}_{\bm{I}_{S}\rightarrow\hat{\bm{I}}} = \mathcal{F}_{flow} (\bm{I}_S, \hat{\bm{I}}). 
\end{equation}
Then, $\bm{I}_S$ is deformed towards $\bm{\hat{I}}$ using the estimated optical flow
\begin{equation}
\bm{I}_S^{w} = \mathcal{W} (\bm{I}_S, \bm{\Phi}_{\bm{I}_{S}\rightarrow\hat{\bm{I}}}),
\end{equation}
where $\mathcal{W}$ denotes linear warping operation \cite{sun2018pwc}. 

We utilize a calibration mask, denoted as $\bm{M}$, to identify and exclude regions with inaccurate optical flow estimations. 
The process begins by calculating the average optical flow value, denoted
by $\overline{\bm{\Phi}}$. Based on this, the calibration mask is defined as follows
\begin{equation}\label{eq:mask}
\begin{split}
	\bm{M}_{\bm{I}_{S}\rightarrow\hat{\bm{I}}} = [&(1-\lambda) \times \overline{\bm{\Phi}}_{\bm{I}_{S}\rightarrow\hat{\bm{I}}}<\bm{\Phi}_{\bm{I}_{S}\rightarrow\hat{\bm{I}}}\\
	&<(1+\lambda)\times \overline{\bm{\Phi}}_{\bm{I}_{S}\rightarrow\hat{\bm{I}}}],
\end{split}
\end{equation}
where the value of $\bm{M}$ is 1 if the condition in $[\cdot]$ is satisfied, and otherwise 0.

The effectiveness of Eq.~\eqref{eq:mask} can be attributed to the slight variation in the misalignment between $\bm{I}_S$ and $\hat{\bm{I}}$ across different spatial positions, enabling the detection of inaccurate optical flow estimation through anomalies in its magnitude. This process is illustrated in Fig. \ref{fig:mask}. Additionally, we incorporate a cycle deformation strategy by calculating the optical flow $\bm{\Phi}_{\hat{\bm{I}}\rightarrow \bm{I}_{S}}$ in reverse order and applying reverse deformation, significantly enhancing the robustness of the deformation process. 
Finally, the adaptive deblurring loss can be calculated based on Charbonnier loss~\cite{charbonnier1994two}
\begin{equation}\label{eq:deblurloss}
\begin{split}
	{ \mathcal{L}_{D}}=&\sqrt{\parallel \bm{M}_{\bm{I}_{S}\rightarrow\hat{\bm{I}}}*(\bm{I}_S^{w} - \hat{\bm{I}}^{}) \parallel^{2} + \varepsilon^2} \\
	&+ \sqrt{\parallel \bm{M}_{\hat{\bm{I}}\rightarrow \bm{I}_{S}}*(\hat{\bm{I}}^{w} - \bm{I}_S) \parallel^{2} + \varepsilon^2},
\end{split}
\end{equation}
where $*$ is element-wise product, and $\varepsilon$ is empirically set as $1\times 10^{-3}$ in all the experiments.

To address the significant deviation of the estimated defocus blur map $\hat{\bm{B}}$ from the true defocus blur map, we introduce an additional loss term $\mathcal{L}_P$ that uses the pseudo supervision $\bm{B}_P$ derived in Section~\ref{subsec:pseudo} as supervision
\begin{equation} \label{eq:L_p}
{\mathcal{L}_{P}}=\sqrt{\parallel \bm{B}_P - \hat{\bm{B}} \parallel^{2} + \varepsilon^2}.
\end{equation}

% \vspace{-2em}

\subsection{A New Dataset for Image Defocus Deblurring}
To validate the effectiveness of our approach on a specific device, we employed a HUAWEI X2381-VG camera to establish a new dataset for image defocus deblurring, referred to SDD dataset. This camera has adjustable DOF through vertical lens movement and aperture modulation, allowing us to collect pairs of blurry images and corresponding ground-truth sharp images from the same scene. Despite our best efforts to ensure proper alignment between the blurry images and ground-truth sharp images, the SDD dataset exhibits noticeable misalignment in training pairs compared to DPDD~\cite{abuolaim2020defocus}, as depicted in Fig. \ref{intro}.
This discrepancy can be attributed to adjustments made by using an electronic motor to manipulate the camera lens.  

\begin{figure}[!t]
\centering
\setlength{\abovecaptionskip}{5pt} 
\setlength{\belowcaptionskip}{0pt}
\begin{overpic}[width=0.48\textwidth]{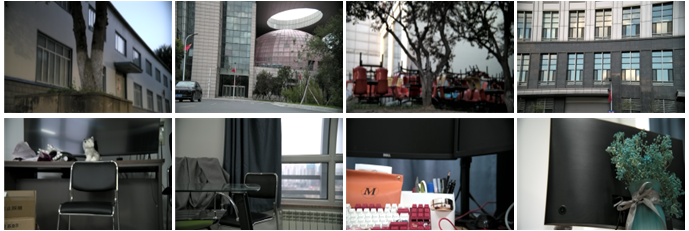}
	
\end{overpic}
\caption{\small Sample images from our SDD dataset. The first and second rows show the outdoor and indoor scenes respectively. 
}
\label{fig:data}
\end{figure}

This dataset comprises 150 high-resolution blurry and sharp image pairs with dimensions $4096\times 2160$. 
Sample images from the SDD dataset are shown in Fig.~\ref{fig:data}.
These pairs are divided into 115 training pairs and 35 testing pairs. Similar to \cite{abuolaim2020defocus}, the training image pairs are resized and cropped into 4,830 patches sized $512\times 512$. The SDD dataset encompasses diverse indoor and outdoor scenes including 50 indoor scenes and 65 outdoor scenes in the training set, along with 11 indoor scenes and 24 outdoor scenes in the test set. Misalignment between blurry and sharp images occurs in two forms: zoom misalignment obtained by vertical camera movement and shift misalignment achieved through horizontal movement, referring to Fig.~\ref{intro}. 

\section{Experiments}\label{sec4}

\noindent \textbf{Datasets:}
We evaluate the performance of our proposed method on four datasets: DPDD \cite{abuolaim2020defocus}, RealDOF \cite{lee2021iterative}, DED \cite{ma2021defocus} and our SDD. 
\begin{itemize}
\item {DPDD} dataset was built by using a dual-pixel camera, capturing the defocus and all-in-focus pairs in two successive shots. It contains 350/76/74 image triplets for training/testing/validation, respectively. Each blurry image is paired with a corresponding sharp image. We use 7,000 processed blurry-sharp pairs for training and 76 blurry images for testing.
\item {RealDOF} dataset was constructed using a dual-camera system with a beam splitter, as described in \cite{lee2021iterative}. It provides only a test set, consisting of 50 scenes, for evaluation. 
\item {DED} dataset was the first large-scale realistic dataset for defocus map estimation and defocus image deblurring. It comprises a total of 1,112 image pairs, with some sourced from the multi-view dataset~\cite{ma2021defocus} and others captured using a light field camera.
\item {SDD} dataset contains 4,830 training pairs, and 35 image testing images with high resolution $4096\times 2160$.
\end{itemize}
\noindent \textbf{Evaluation Metrics:} We report five metrics to quantitatively assess the compared methods: Peak Signal-to-Noise Ratio (PSNR), Structural Similarity Index (SSIM)\cite{wang2004image}, Learned Perceptual Image Patch Similarity (LPIPS)\cite{zhang2018unreasonable}, Fréchet Inception Distance (FID)\cite{blau2018perception} and Deep Image Structure and Texture Similarity (DISTS)\cite{jinjin2020pipal}.
In defocusing deblurring scenarios with misaligned training data, existing methods may restore the image, but they often lead to severe distortions and truncations. Quantitative metrics, \emph{e.g}., PSNR and SSIM, are sensitive to pixel-level differences and may not fully capture the local structural improvements, such as edge sharpness and texture clarity, which are more perceptually significant.
This aligns with the findings in image restoration research~\cite{jinjin2020pipal,blau2018perception}, where perceptual quality is increasingly valued over strict pixel-level fidelity.
The study \cite{jinjin2020pipal} also analyzes the unsatisfactory performance of existing metrics on certain image restoration techniques, partially due to their low tolerance to spatial misalignment. Thus, in addition to PSNR and SSIM \cite{wang2004image}, we also calculated LPIPS \cite{zhang2018unreasonable}, FID \cite{blau2018perception}, and DISTS \cite{jinjin2020pipal}.

\subsection{Implementation Details}

At the beginning of training, the predicted optical flow $\bm{\Phi}_{\bm{I}_{S}\rightarrow\hat{\bm{I}}}$ and $\bm{\Phi}_{\hat{\bm{I}}\rightarrow \bm{I}_{S}}$ are hardly informative due to the low quality of $\hat{\bm{I}}$. Therefore, during the initial training stage, we calculate $\bm{\Phi}_{\bm{I}_{S}\rightarrow\hat{\bm{I}}}$ using $\mathcal{F}_{flow} (\bm{I}_S, \bm{I}_B)$ over $T$ training epochs, and subsequently by Eq.~\eqref{eq:flow}. To ensure the quality of $\hat{\bm{I}}$, we empirically set $T$ as 15 in our experiments.  
During training, we set $\lambda$ as 0.35 to generate the calibration masks. The maximal radius of blur kernels $m$ is set as 8, and the sampling points number $N_p$ in deformable attention is set as 4. The trade-off parameters $\alpha$ and $\beta$ in Eq. \eqref{eq:jdrl} are both set as 0.5. 

All the experiments are conducted using PyTorch on two A100 GPUs. 
The input images, along with the corresponding sharp ground truths and defocus maps, are randomly cropped to a size of $512\times512$. The batch size is set to 1.
The parameters are initialized using the strategy proposed by He et al.~\cite{he2015delving}, and are optimized using the Adam optimizer~\cite{kingma2014adam} by setting $\beta_{1}=0.9, \beta_{2}=0.999$ and $\epsilon=10^{-8}$. 
The learning rate is initialized as 2$\times$10$^{-5}$ and is halved every 60 epochs. 
The entire training stage ends with 200 epochs.

\subsection{Evaluation on SDD Dataset}

In this section, we compare our proposed defocus deblurring method against state-of-the-art single-image defocus deblurring approaches.
we retrain state-of-the-art deblurring methods on the SDD dataset for further comparison, including DMPHN ~\cite{zamir2021multi}, MPRNet~\cite{zhang2019deep}, UformerT~\cite{wang2022uformer}, Restormer~\cite{zamir2022restormer} and Loformer \cite{mao2024loformer}. 
To validate the effectiveness of our method, We have applied the proposed framework to both CNN-based and Restormer-based deblurring methods. 
As shown in Table \ref{tab:SDD more metrics}, our proposed framework is evaluated from two perspectives: (\emph{i}) reblurring-based learning framework, \emph{i.e}., deblurring model+Ours(Eq. (1)), and (\emph{ii}) reblurring-based blurring framework and deblurring blocks $\mathcal{E}$ and $\mathcal{F}$, \emph{i.e}., deblurring model+Ours(Eq. (2)).
One can see that for plain UNet, PSNR gains +1.2dB and +1.58dB are obtained by Ours(Eq. (1)) and Ours(Eq. (2)), respectively.
For Loformer with top performance, PSNR gains +0.44dB and +0.68dB are obtained by Ours(Eq. (1)) and Ours(Eq. (2)), respectively.
Considering that UNet, Restormer and Loformer are representative network architectures, covering from plain CNN to top Transformer, we believe that our proposed framework can be successfully applied to improve existing deblurring models. 

% In this section, we compare our proposed defocus deblurring method against state-of-the-art single-image defocus deblurring approaches, including JNB~\cite{Shi_2015_CVPR}, EBDB~\cite{karaali2017edge}, DMENet~\cite{Lee_2019_CVPR} and DPDNet$_{S}$~\cite{abuolaim2020defocus}. 
% %
% Notably, the first three methods are primarily developed for defocus map estimation, we follow \cite{krishnan2009fast,fish1995blind} to get the final estimated sharp images using non-blind deconvolution. 
%
% Additionally, we retrain some state-of-the-art deblurring methods on the SDD dataset for further comparison, including DMPHN ~\cite{zamir2021multi}, MPRNet~\cite{zhang2019deep}, Restormer\cite{zamir2022restormer}, Loformer \cite{mao2024loformer} and GRL$_{S}$-B \cite{li2023efficient}. 
% %
% It is noteworthy that this study focuses on single-image defocus deblurring using misaligned data pairs.  GRL \cite{li2023efficient}, for instance, has a dual-pixel version, but only its single image version (GRL$_{S}$-B \cite{li2023efficient}) is employed in our experiments.
%
Since our dataset is designed for tackling misalignment tasks, directly calculating pixel-level metrics may not be accurate. 
Therefore, when testing all methods, we used an existing pretrained optical flow network to warp the sharp images $\bm{I}_{S}$, aligning them with the deblurred images $\hat{\bm{I}}$, and then calculated the evaluation metrics between $\hat{\bm{I}}$ and $\bm{I}_S^w$.

Table \ref{tab:SDD more metrics} summarizes the results of all competing methods on SDD dataset.
In our conference paper \cite{li2023learning}, our method directly adopt MPRNet as the deblurring model, and achieves notable performance gains over MPRNet. 
In this work, our reblurring-guided learning frameworks Eq. \eqref{eq:jdrl-simple} and Eq. \eqref{eq:jdrl} both obtain better deblurring performance, while Ours (Eq. \eqref{eq:jdrl}) further improves the quantitative metrics benefiting from the defocus map as degradation prior in the baseline deblurring network. 
Nevertheless, all the models trained by reblurring-guided framework can well handle the misalignment issues in training data. 
%
% In the following, the term ``Ours" refers to our proposed method described by Eq. \eqref{eq:jdrl}, unless otherwise specified for clarity. 

{In Table \ref{tab:flops}, 
we provide comparison of parameters and FLOPs for several representative methods, including UFormerT~\cite{wang2022uformer}, MPRNet~\cite{zhang2019deep}, Loformer~\cite{mao2024loformer} and Restormer~\cite{zamir2022restormer}.
Since Ours(Eq. (1)) does not modify the architecture of deblurring model, parameters and FLOPs of the methods with * are exactly same with their original models. 
For the methods with ${\dagger}$, the blocks $\mathcal{E}$ and $\mathcal{F}$ introduce additional computational cost, but the increases are very slight, where only $\sim$0.1M parameters and $\sim$18G FLOPs are negligible in comparison to these deblurring models. 
}

\begin{figure*}[t]
					\small
					\centering
					\setlength{\abovecaptionskip}{5pt} 
					\setlength{\belowcaptionskip}{0pt}
					\begin{tabular}{cc}
						\footnotesize
						\begin{adjustbox}{valign=t}
							\begin{tabular}{cccccccc}
								\hspace{-4mm}
								\includegraphics[width=0.12\textwidth]{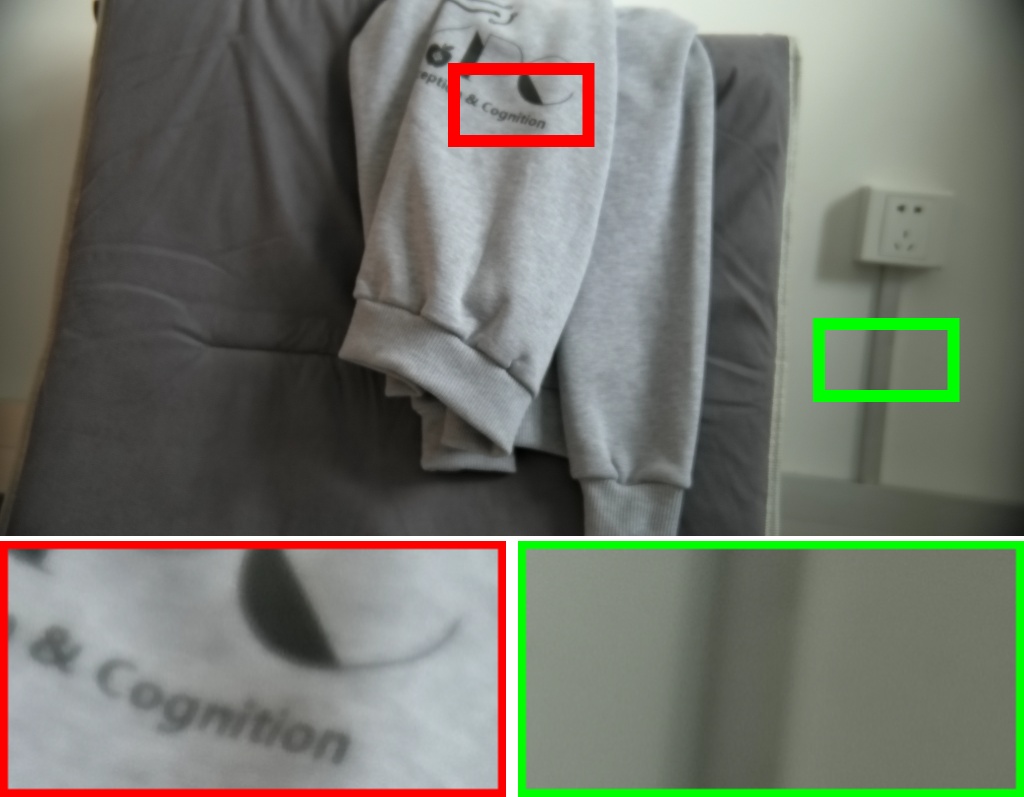}&\hspace{-4mm}
								\includegraphics[width=0.12\textwidth]{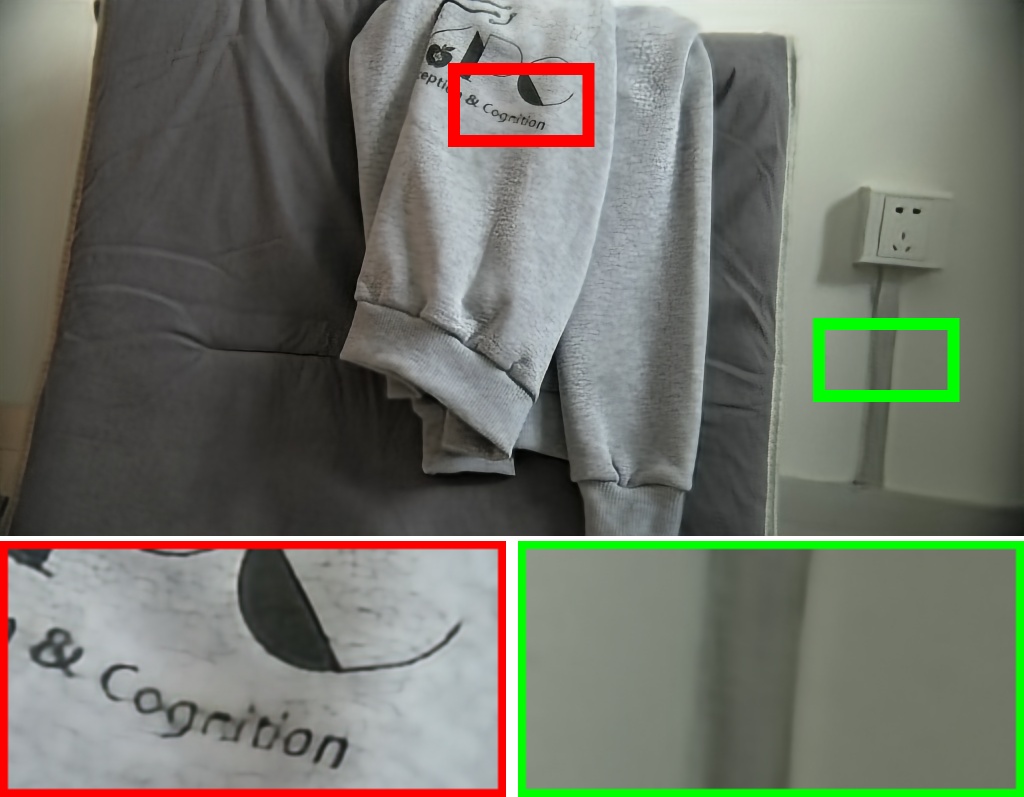}&\hspace{-4mm}
								\includegraphics[width=0.12\textwidth]{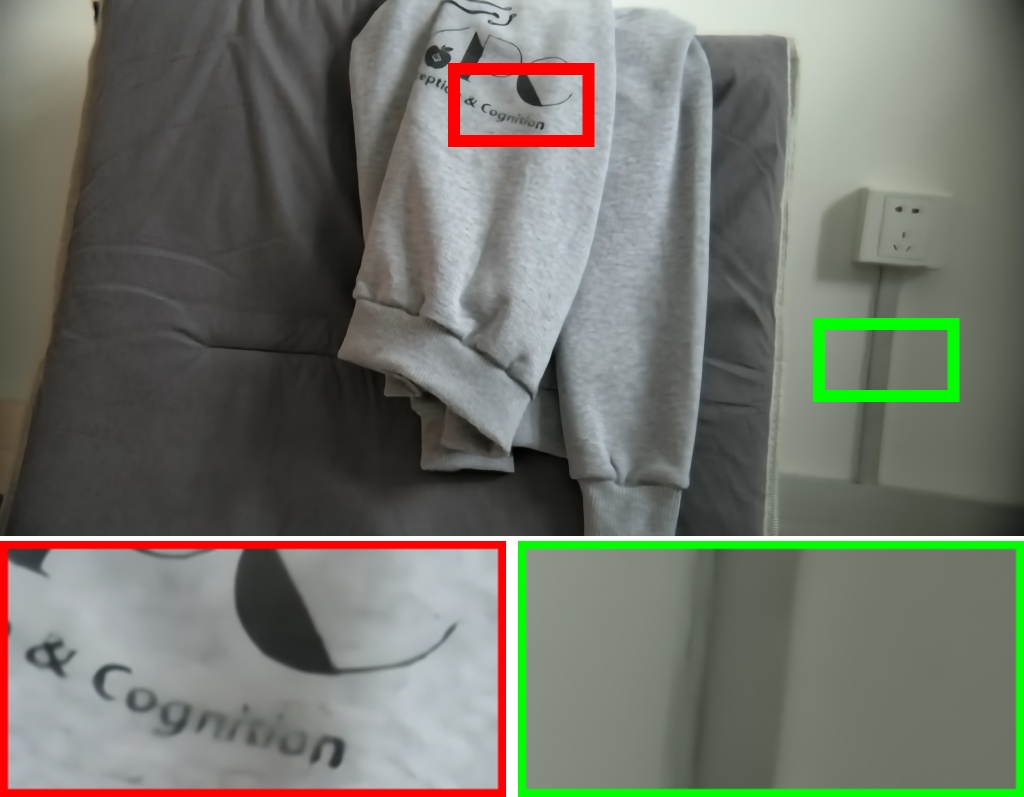}&\hspace{-4mm}
								\includegraphics[width=0.12\textwidth]{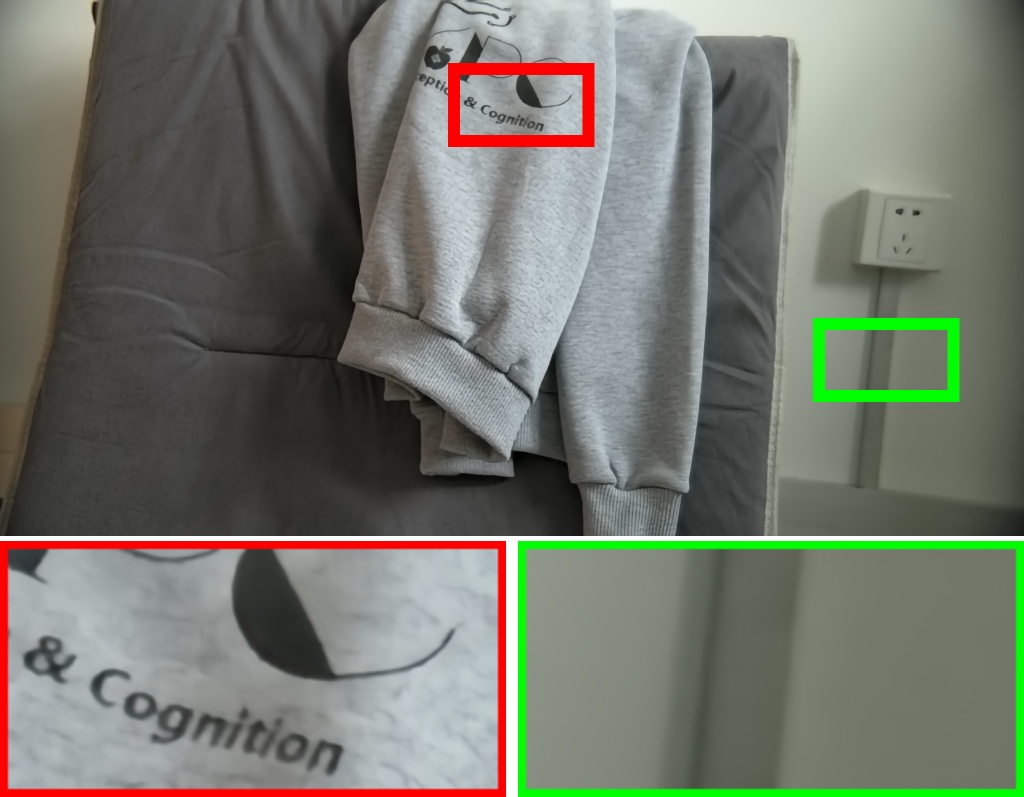}&\hspace{-4mm}
								\includegraphics[width=0.12\textwidth]{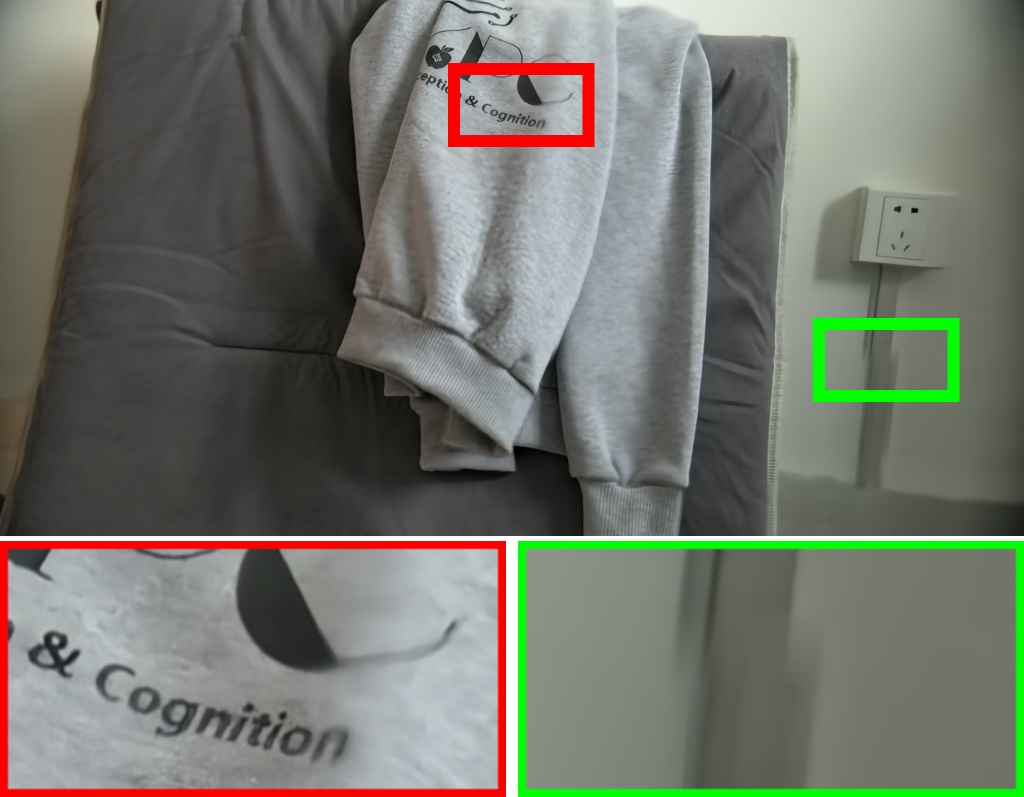}&\hspace{-4mm}
								\includegraphics[width=0.12\textwidth]{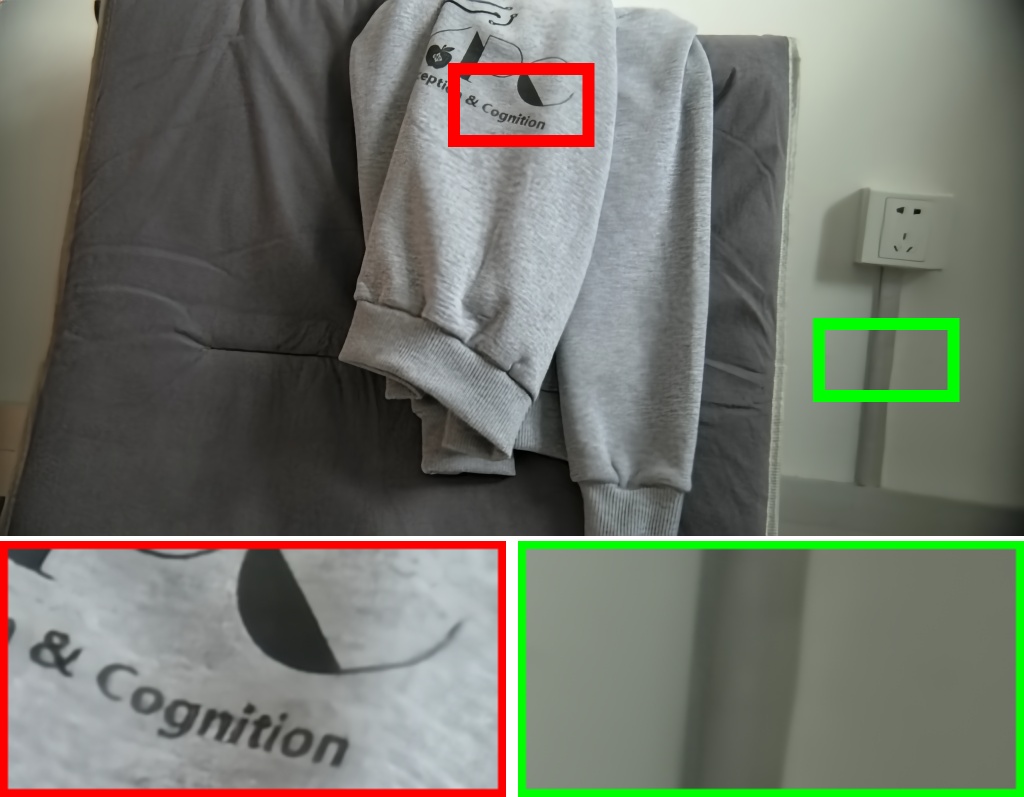}&\hspace{-4mm}
								\includegraphics[width=0.12\textwidth]{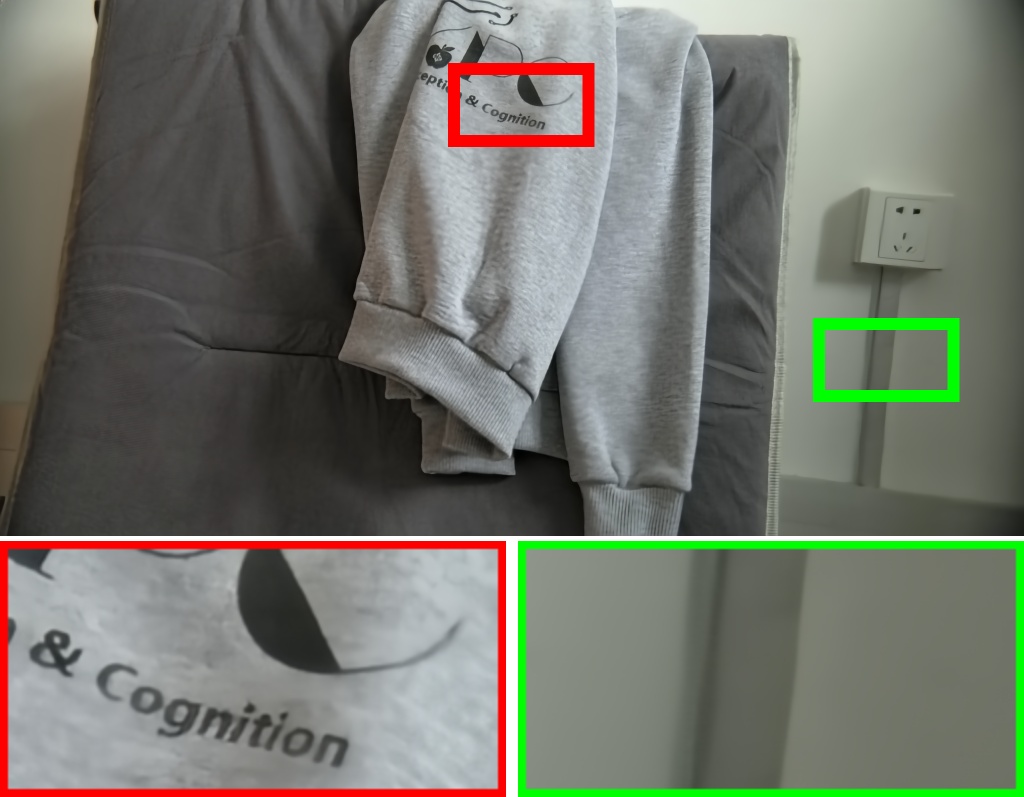}&\hspace{-4mm}
								\includegraphics[width=0.12\textwidth]{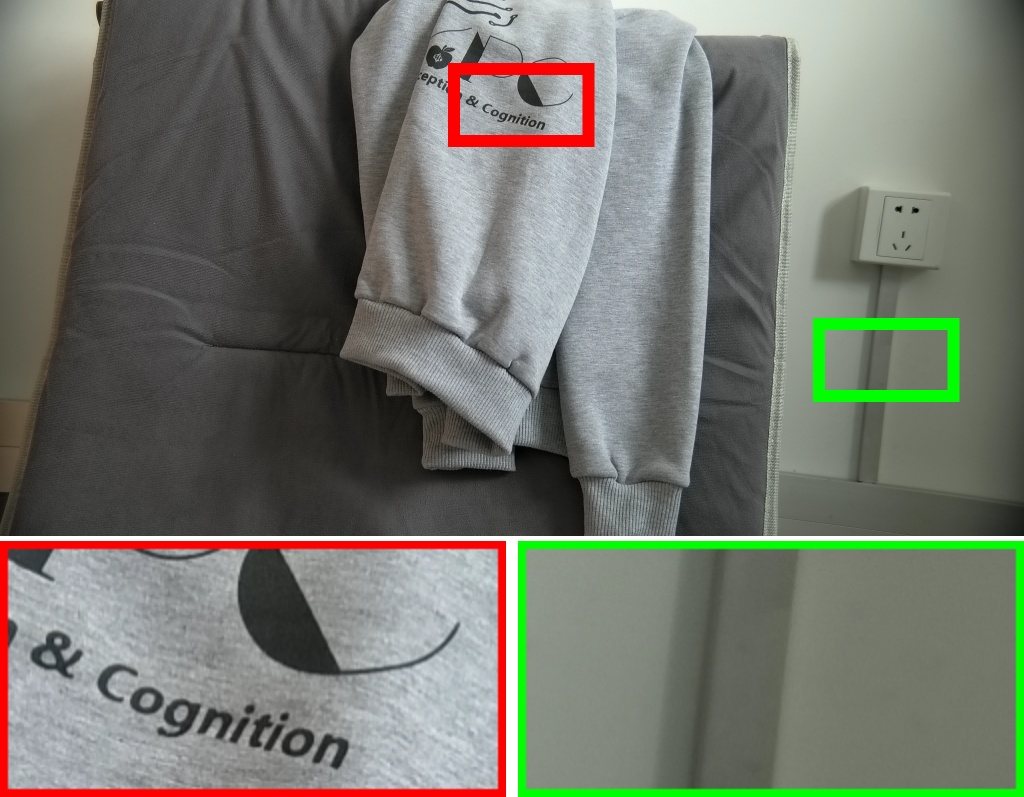}
								\\
								\hspace{-4mm}
								\includegraphics[width=0.12\textwidth]{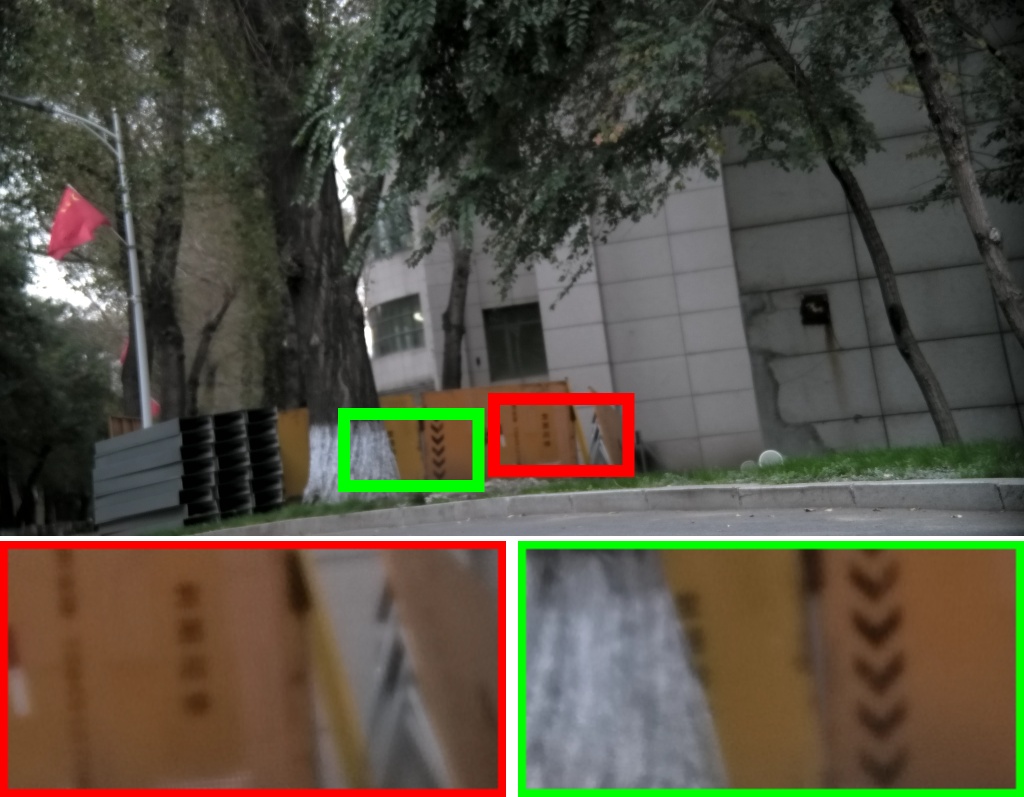}&\hspace{-4mm}
								\includegraphics[width=0.12\textwidth]{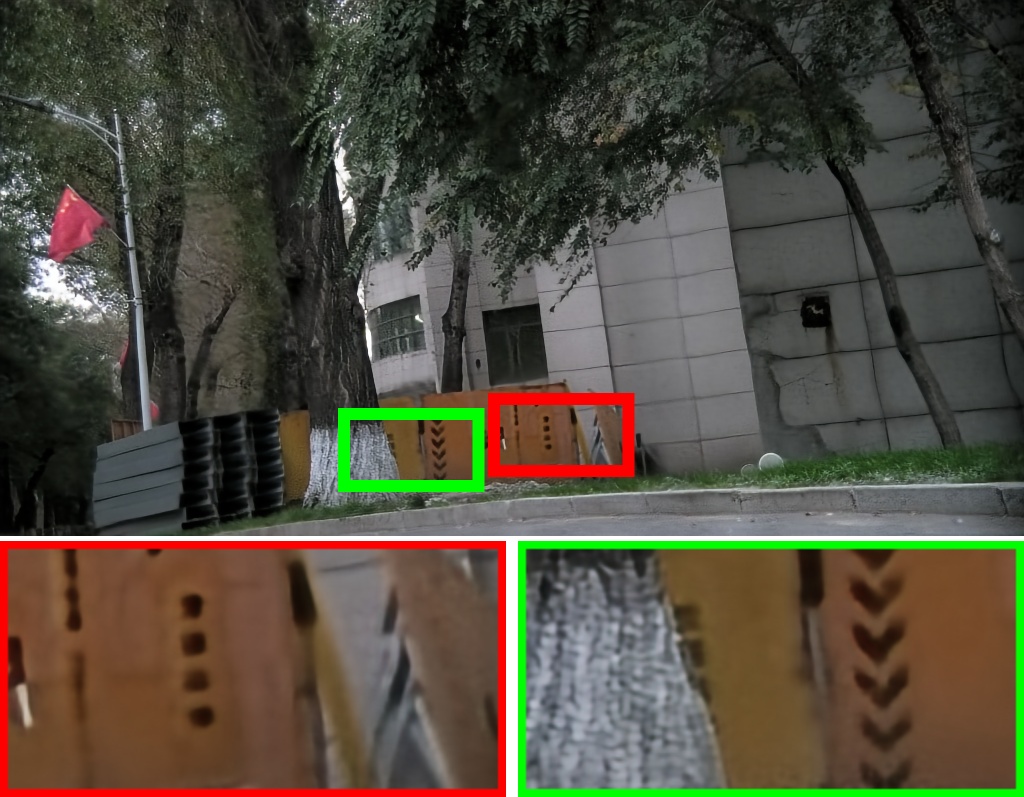}&\hspace{-4mm}
								\includegraphics[width=0.12\textwidth]{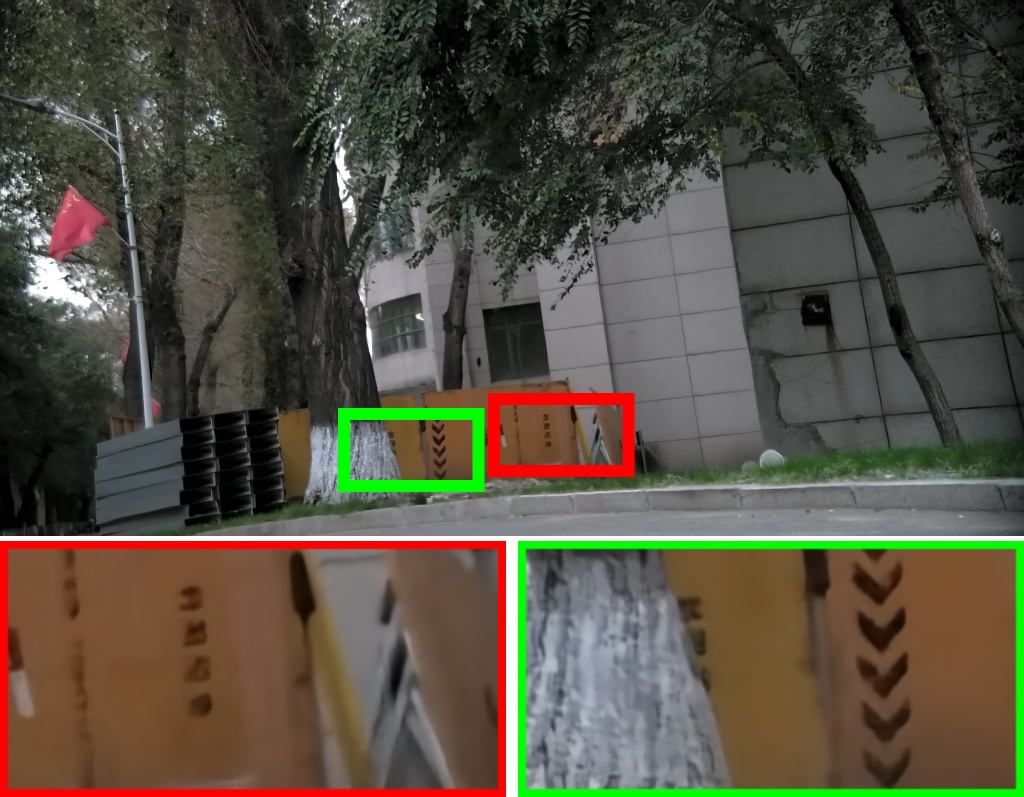}&\hspace{-4mm}
								\includegraphics[width=0.12\textwidth]{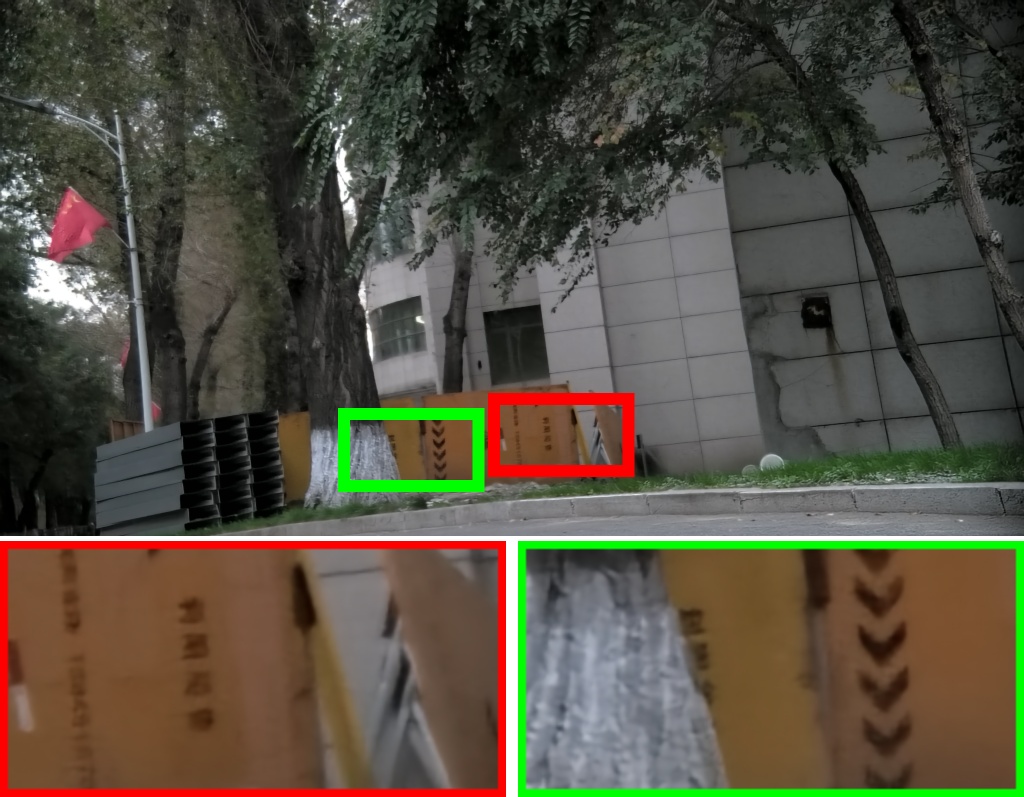}&\hspace{-4mm}
								\includegraphics[width=0.12\textwidth]{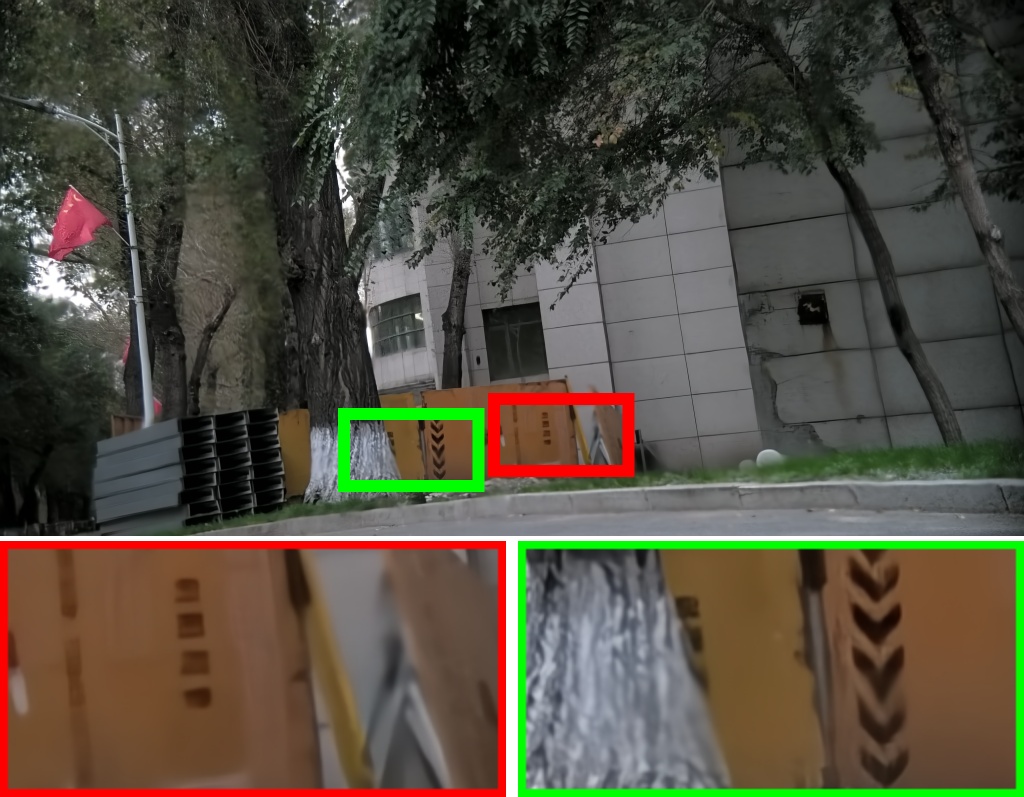}&\hspace{-4mm}
								\includegraphics[width=0.12\textwidth]{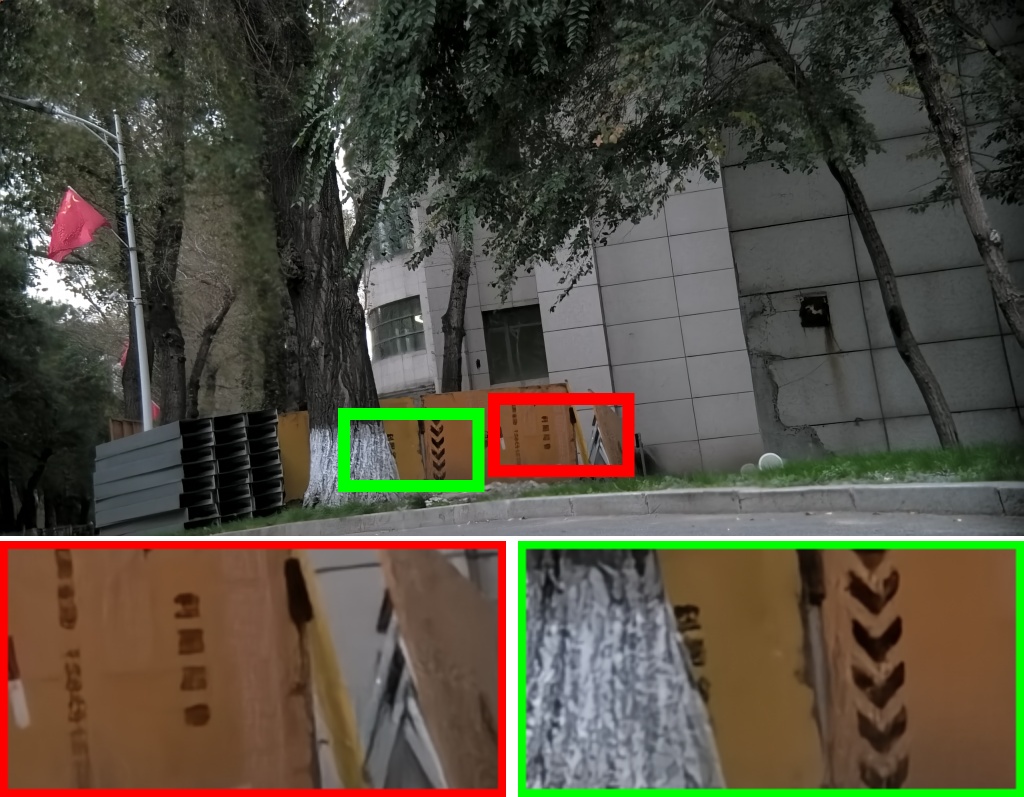}&\hspace{-4mm}
								\includegraphics[width=0.12\textwidth]
								{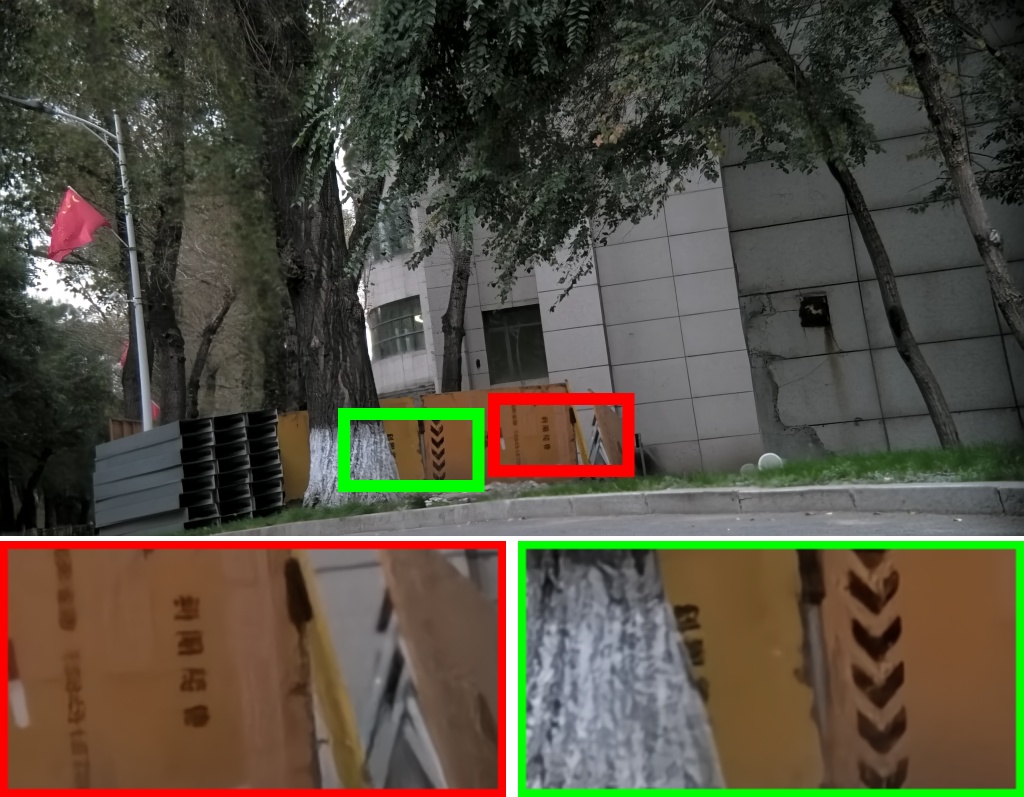}&\hspace{-4mm}
								\includegraphics[width=0.12\textwidth]
								{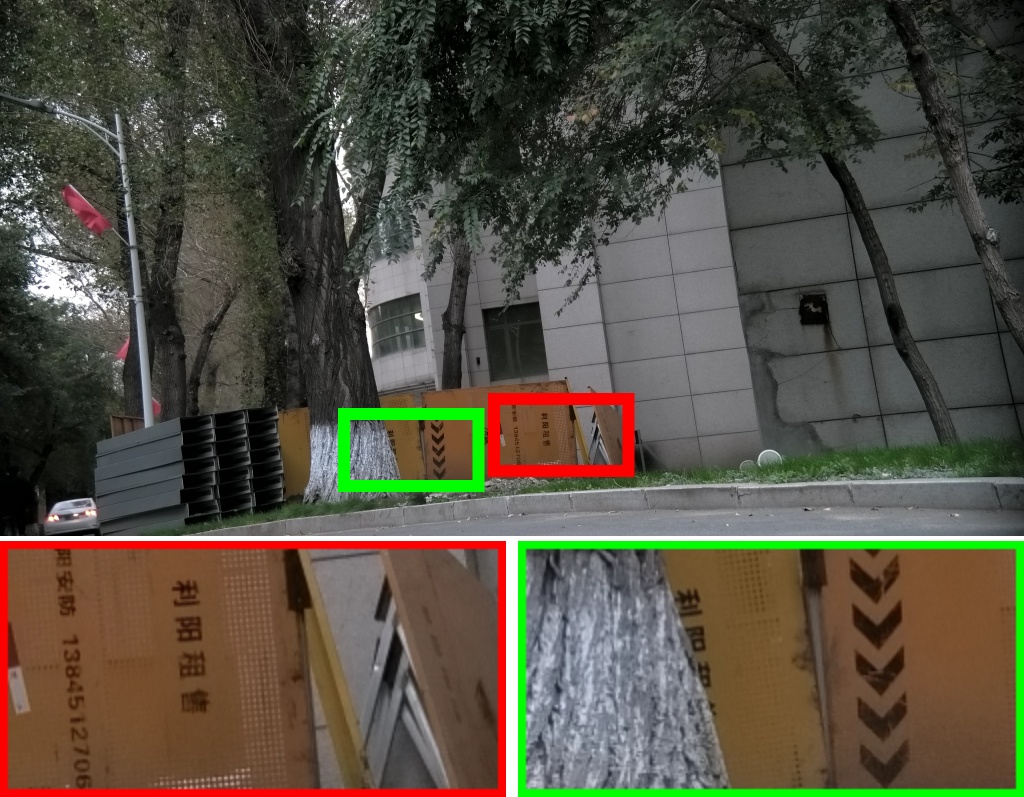}
								\\
								\hspace{-4mm}
								\includegraphics[width=0.12\textwidth]{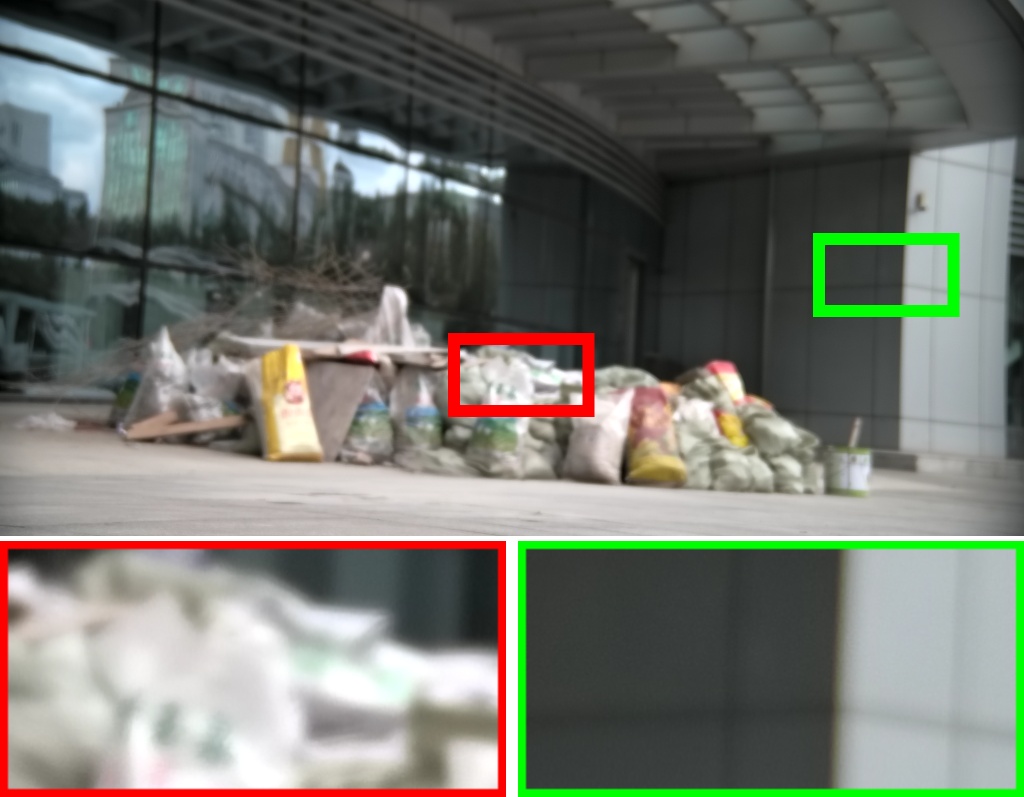}&\hspace{-4mm}
								\includegraphics[width=0.12\textwidth]{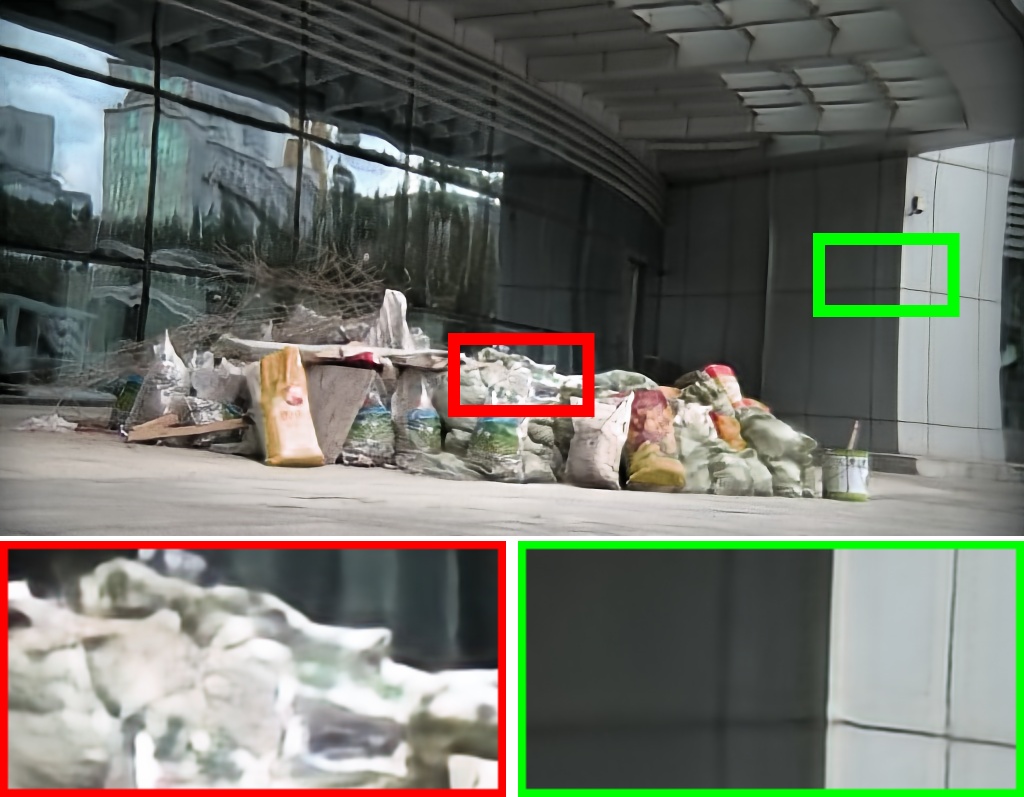}&\hspace{-4mm}
								\includegraphics[width=0.12\textwidth]{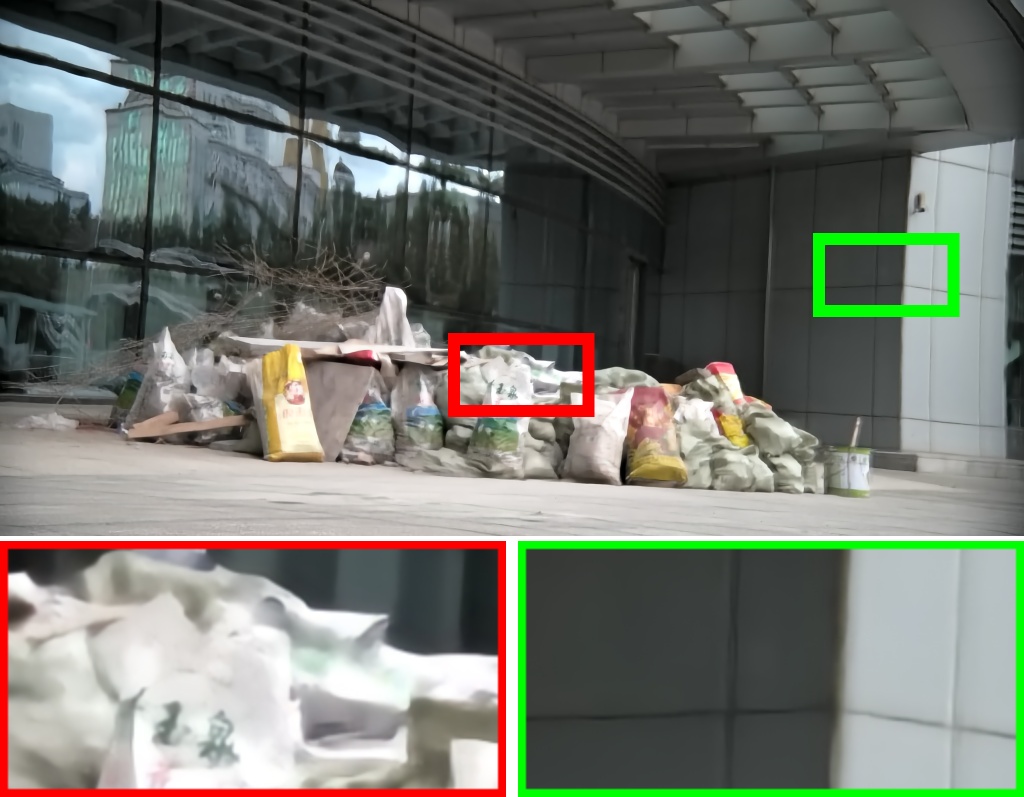}&\hspace{-4mm}
								\includegraphics[width=0.12\textwidth]{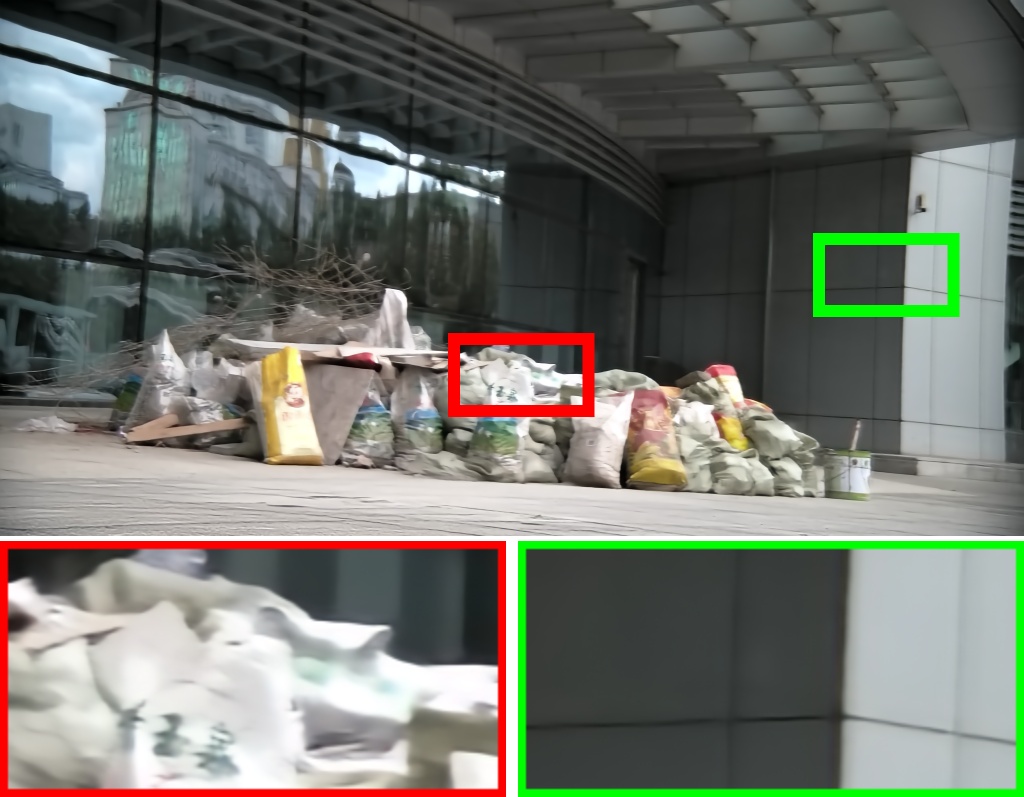}&\hspace{-4mm}
								\includegraphics[width=0.12\textwidth]{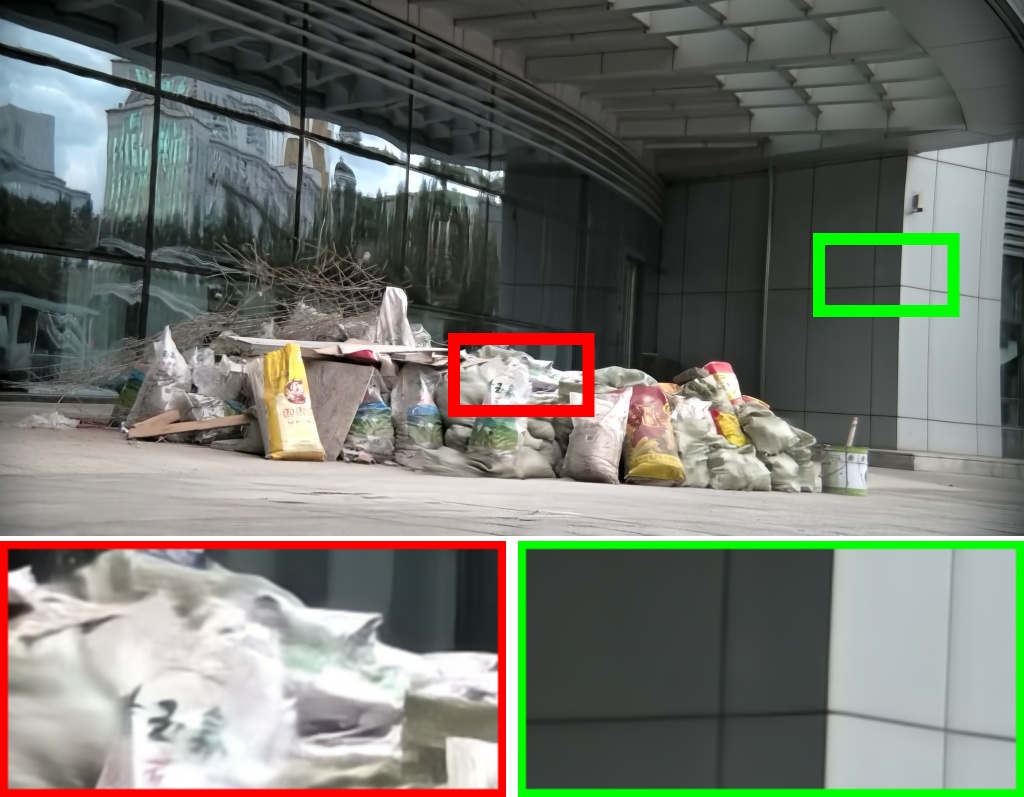}&\hspace{-4mm}
								\includegraphics[width=0.12\textwidth]{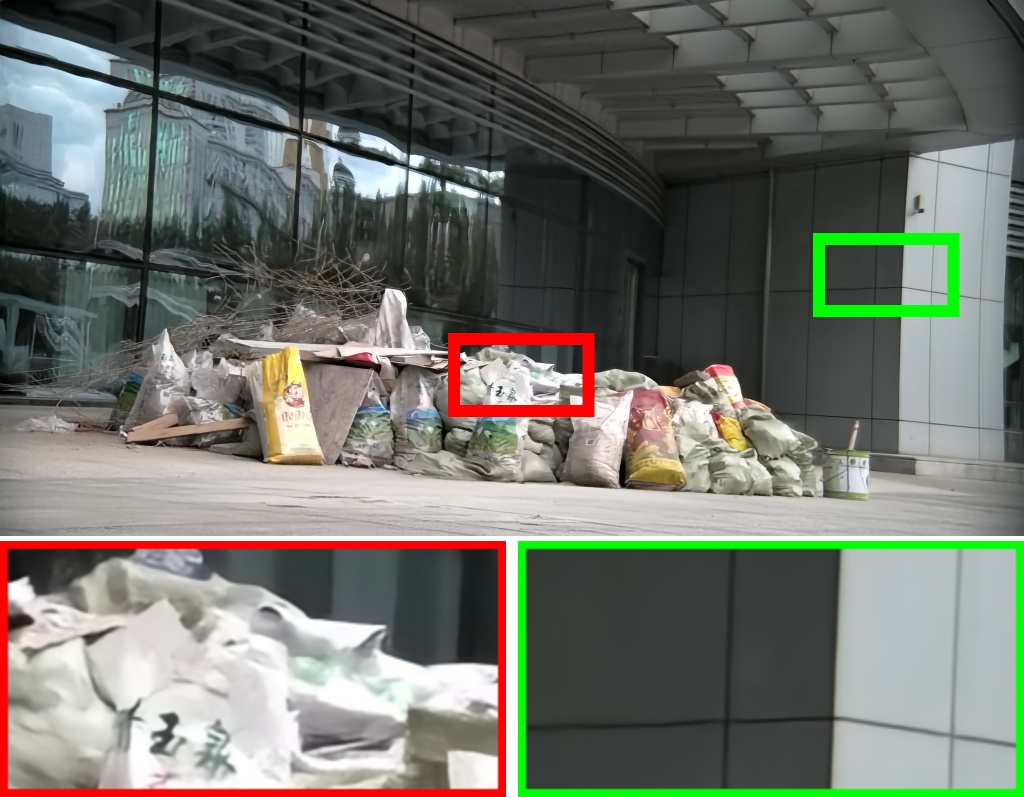}&\hspace{-4mm}
								\includegraphics[width=0.12\textwidth]{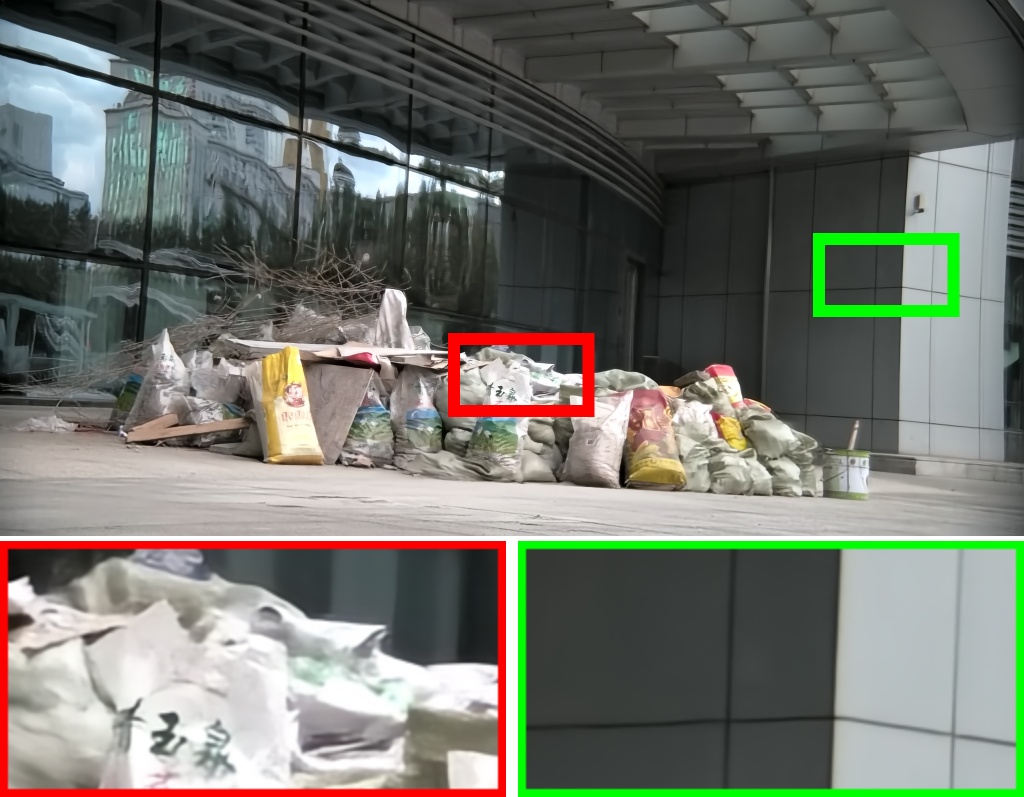}&\hspace{-4mm}
								\includegraphics[width=0.12\textwidth]{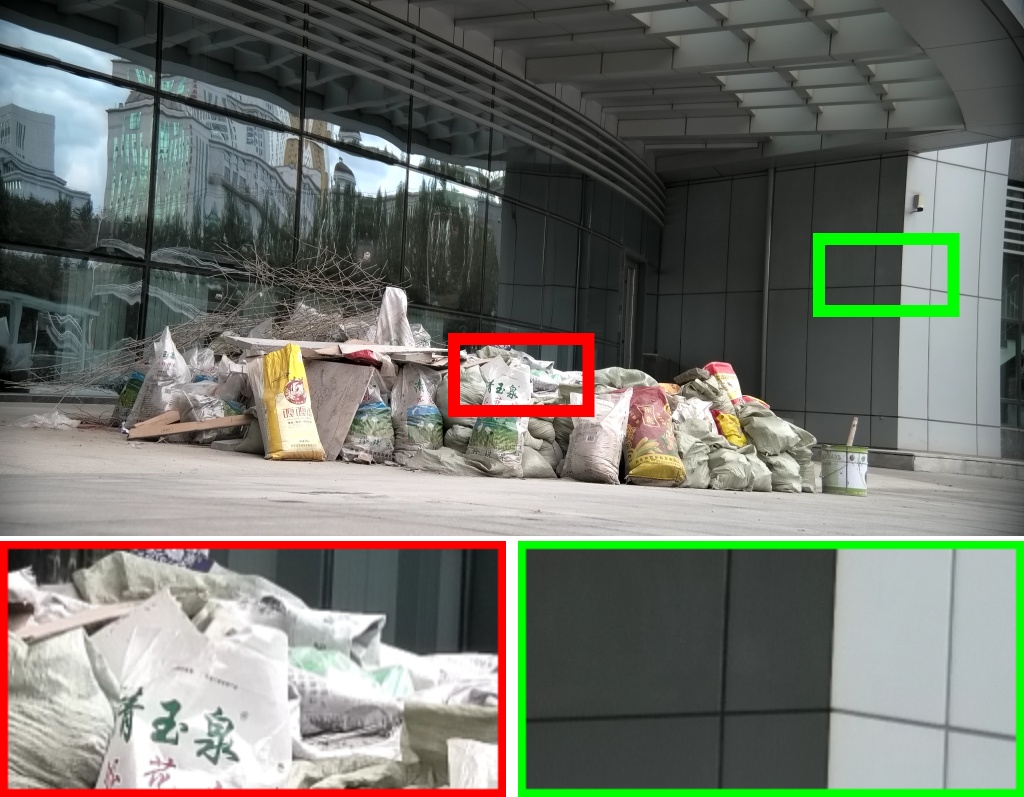}
								\\
								\hspace{-4mm}
								\includegraphics[width=0.12\textwidth]{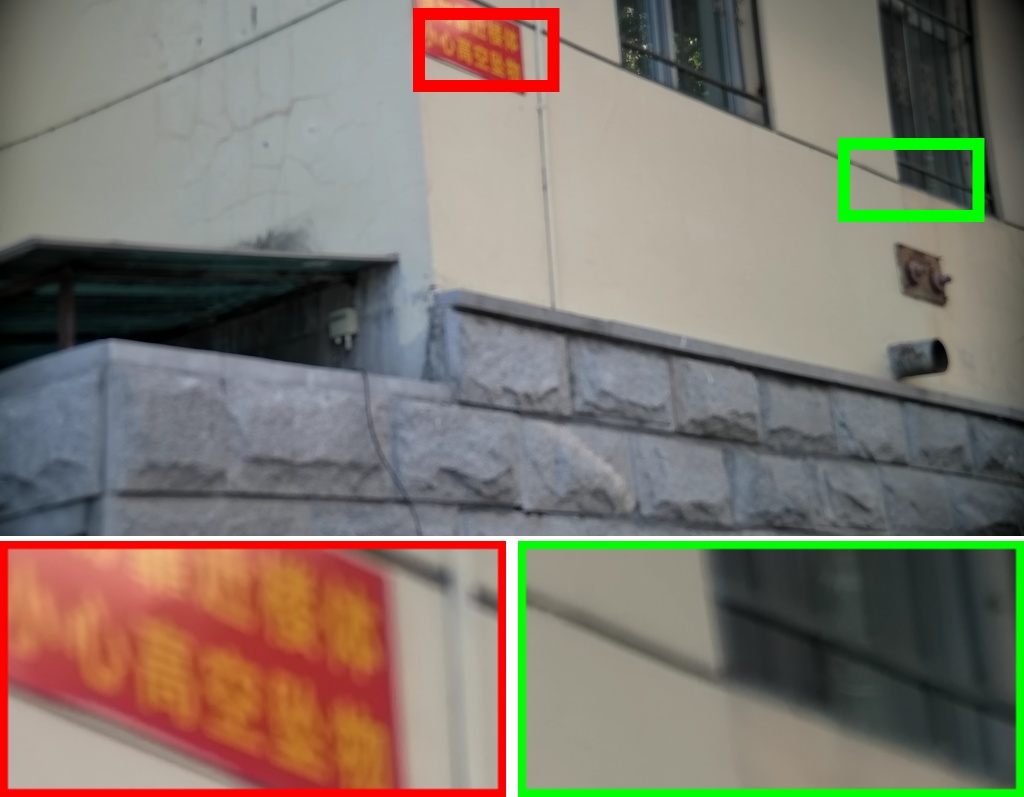}&\hspace{-4mm}
								\includegraphics[width=0.12\textwidth]{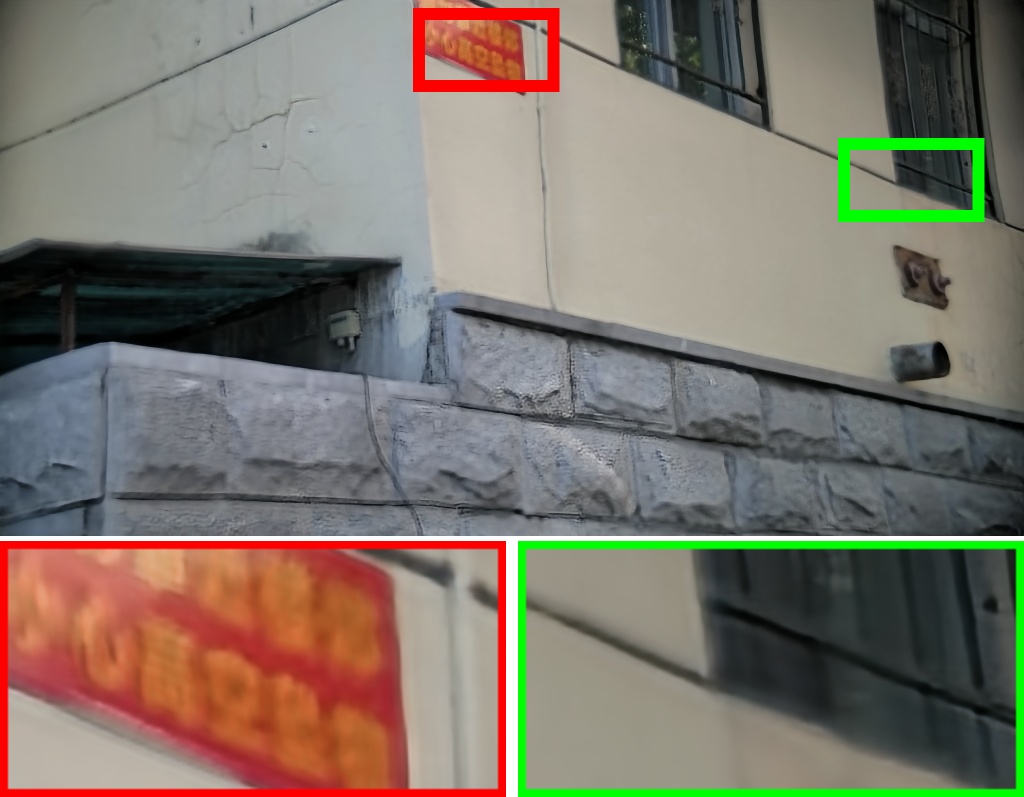}&\hspace{-4mm}
								\includegraphics[width=0.12\textwidth]{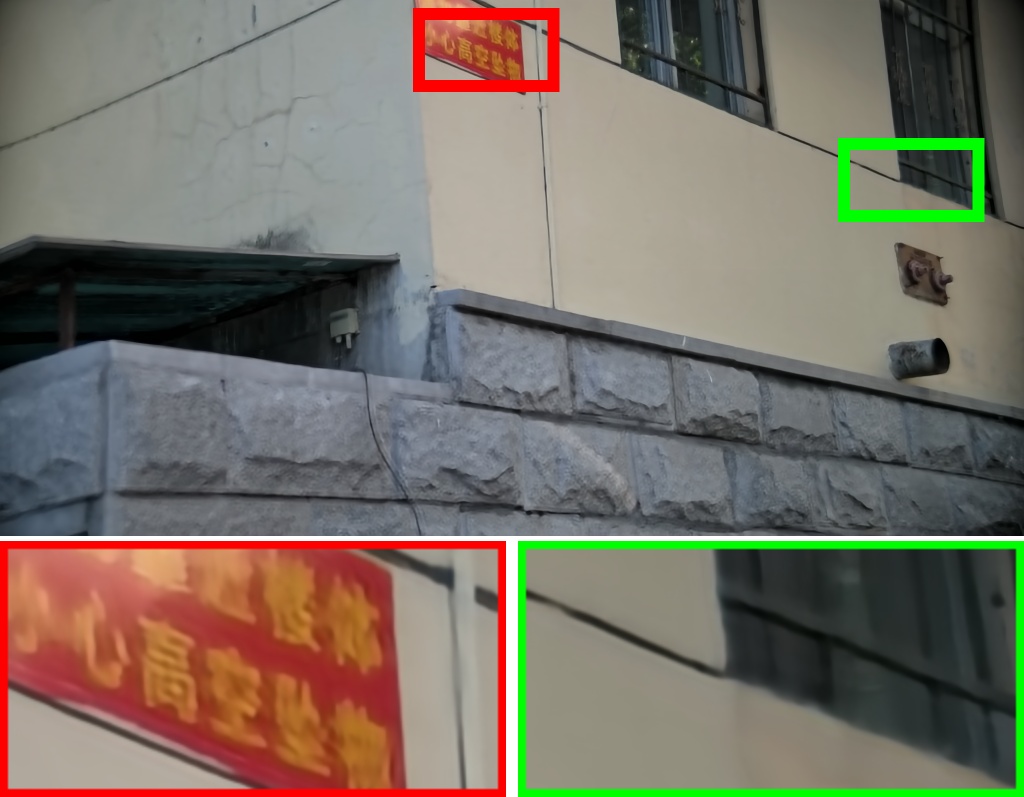}&\hspace{-4mm}
								\includegraphics[width=0.12\textwidth]{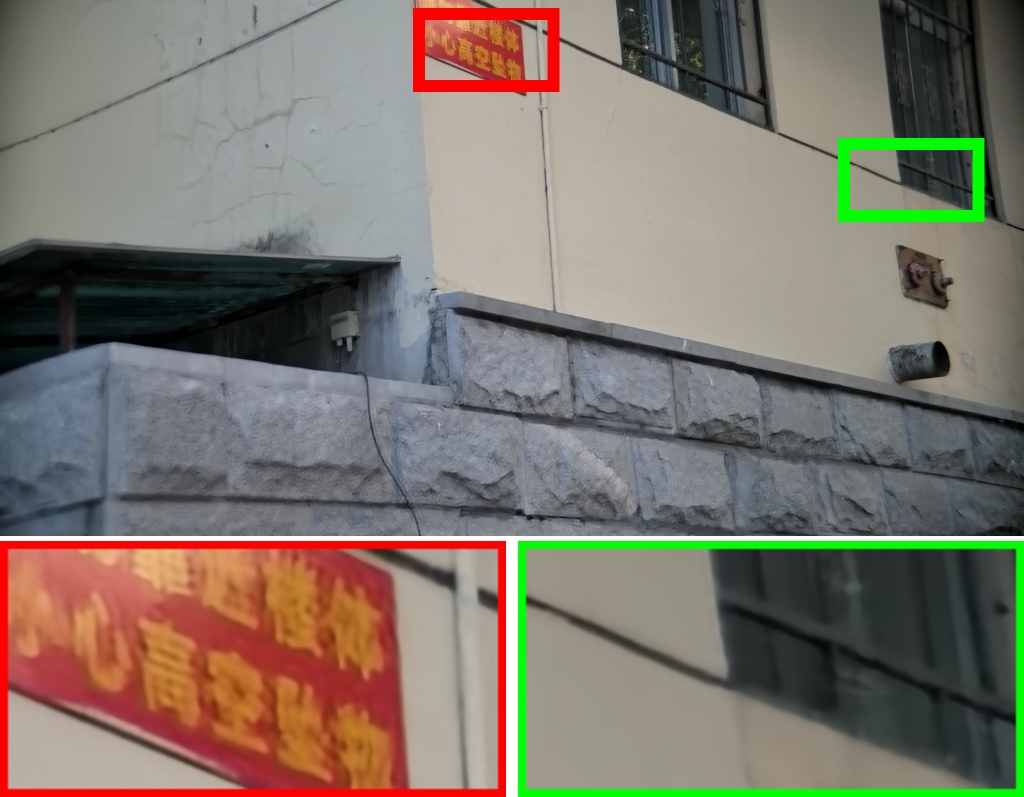}&\hspace{-4mm}
								\includegraphics[width=0.12\textwidth]{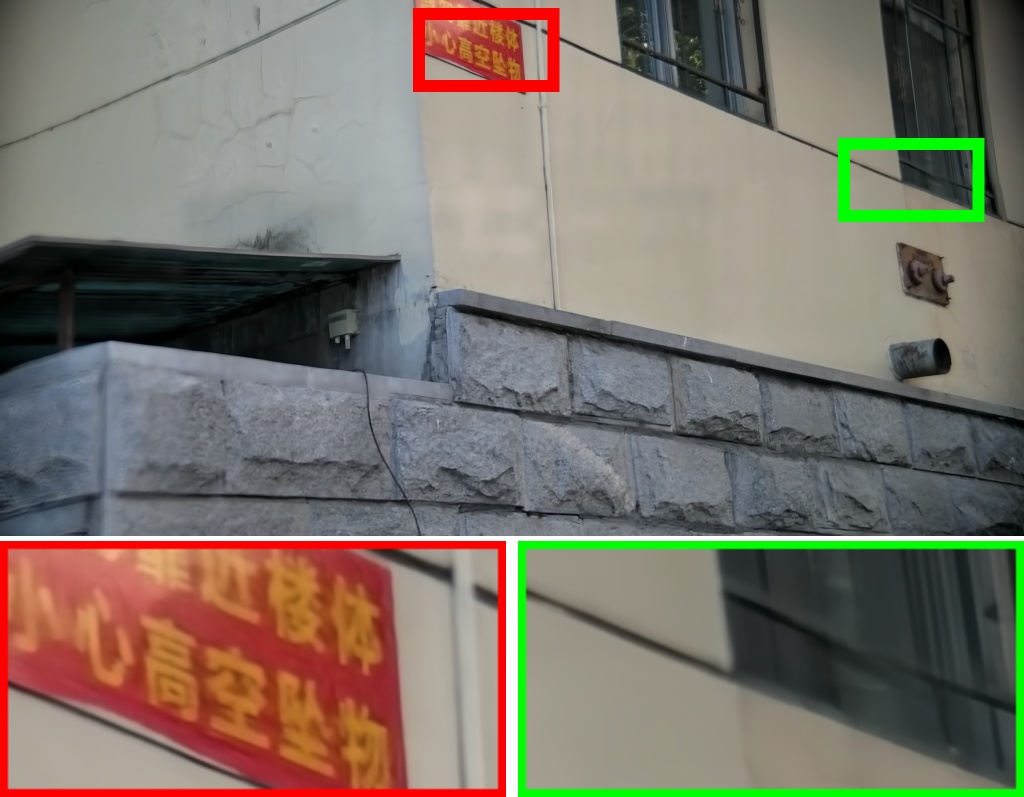}&\hspace{-4mm}
								\includegraphics[width=0.12\textwidth]{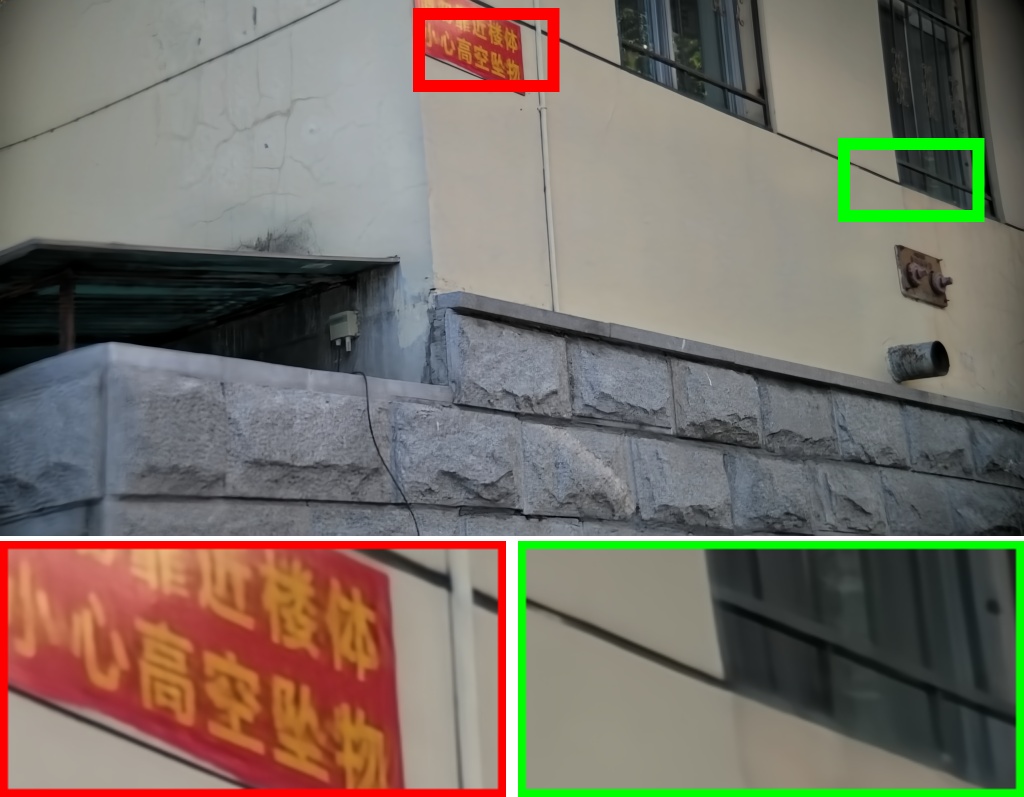}&\hspace{-4mm}
								\includegraphics[width=0.12\textwidth]{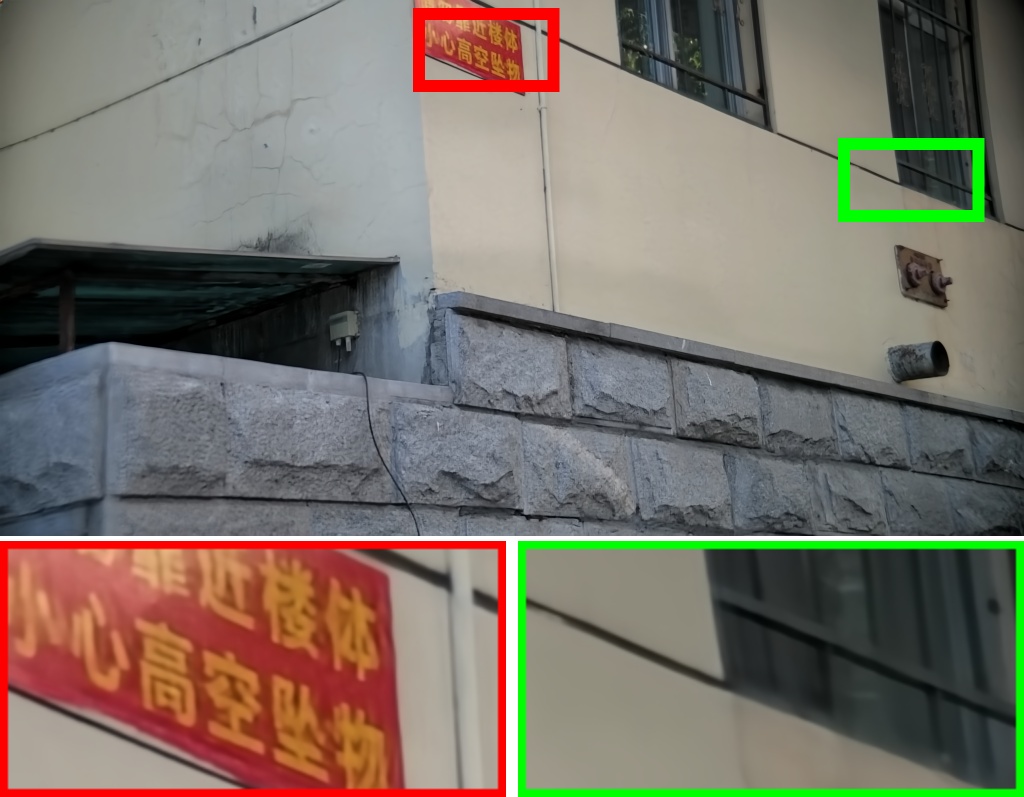}&\hspace{-4mm}
								\includegraphics[width=0.12\textwidth]{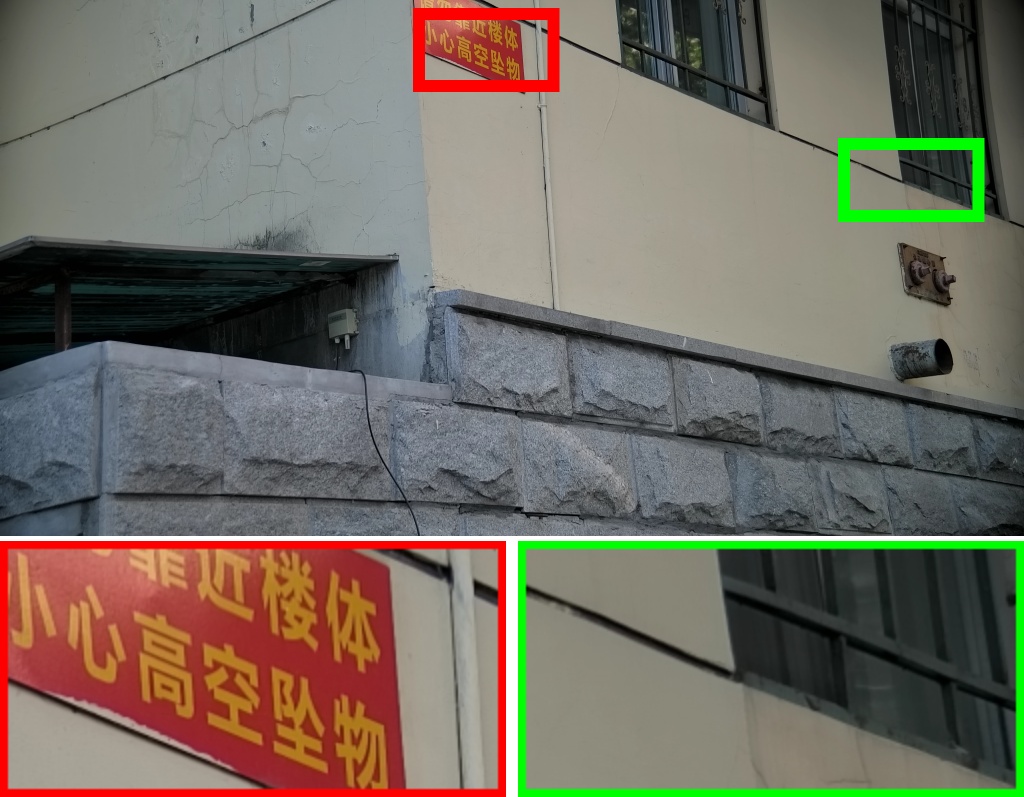}
								\\
								\hspace{-4mm}
								\includegraphics[width=0.12\textwidth]{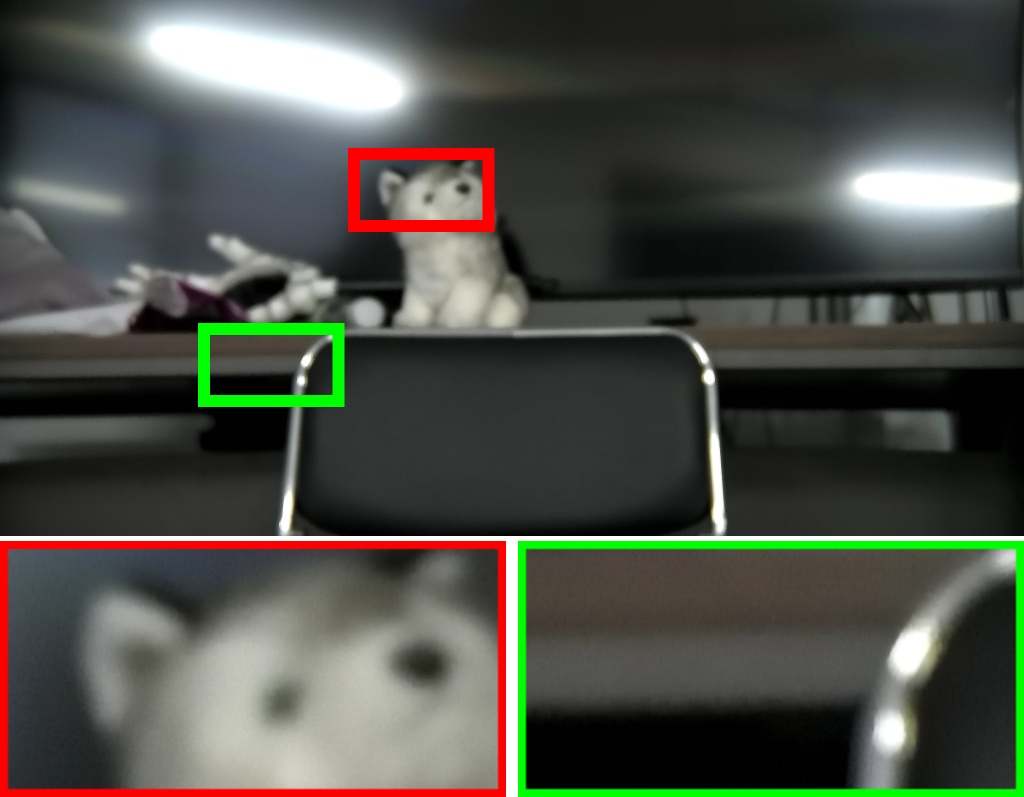}&\hspace{-4mm}
								\includegraphics[width=0.12\textwidth]{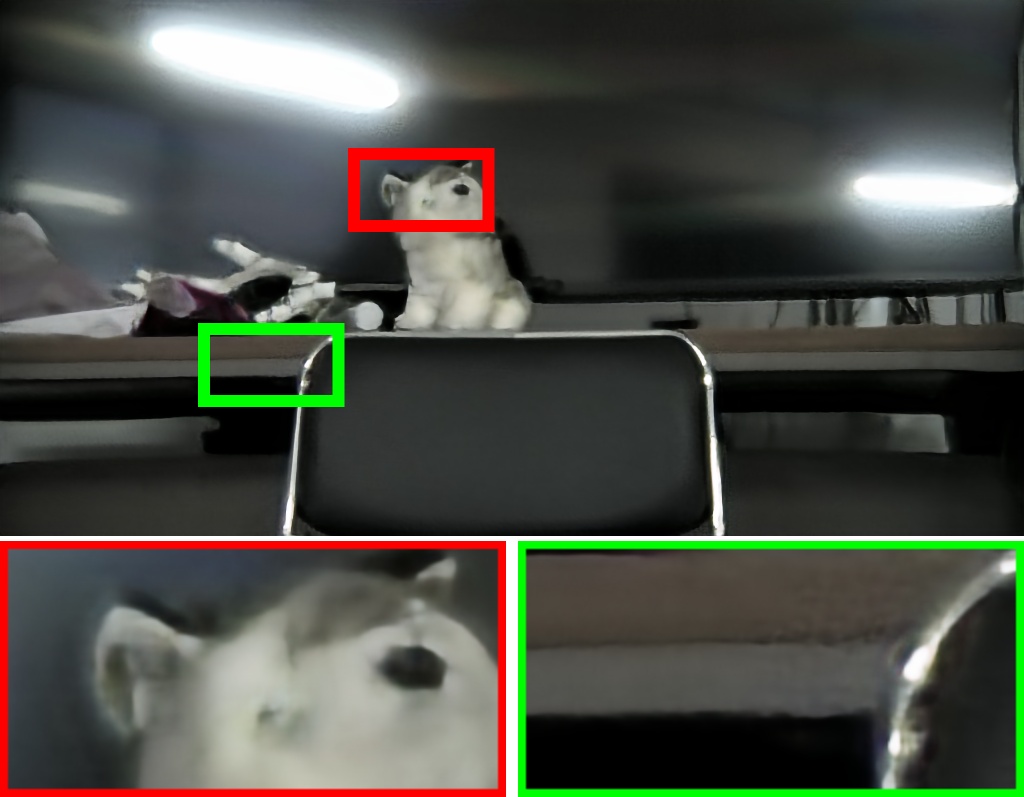}&\hspace{-4mm}
								\includegraphics[width=0.12\textwidth]{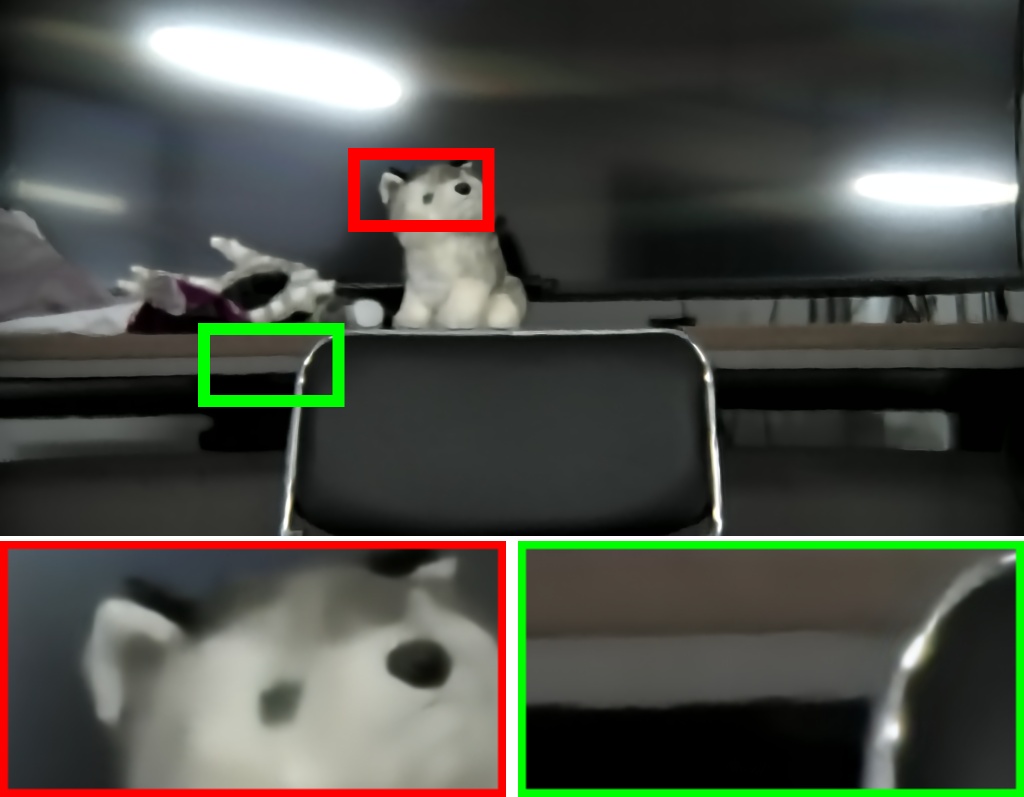}&\hspace{-4mm}
								\includegraphics[width=0.12\textwidth]{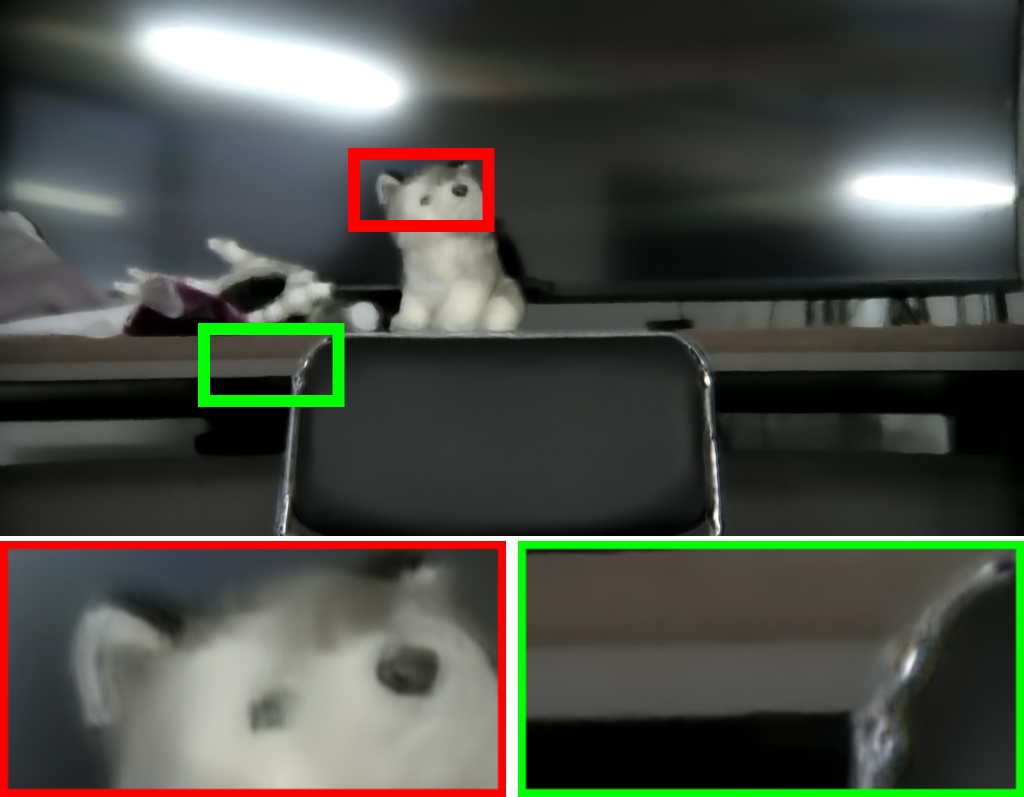}&\hspace{-4mm}
								\includegraphics[width=0.12\textwidth]{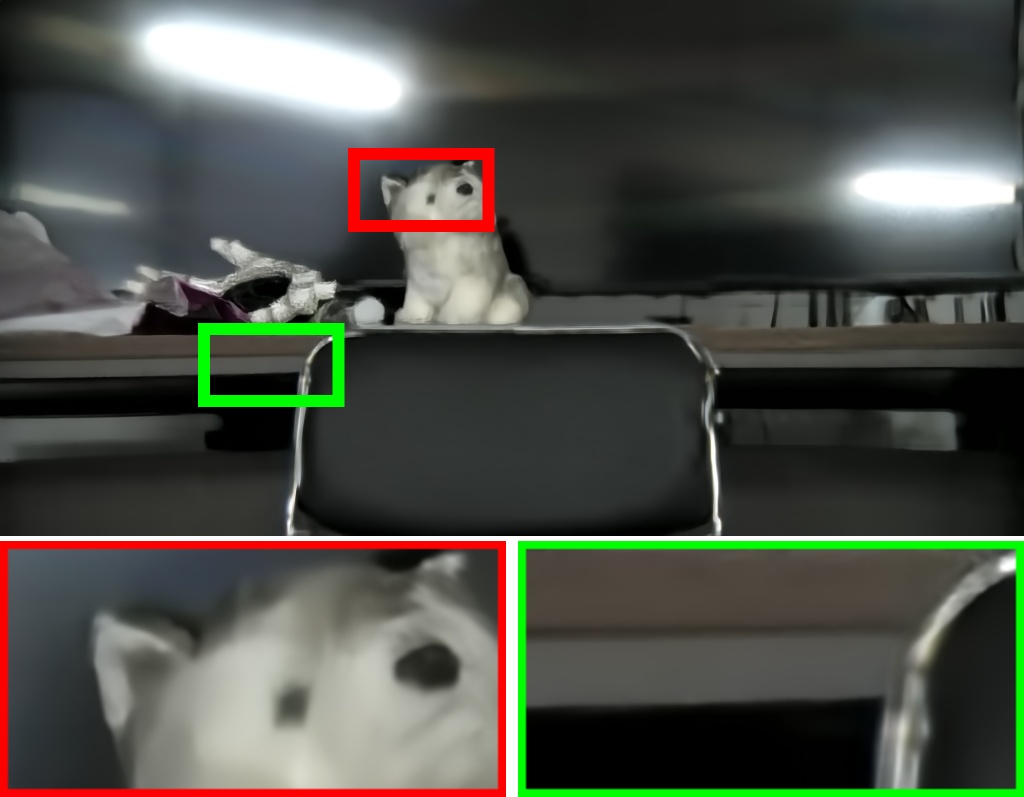}&\hspace{-4mm}
								\includegraphics[width=0.12\textwidth]{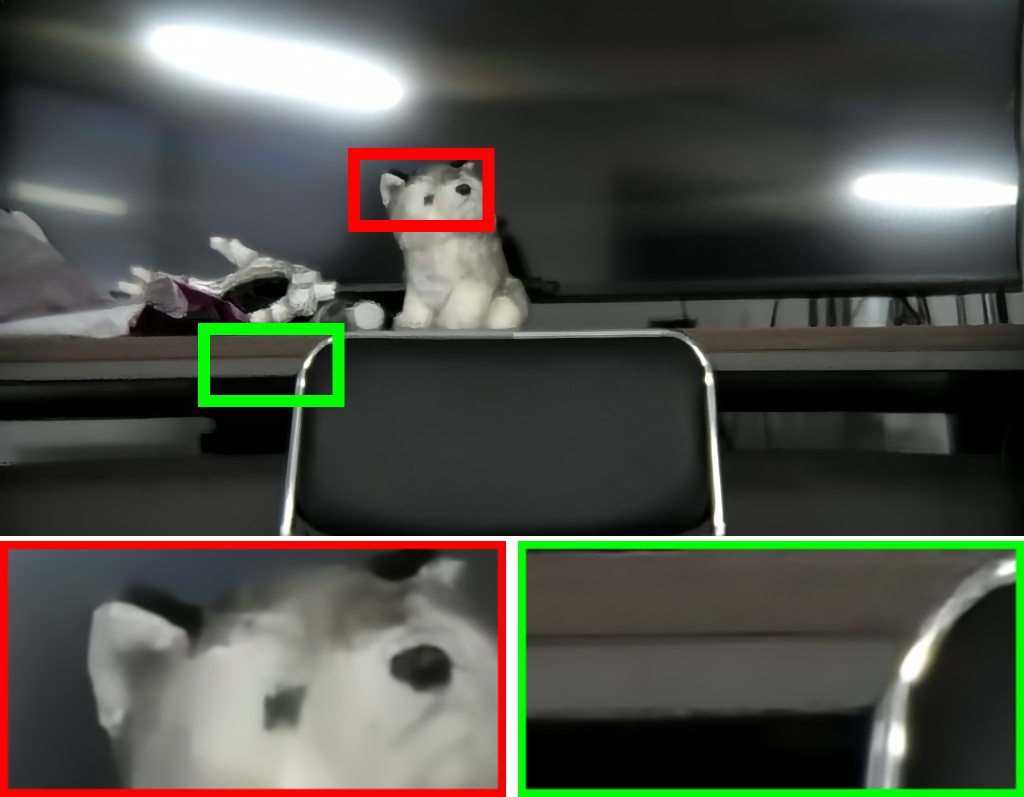}&\hspace{-4mm}
								\includegraphics[width=0.12\textwidth]{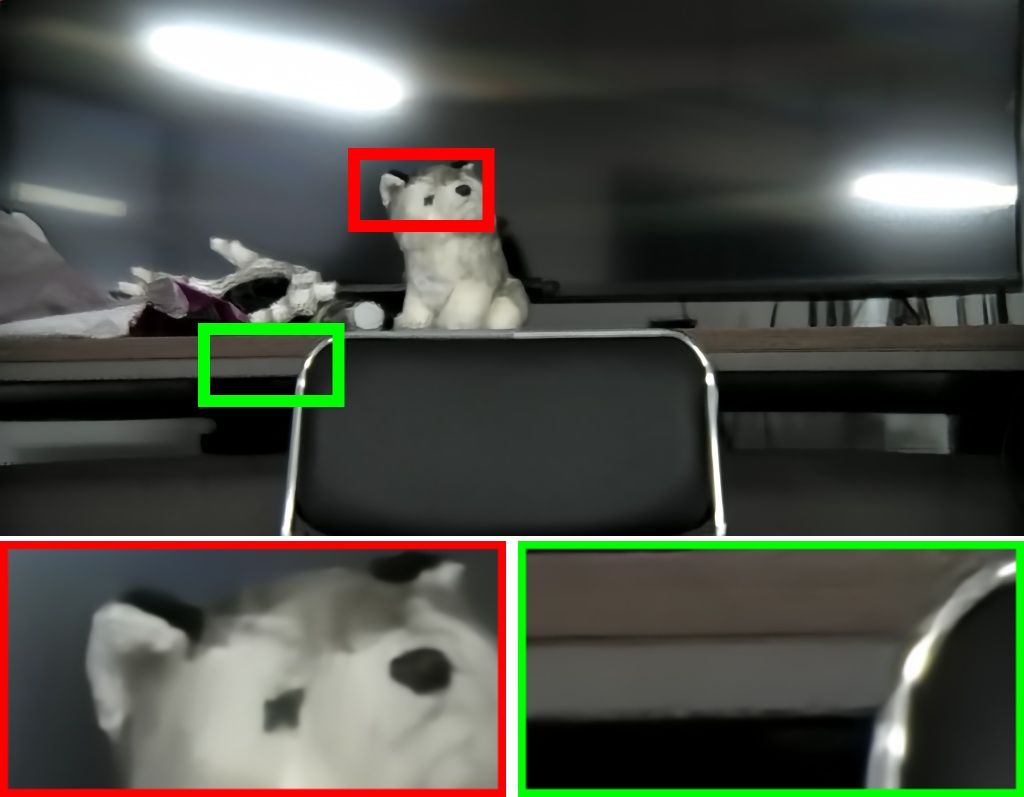}&\hspace{-4mm}
								\includegraphics[width=0.12\textwidth]{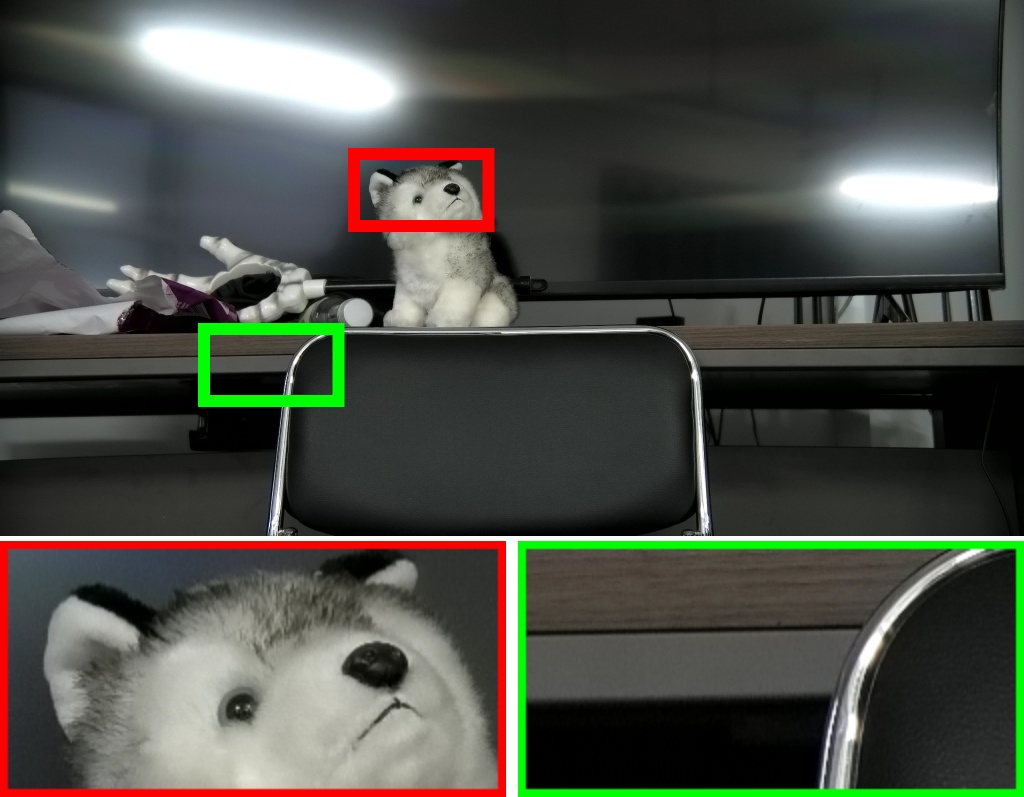}
								\\
								\hspace{-4mm}
								\includegraphics[width=0.12\textwidth]{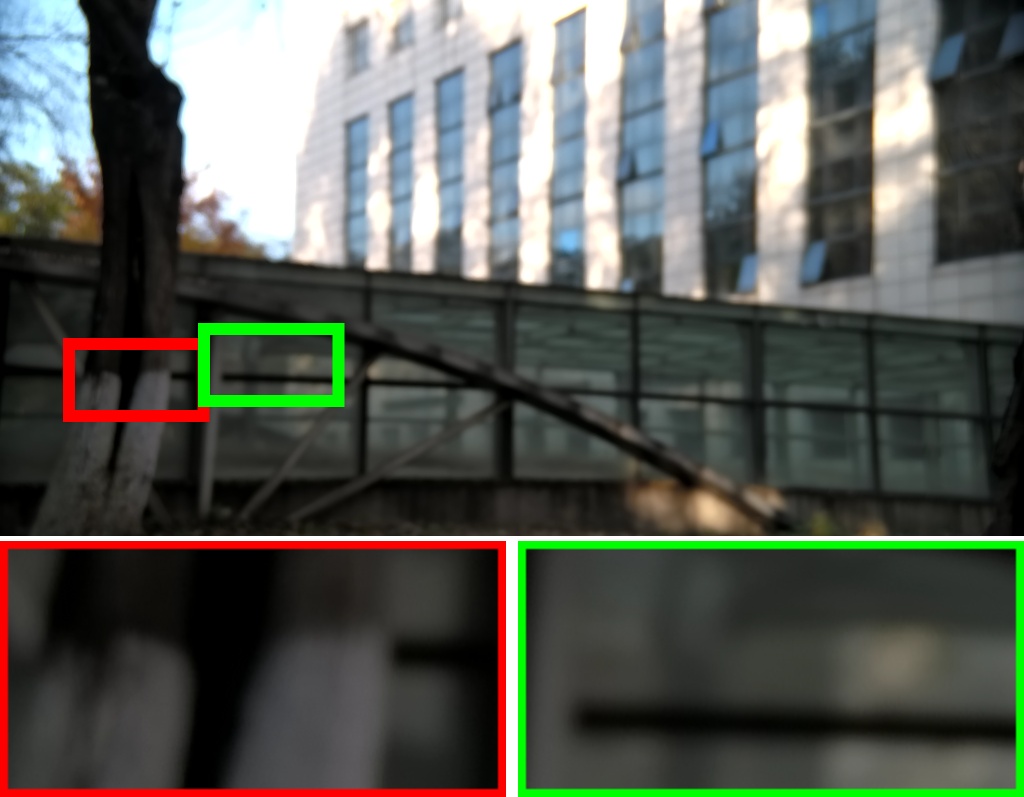}&\hspace{-4mm}
								\includegraphics[width=0.12\textwidth]{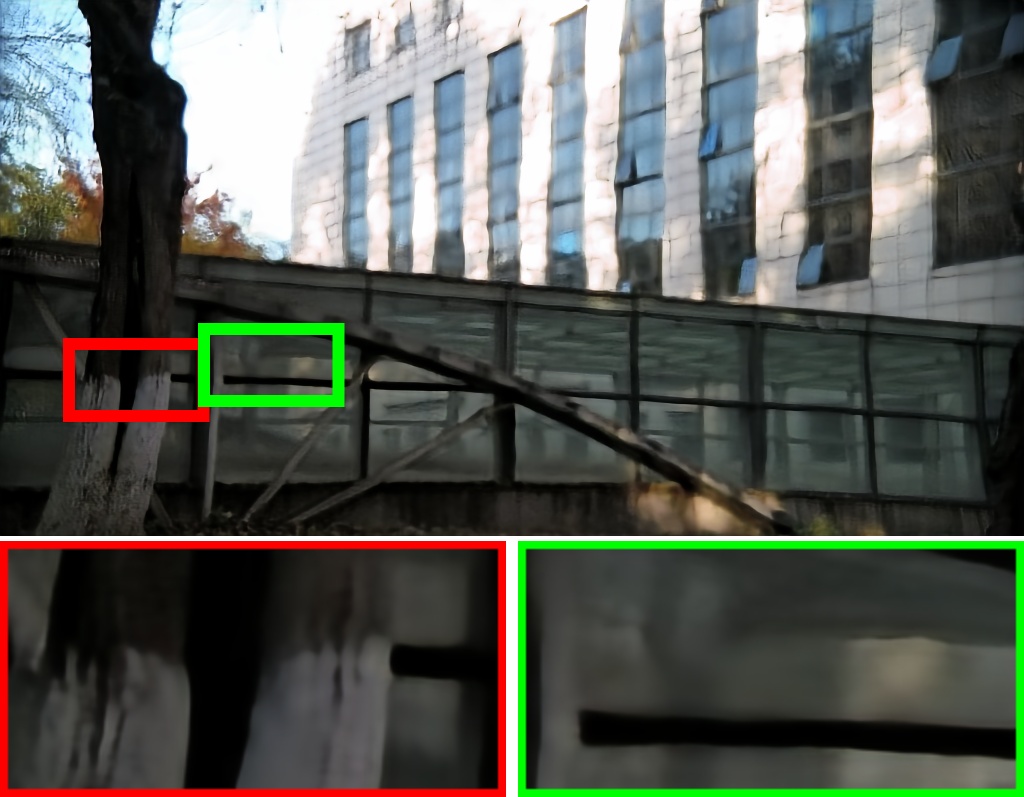}&\hspace{-4mm}
								\includegraphics[width=0.12\textwidth]{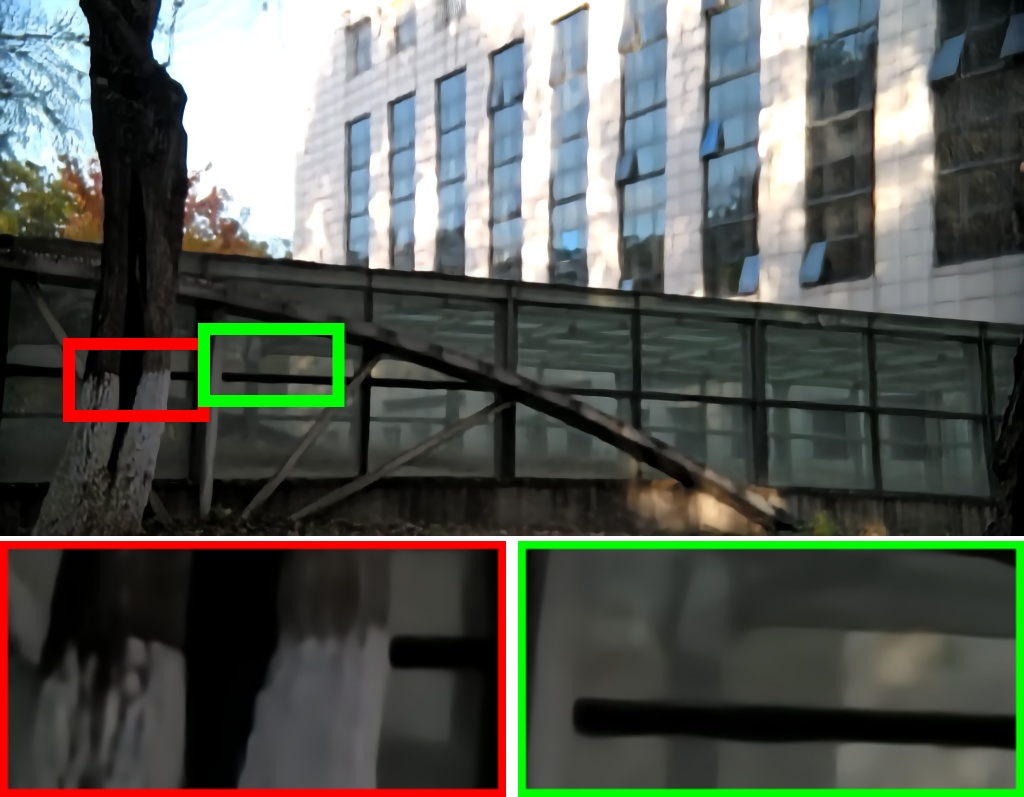}&\hspace{-4mm}
								\includegraphics[width=0.12\textwidth]{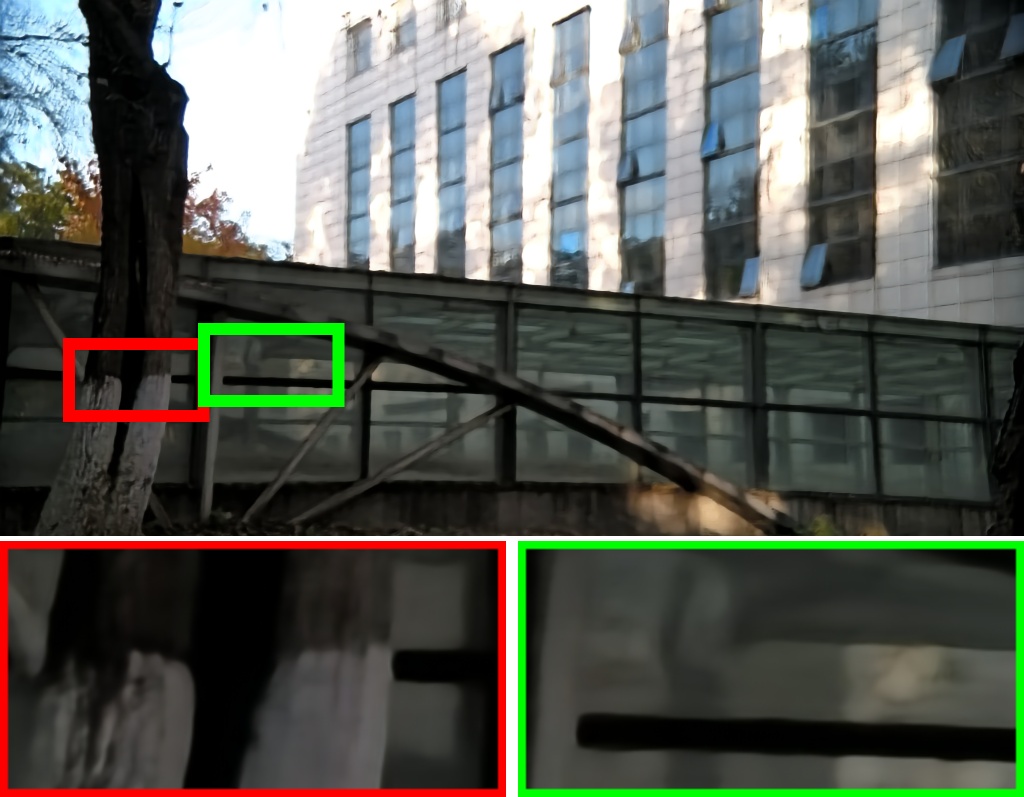}&\hspace{-4mm}
								\includegraphics[width=0.12\textwidth]{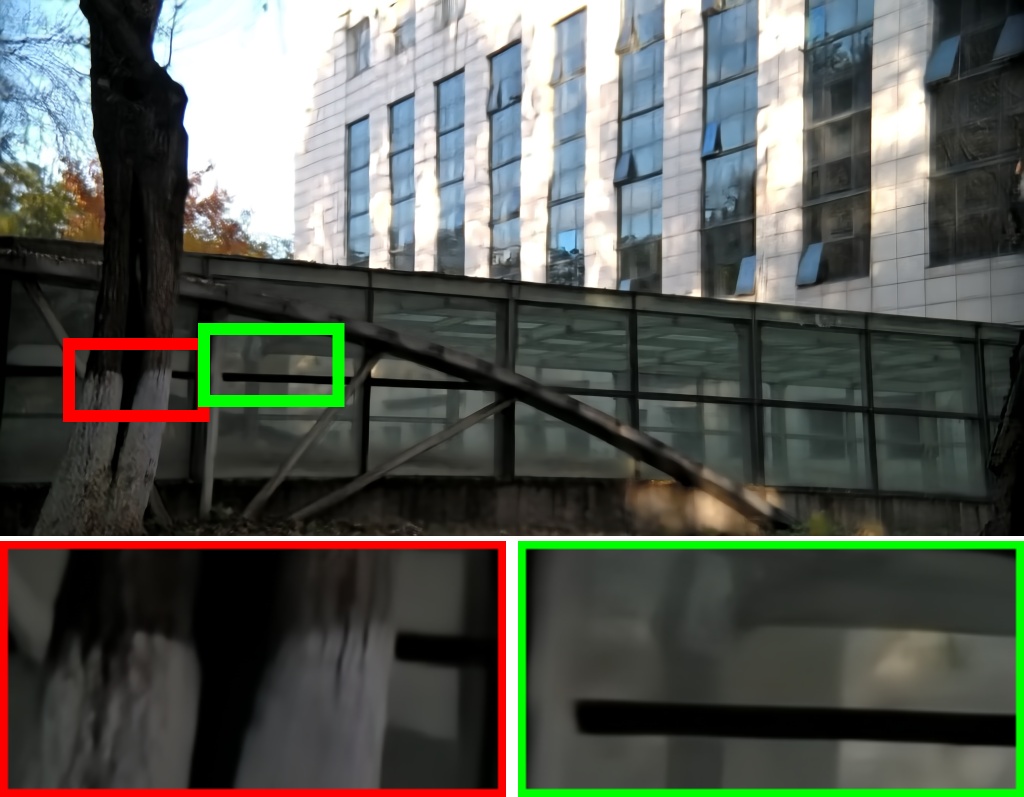}&\hspace{-4mm}
								\includegraphics[width=0.12\textwidth]{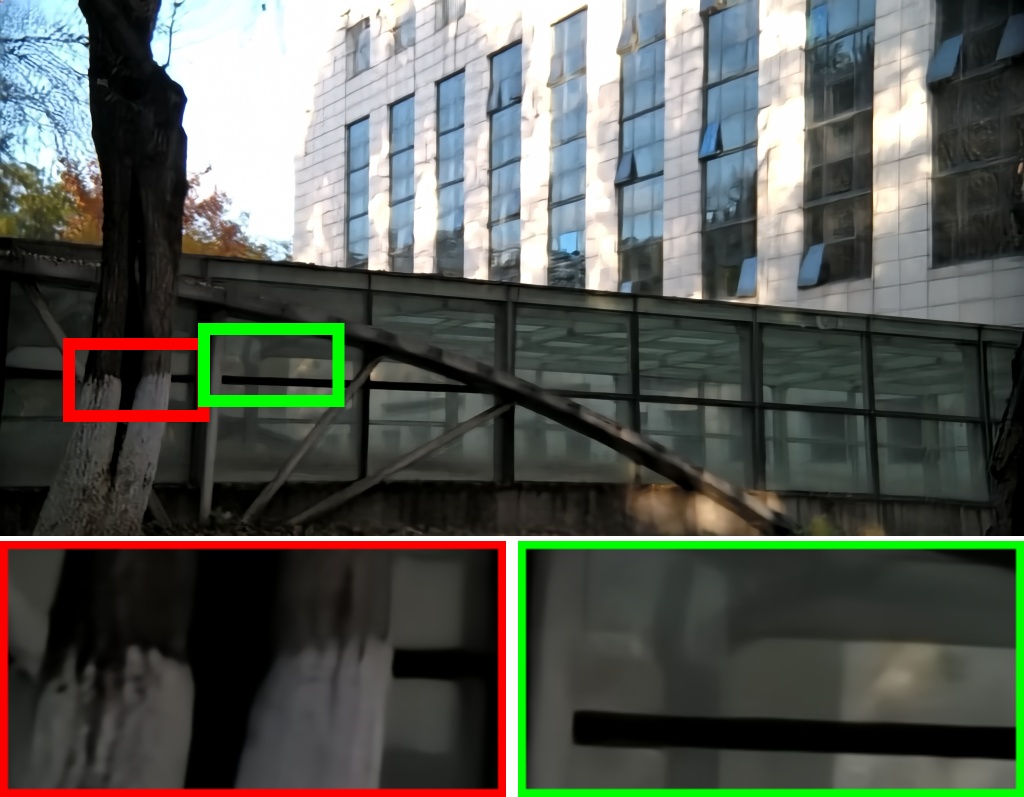}&\hspace{-4mm}
								\includegraphics[width=0.12\textwidth]{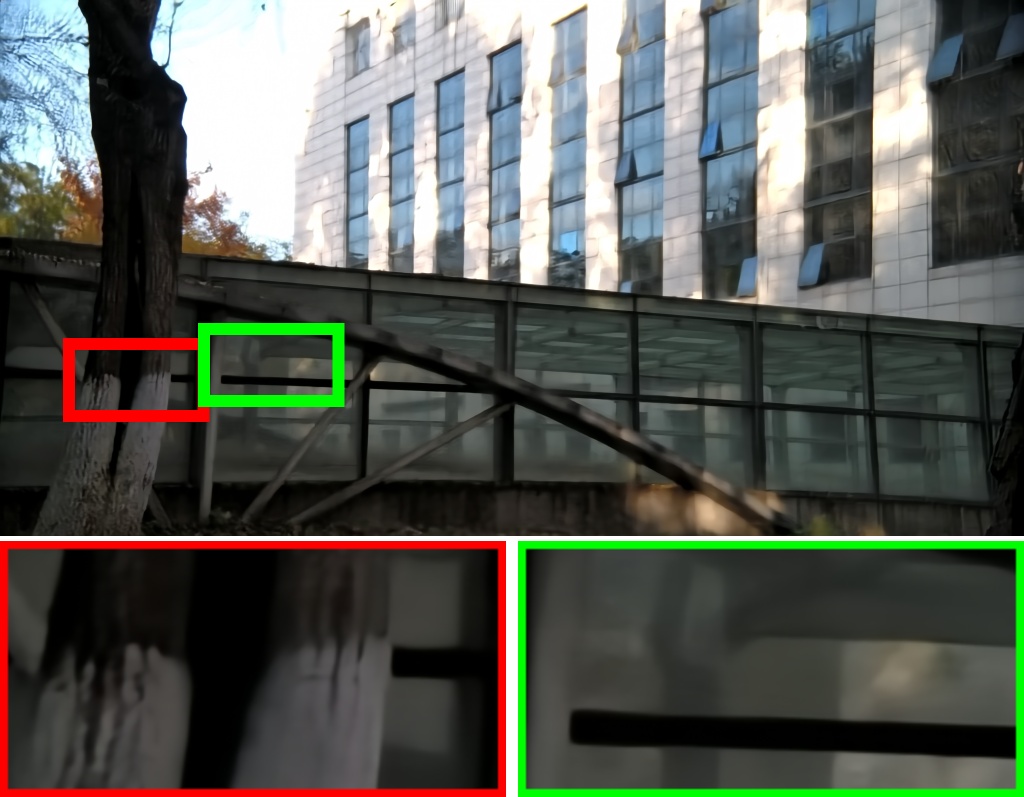}&\hspace{-4mm}
								\includegraphics[width=0.12\textwidth]{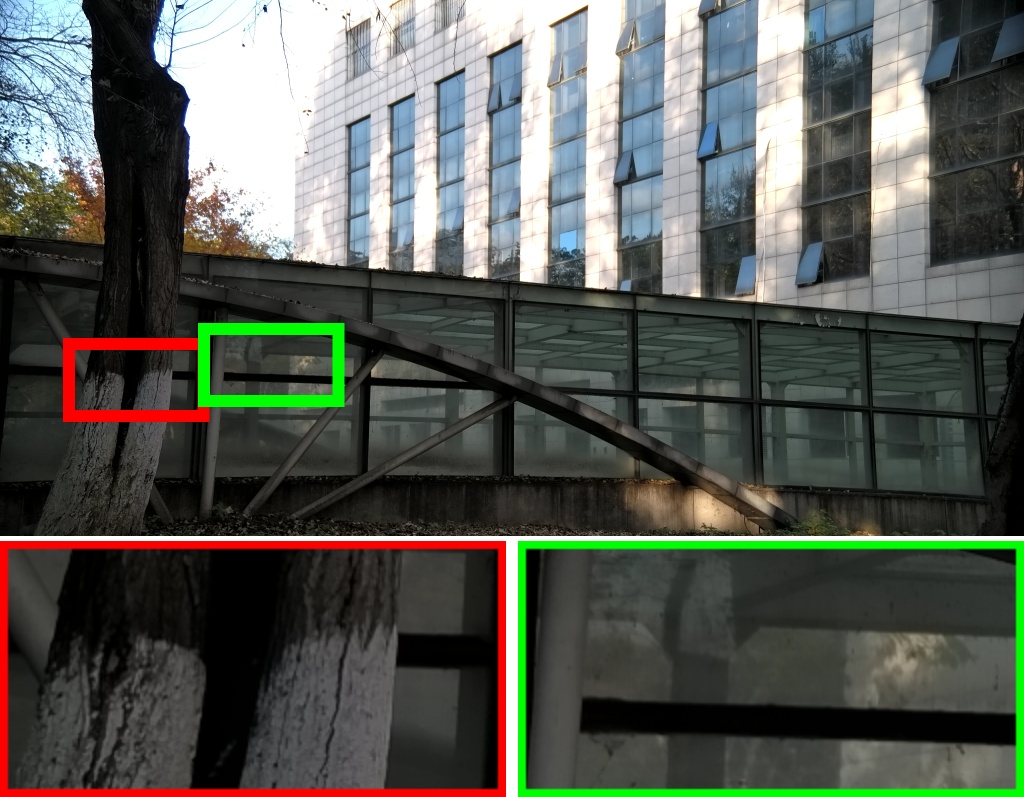}
								\\
								Input&\hspace{-4mm}                DPDNet$_{S}$~\cite{abuolaim2020defocus}&\hspace{-4mm}
								MPRNet~\cite{zhang2019deep}& \hspace{-4mm}
								Restormer\cite{zamir2022restormer}& \hspace{-4mm} Loformer\cite{mao2024loformer}&\hspace{-4mm}
								Loformer*&\hspace{-4mm}
								$\text{Loformer}^{\dagger}$&\hspace{-4mm}
								GT
								\\
							\end{tabular}
						\end{adjustbox}
						
					\end{tabular}
					\caption{\small Visual comparison of competing methods on SDD dataset.Loformer$^*$ is trained using Ours(Eq. \eqref{eq:jdrl-simple}),while $\text{Loformer}^{\dagger}$ is trained using ours(Eq. \eqref{eq:jdrl}). 
					}
					\label{fig:MDD_fig}
				\end{figure*}

The visualizations shown in Fig. \ref{fig:MDD_fig} prove that our
method overcomes the limitations of misaligned training pairs, achieving superior restoration of fine details in the presence of strong defocus blur.
% {\color{red}
In the green box in the fifth row, pixel misalignment leads to optimization discrepancies during training, causing severe distortion in the image restoration process of basic models such as DPDNet\cite{abuolaim2020defocus}, MPRNet\cite{zhang2019deep}, and Restormer\cite{zamir2022restormer}. The edges of the table and chair, which should be straight, are restored into pronounced curves, which contradicts our visual perception.
In the first row, a similar phenomenon can also be observed. Although Loformer\cite{mao2024loformer} performs well in image restoration, it experiences local ``collapse" in training tasks with misaligned data. In the green box, Loformer\cite{mao2024loformer} exhibits a ``truncation" phenomenon during the restoration process, resulting in poor visual quality. It is clear that this issue can be effectively addressed by our training framework.

%
% For example, in the second row of images, our method produces the sharpest lines while avoiding blur artifacts and distortions.
\begin{table}[t]
	\centering
	\vspace{-1.2em}
	\caption{\small{Quantitative comparison of competing methods on SDD dataset. 
			The methods marked with notation $*$ are trained using Ours(Eq. (1)), and the methods marked with notation ${\dagger}$ are trained using Ours(Eq. (2)). 
	}}
	\label{tab:SDD more metrics}
	\footnotesize
	\setlength{\tabcolsep}{4pt}
	\begin{tabular}{c|ccccc}
		\toprule
		Method  &  PSNR$\uparrow$  & SSIM$\uparrow$   & LPIPS$\downarrow$ & FID$\downarrow$ & DISTS$\downarrow$ \\
		\midrule
		UNet & 24.62 & 0.758  & 0.344 & 81.82 & 0.212 \\
		DMPHN~\cite{zamir2021multi} & 25.00 & 0.769 & 0.326 & 71.47 & 0.208 \\
		DPDNet$_{S}$~\cite{abuolaim2020defocus} & 24.81 & 0.760 & 0.343 & 75.66 & 0.210 \\
		MPRNet~\cite{zhang2019deep} & 26.28 & 0.796 & 0.302 & 62.32 & 0.202 \\
		UFormerT~\cite{wang2022uformer} & 25.68 & 0.774 & 0.321 & 70.24 & 0.208 \\
		Restormer\cite{zamir2022restormer} & 26.39 & 0.806 & 0.301 & 58.90 & 0.189 \\
		Loformer\cite{mao2024loformer} & 26.51 & 0.808 & 0.296 & 54.80 & 0.179 \\
		\midrule
		UNet* & 25.82 & 0.783 & 0.305 & 58.53 & 0.181 \\
		MPRNet* & 26.88 & 0.810 & 0.265 & 55.30 & 0.177 \\
		UformerT* & 26.53 & 0.808 & 0.297 & 56.18 & 0.180 \\
		Restormer* & 26.89 & 0.817 & 0.257 & 51.22 & 0.168 \\
		Loformer* & 26.95 & 0.820 & 0.255 & 48.77 & 0.165 \\
		\midrule
		UNet$^{\dagger}$ & 26.20 & 0.796 & 0.298 & 55.49 & 0.180 \\
		Restormer$^{\dagger}$ & 27.08 & 0.819 & 0.253 & 50.64 & 0.166 \\
		Loformer$^{\dagger}$ & \textbf{27.20} & \textbf{0.826} & \textbf{0.251} & \textbf{47.91} & \textbf{0.164} \\
		\bottomrule
	\end{tabular}
\end{table}

\begin{table}[t]
	% \footnotesize%
	% \arrayrulewidth0.5pt
	\centering
	\caption{\small{Comparison of computational costs of competing methods. 
			For the methods marked with $*$, deblurring models are exactly same with their original ones. 
			For the methods marked with ${\dagger}$, deblurring module introduces $\mathcal{E}$ and $\mathcal{F}$. }
	}
	\label{tab:flops}
	\footnotesize
	% \resizebox{\textwidth}{!}{
		\setlength{\tabcolsep}{12pt}
		% \resizebox{0.5\textwidth}{!}{
			\begin{tabular}{c|cc}
				\toprule
				% 		& \multicolumn{4}{|c}{Deformed GT} \\
				% 		\midrule
				Method & FLOPs(G) & Params(M)\\
				\midrule
				% UNet& - & -\\
				MPRNet~\cite{zhang2019deep} & 6830 & 20.1\\
				UFormerT~\cite{wang2022uformer}& 42.7 & 5.20\\
				Restormer\cite{zamir2022restormer} & 564 & 26.1\\
				% GRL$_{S}$-B\cite{li2023efficient}& 2120G & 6.65M\\
				Loformer\cite{mao2024loformer} & 331 & 27.9\\
				% 			\midrule
				% UNet & 24.62 & 0.758  & 0.344 & - & -\\	
				\midrule
				% UNet*  & -&-\\
				MPRNet*  & 6830 & 20.1 \\
				UformerT* & 42.7 & 5.20 \\	
				Restormer* & 564 & 26.1\\
				Loformer*& 331 & 27.9\\
				\midrule
				% UNet^{\dagger}  &-&-\\	
				$\text{Restormer}^{\dagger}$  & 582&26.2 \\	
				$\text{Loformer}^{\dagger}$&348&28.0\\
				\bottomrule
				\hline
			\end{tabular}
		\end{table}

				\subsection{Evaluation on DPDD Dataset}
				For the DPDD~\cite{abuolaim2020defocus} dataset, besides the methods mentioned above, two state-of-the-art single image defocus deblurring methods Son et al. \cite{son2021single} and IFAN \cite{lee2021iterative} are also compared. 
				It is worth noting that although the DPDD dataset is meant to be aligned, there actually exists slight misalignment as shown in Fig.~\ref{intro}. 
				Therefore, we calculate the evaluation metrics in view of  $\hat{\bm{I}}$ \emph{v.s.} $\bm{I}_S^w$.
				The experimental results are reported in Table~\ref{tab:DPDD_tab}.
				%  \begin{table}[t]%
					% % \footnotesize%
					% % \arrayrulewidth0.5pt
					%     \centering
					%     \caption{\small{Quantitative comparison on DPDD dataset \cite{abuolaim2020defocus}.
							%     % The results are tested with warped groundtruth
							%     }}
					%     \label{tab:DPDD_tab}
					%     \footnotesize
					%     % \resizebox{0.5\textwidth}{!}{
						%     \setlength{\tabcolsep}{8pt}
						%     \begin{tabular}{c|cccc}
							%         \toprule
							%         Method  &  \ PSNR$\uparrow$ \  & \ SSIM$\uparrow$\    &\ LPIPS$\downarrow$\ \\
							%         \midrule
							%         JNB~\cite{Shi_2015_CVPR} &  21.30 & 0.692  & 0.458 \\
							%         EBDB~\cite{karaali2017edge} & 22.66 & 0.737 & 0.397 \\
							%         DMENet~\cite{Lee_2019_CVPR} &  22.89 & 0.740 & 0.388 \\
							%         \midrule
							%         DPDNet$_{S}$~\cite{abuolaim2020defocus} & 24.48 & 0.778  & 0.262 \\	
							%         Son et al.\cite{son2021single} & 25.49 & 0.807  & 0.212 \\
							%         IFAN \cite{lee2021iterative}& 25.86 & 0.825  & 0.192 \\
							%         % 			\midrule
							%         %UNet & 24.63 & 0.785 & 0.0424 & 0.261 \\
							%         MPRNet~\cite{zhang2019deep} & 26.03 & 0.820  & 0.214 \\
							
							%         Restormer\cite{zamir2022restormer} & 26.65 & 0.850  & \underline{0.158} \\
							%         Loformer\cite{mao2024loformer} & 26.10 & 0.840  & 0.197\\
							%         GRL$_{S}$-B\cite{li2023efficient} & \underline{26.67} & \textbf{0.853}  & 0.159 \\
							%         \midrule
							%         \ Ours \  & \textbf{26.75} &\underline{0.852} & \textbf{0.155} \\		
							%         \bottomrule
							%         \hline
							%     \end{tabular}
						%     % }
					% \end{table}
				
				\begin{table}[t]%
					% \footnotesize%
					% \arrayrulewidth0.5pt
					\centering
					\caption{\small{Quantitative comparison on DPDD dataset \cite{abuolaim2020defocus}. We only apply our framework on Restormer that achieves top performance on this dataset, resulting in Restormer$^\dagger$. 
							% The results are tested with warped groundtruth
					}}
					\label{tab:DPDD_tab}
					\footnotesize
					% \resizebox{0.5\textwidth}{!}{
						\setlength{\tabcolsep}{4pt}
						\begin{tabular}{c|ccccc}
							\toprule
							Method  &   PSNR$\uparrow$   & \ SSIM$\uparrow$ &\ LPIPS$\downarrow$&FID$\downarrow$ &DISTS$\downarrow$\\
							% \midrule
							% JNB~\cite{Shi_2015_CVPR} &  21.30 & 0.692  & 0.458 \\
							% EBDB~\cite{karaali2017edge} & 22.66 & 0.737 & 0.397 \\
							% DMENet~\cite{Lee_2019_CVPR} &  22.89 & 0.740 & 0.388 \\
							\midrule
							DPDNet$_{S}$~\cite{abuolaim2020defocus} & 24.48 & 0.778  & 0.262 &76.46 &0.158\\	
							Son et al.\cite{son2021single} & 25.49 & 0.807  & 0.212 &60.93 &0.135\\
							IFAN \cite{lee2021iterative}& 25.86 & 0.825  & 0.192 &52.79 &0.133 \\
							% 			\midrule
							%UNet & 24.63 & 0.785 & 0.0424 & 0.261 \\
							MPRNet~\cite{zhang2019deep} & 26.03 & 0.820  & 0.214 &60.68 & 0.140\\
							
							Restormer\cite{zamir2022restormer} & 26.65 & 0.850  & 0.158 &45.57 &0.104\\
							Loformer\cite{mao2024loformer} & 26.10 & 0.840  & 0.197 &53.24 &0.126\\
							% GRL$_{S}$-B\cite{li2023efficient} & 26.67 & 0.853  & 0.159 \\
							\midrule
							$\text{Restormer}^{\dagger}$  & \textbf{26.81} &\textbf{0.854} & \textbf{0.152} &\textbf{43.22} &\textbf{0.102} \\		
							\bottomrule
							\hline
						\end{tabular}
						% }
				\end{table}
				
				%
				% From Table~\ref{tab:DPDD_tab}, we have a similar observation as in Table~\ref{tab:MDD}, i.e., the learning-based methods outperform the traditional ones, and the deblurring performance can be further boosted by adopting our reblurring-guided learning framework. The best performed metrics are highlighted in bold, while the second-best metrics are underlined.
				%
				By tolerating the slight misalignment existing in DPDD dataset~\cite{abuolaim2020defocus}, JDRL contributes to better performance on testing set of DPDD. Notably, the DPDD dataset was collected under strictly controlled conditions using remote control and strictly controlled capture conditions to ensure precise alignment. However, in real-world scenarios, it is impossible to achieve perfectly aligned blurry-sharp pairs through manual control. Our misalignment training strategy can, to some extent, compensate for the discrepancies introduced during the data collection process.

				\begin{figure*}[ht]
					\small
					\centering
					\setlength{\abovecaptionskip}{5pt} 
					\setlength{\belowcaptionskip}{0pt}
					\setlength{\tabcolsep}{2pt}
					\begin{tabular}{cc}
						\footnotesize
						% \begin{adjustbox}{valign=t}
							\begin{tabular}{ccccccc}
								\includegraphics[width=0.155\textwidth]{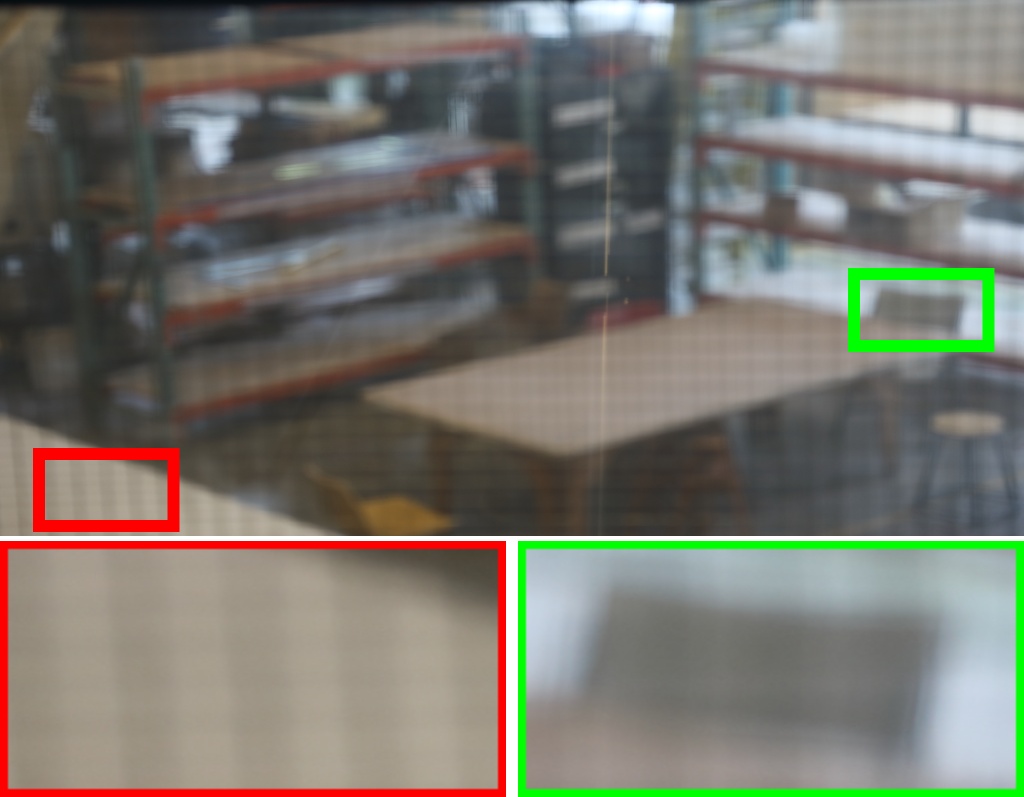}&
								\includegraphics[width=0.155\textwidth]{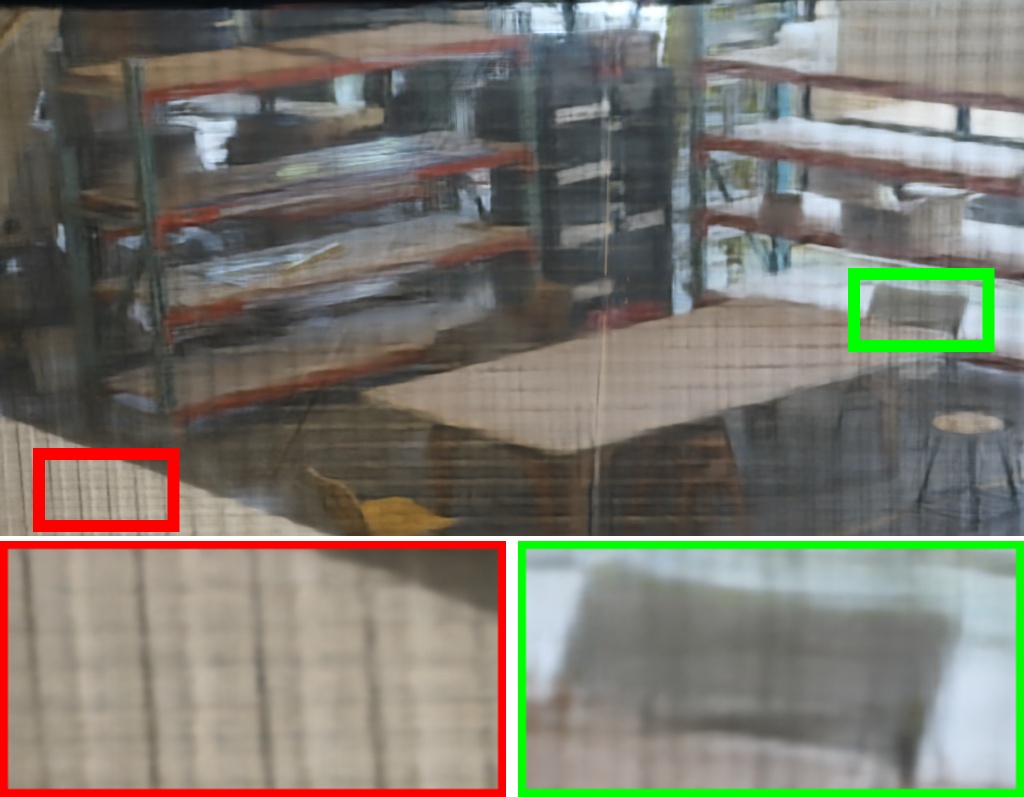}&
								\includegraphics[width=0.155\textwidth]{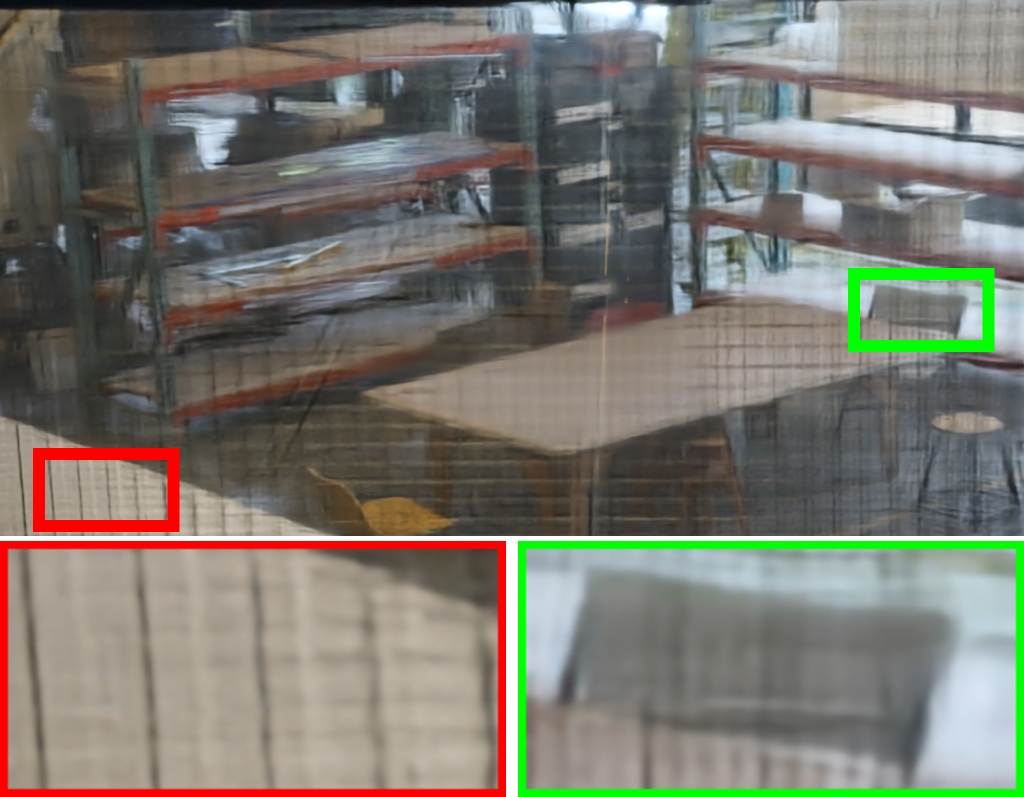}&
								\includegraphics[width=0.155\textwidth]{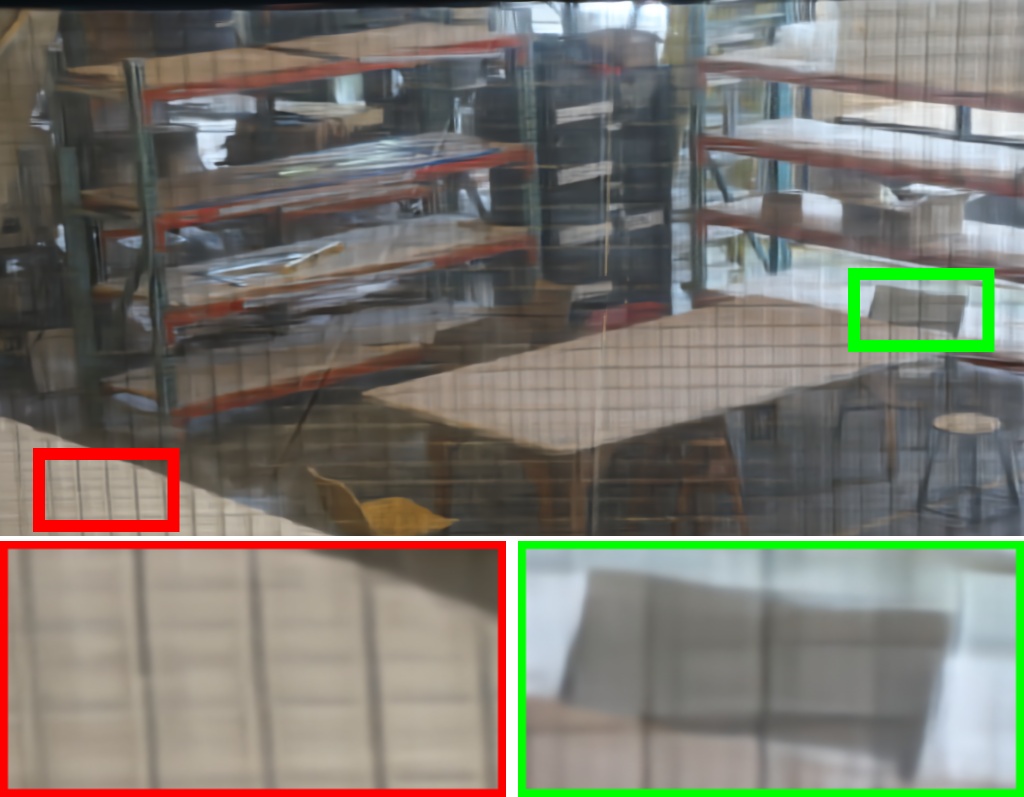}&
								\includegraphics[width=0.155\textwidth]{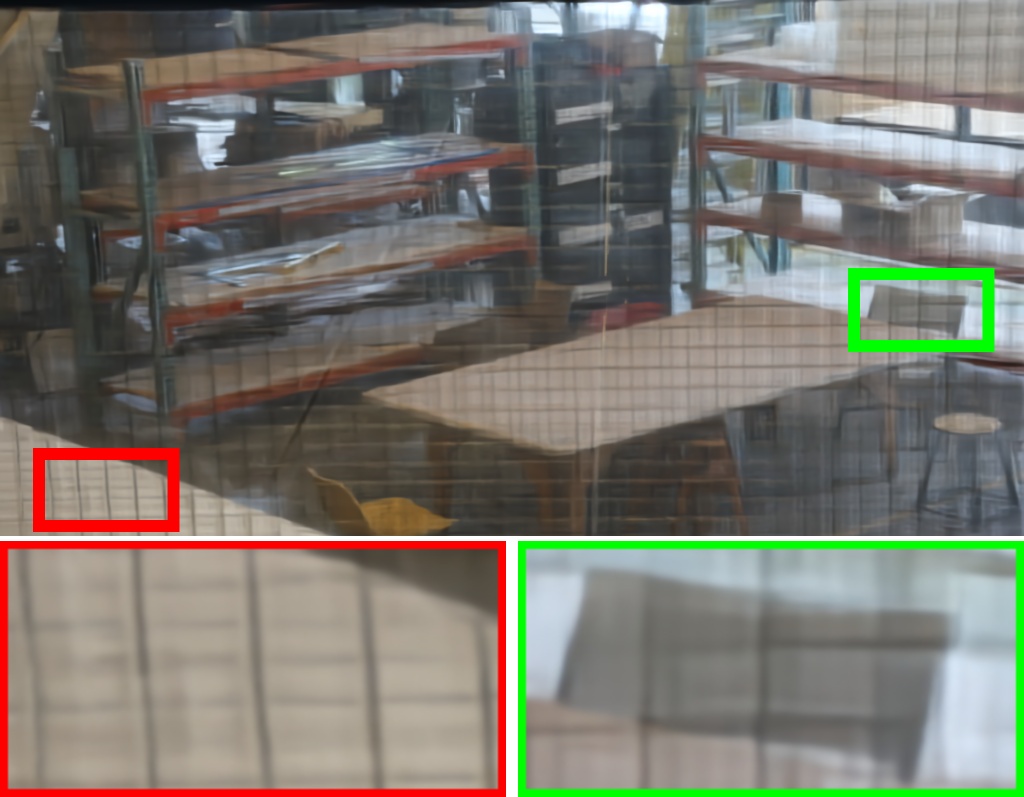}&
								\includegraphics[width=0.155\textwidth]{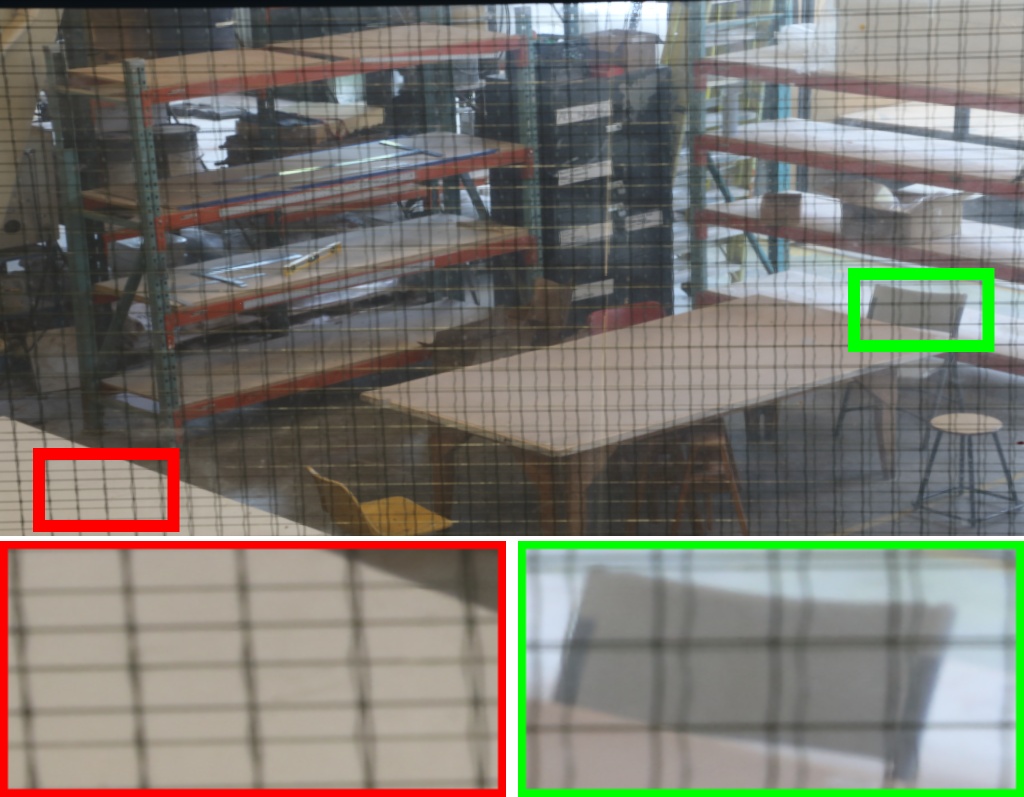}
								\\ 
								\includegraphics[width=0.155\textwidth]{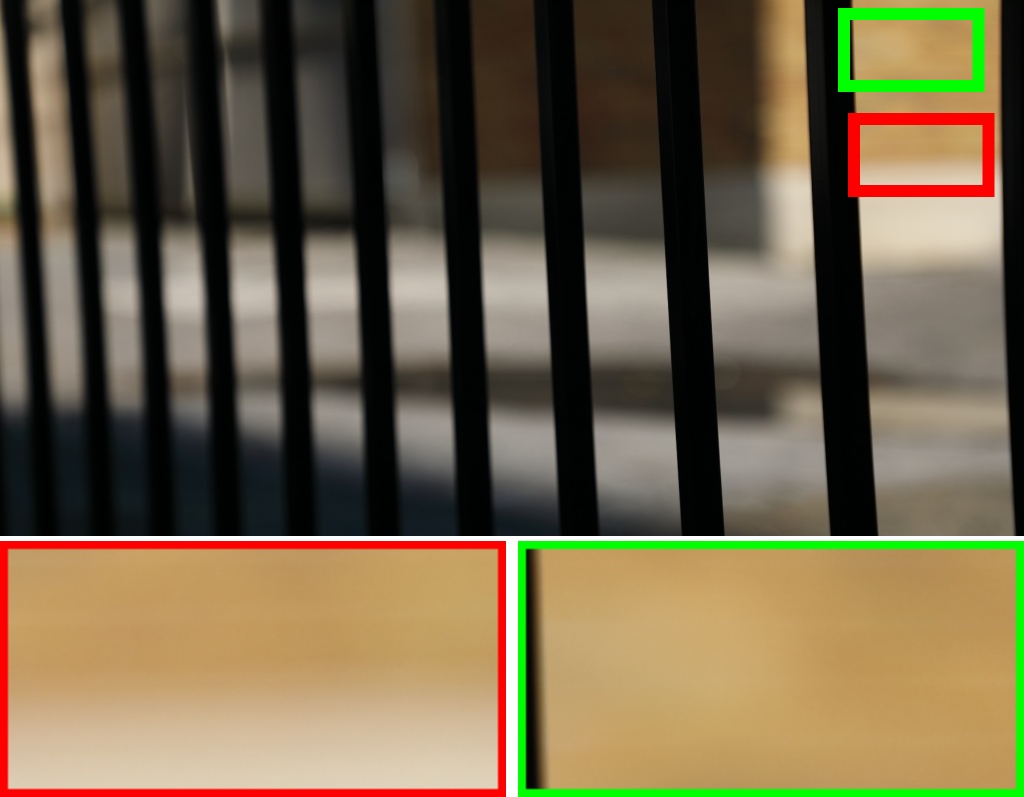}&
								\includegraphics[width=0.155\textwidth]{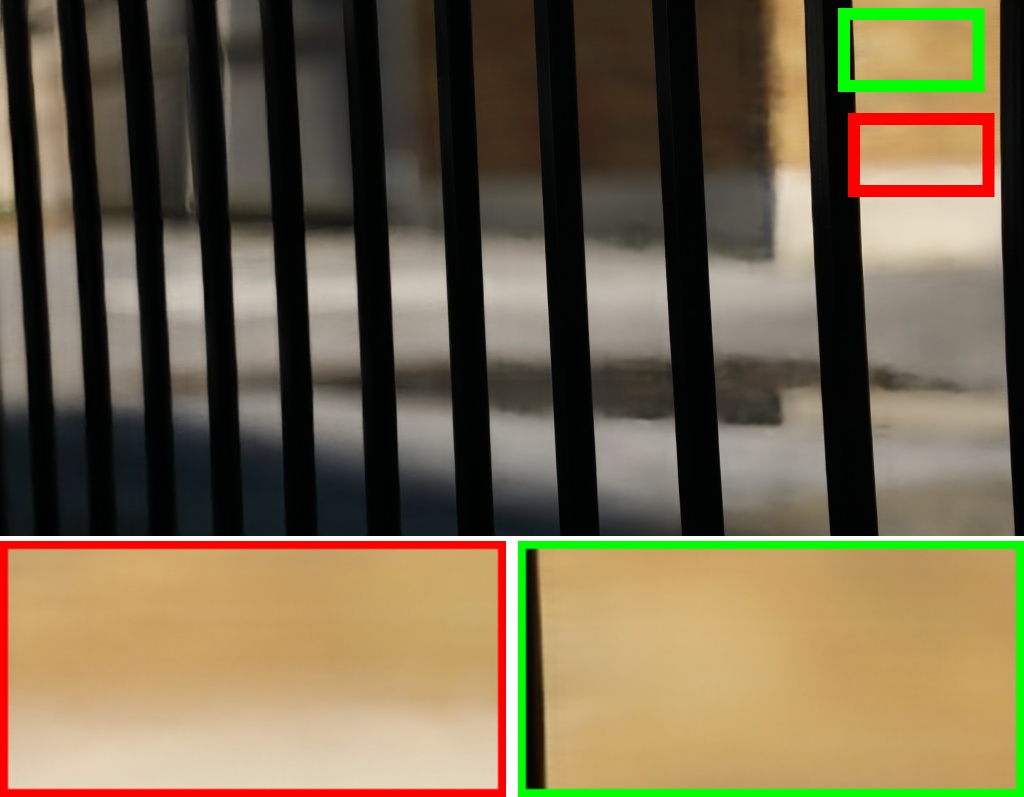}&
								\includegraphics[width=0.155\textwidth]{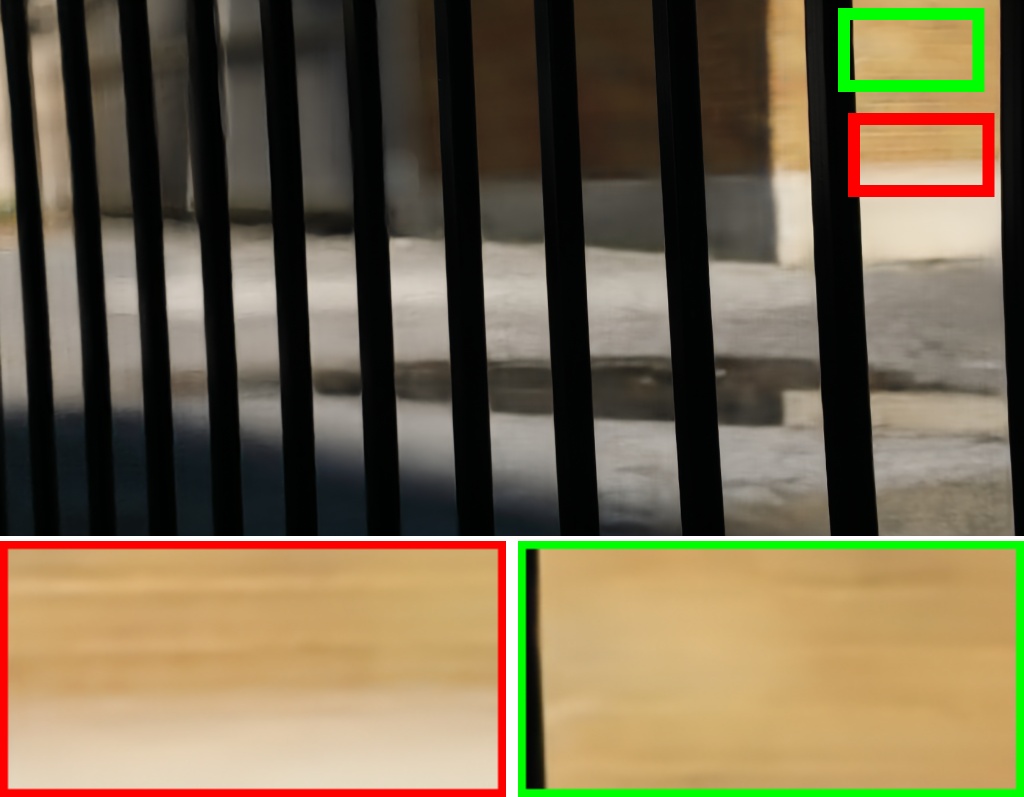}&
								\includegraphics[width=0.155\textwidth]{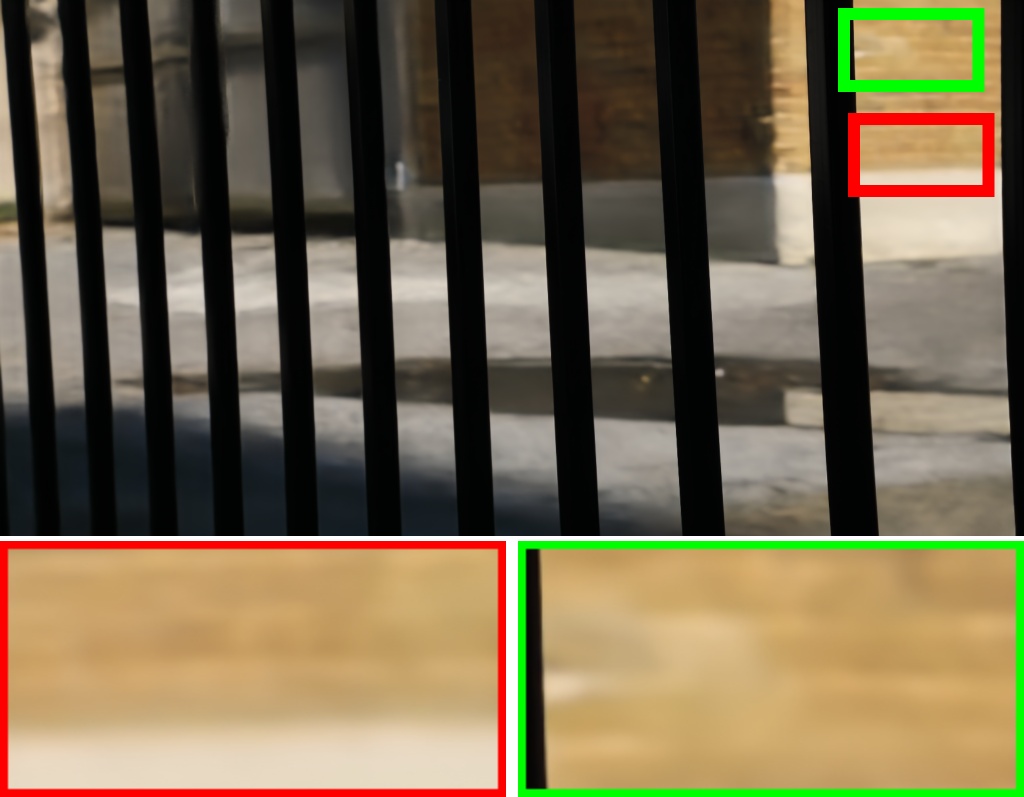}&
								\includegraphics[width=0.155\textwidth]{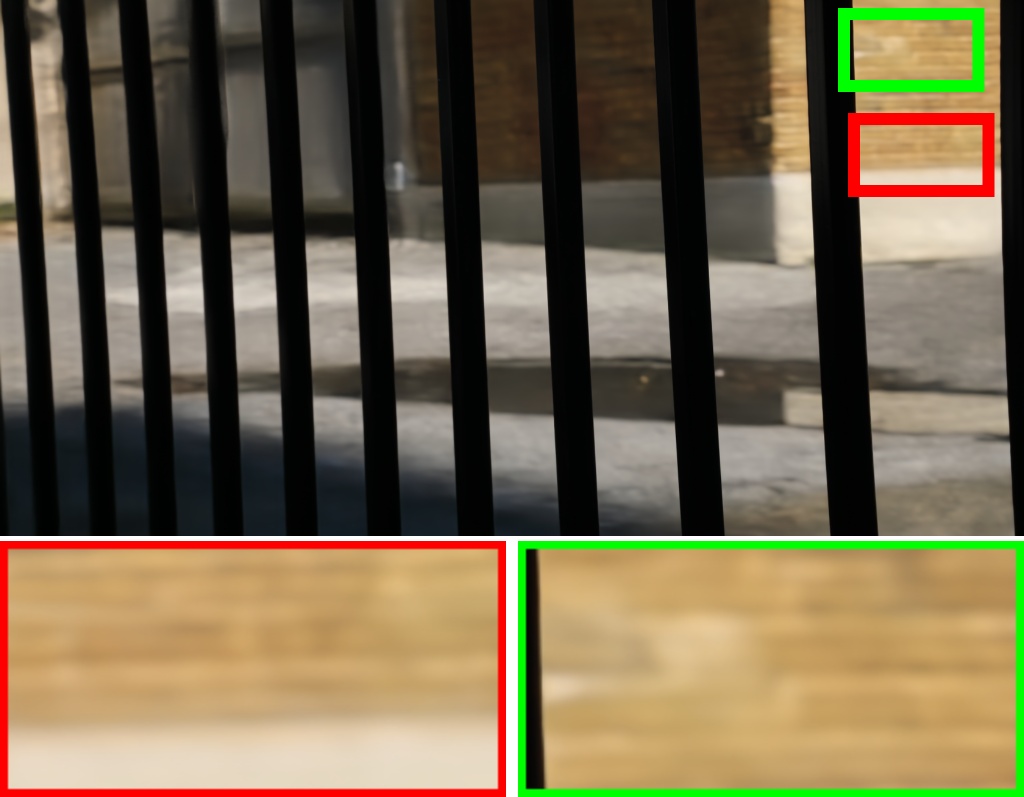}&
								\includegraphics[width=0.155\textwidth]{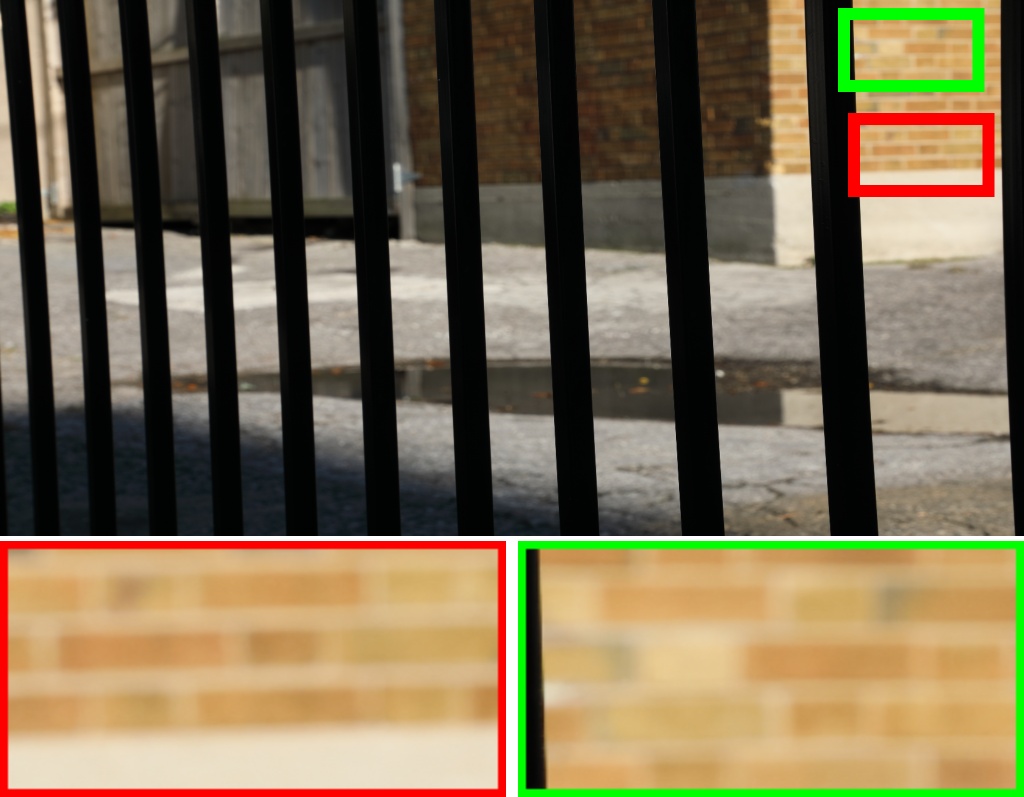}
								\\
								\includegraphics[width=0.155\textwidth]{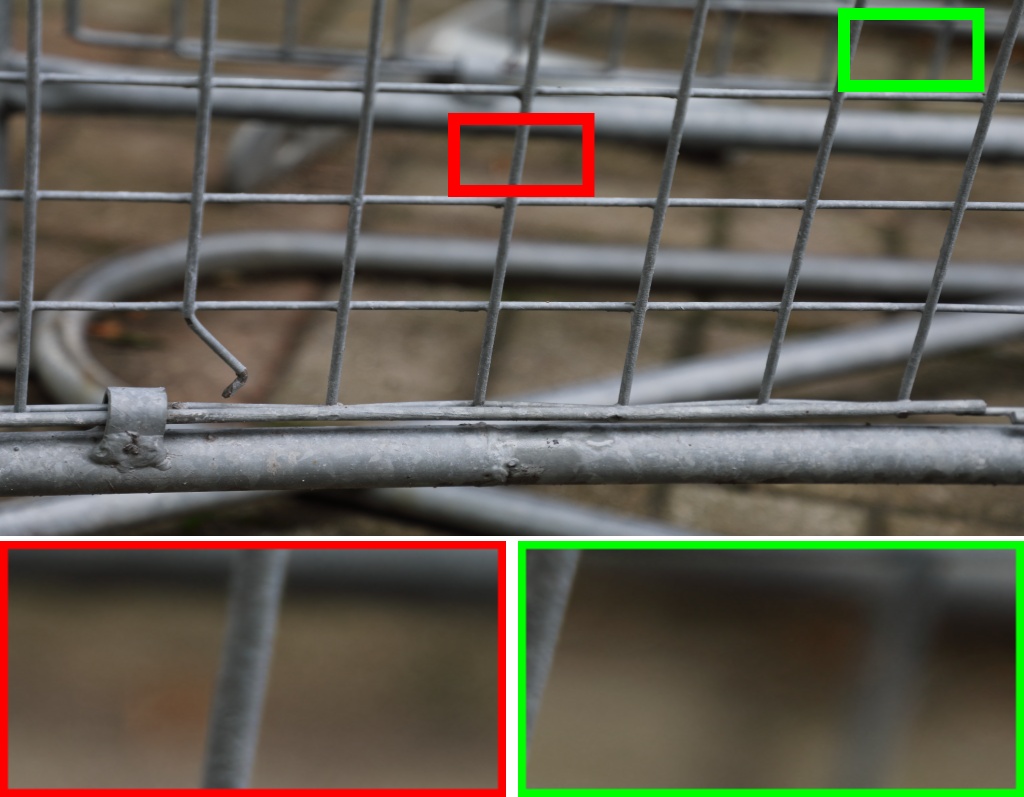}&
								\includegraphics[width=0.155\textwidth]{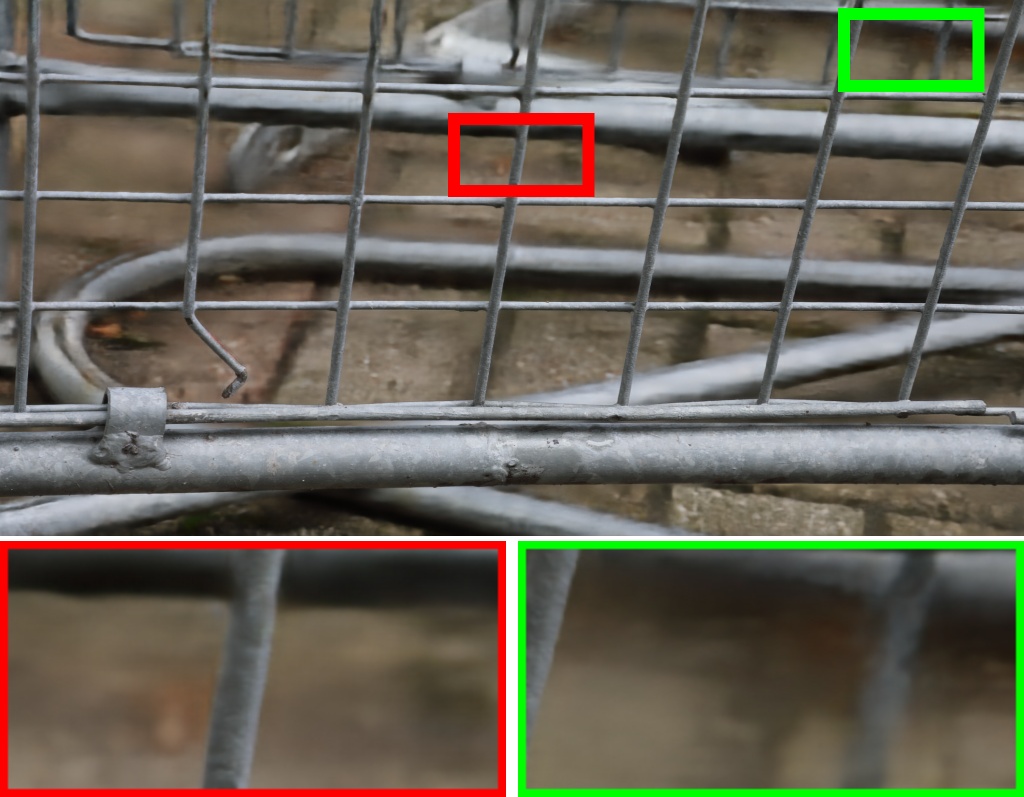}&
								\includegraphics[width=0.155\textwidth]{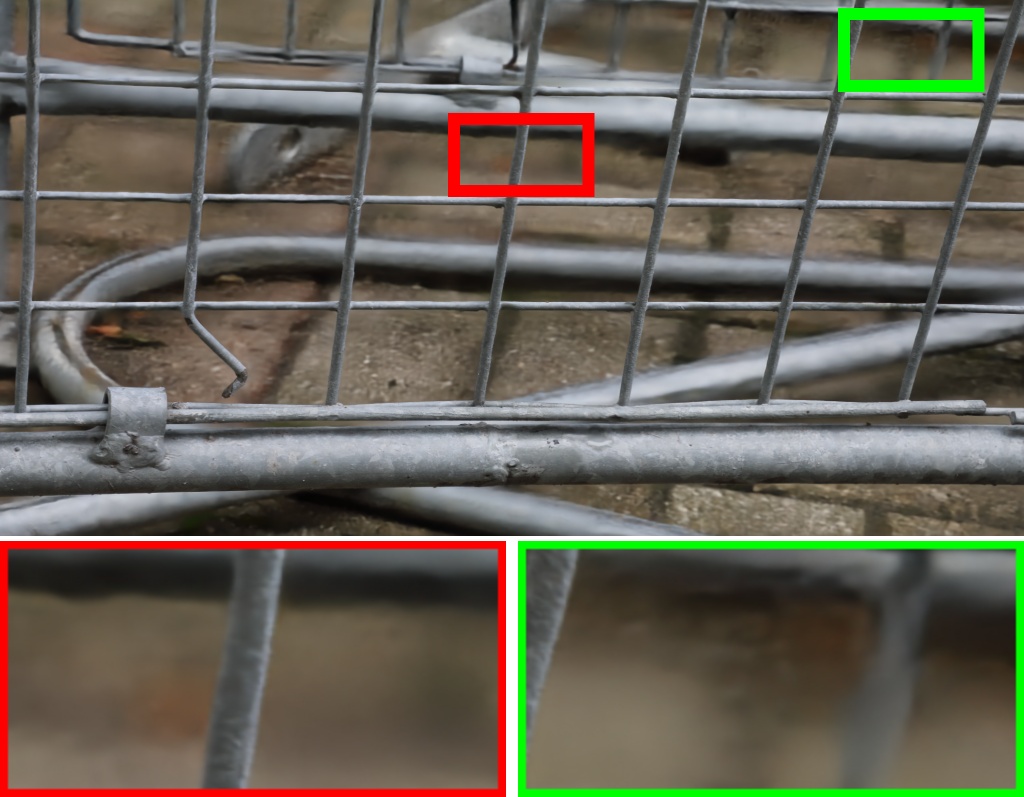}&
								\includegraphics[width=0.155\textwidth]{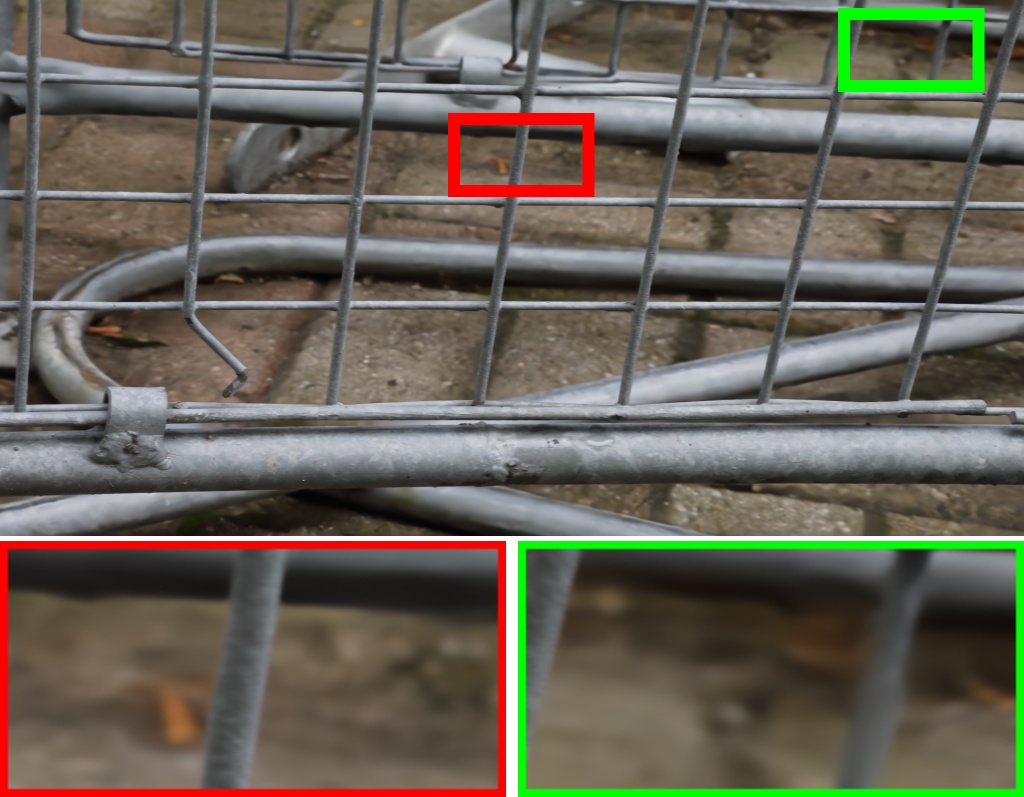}&
								\includegraphics[width=0.155\textwidth]{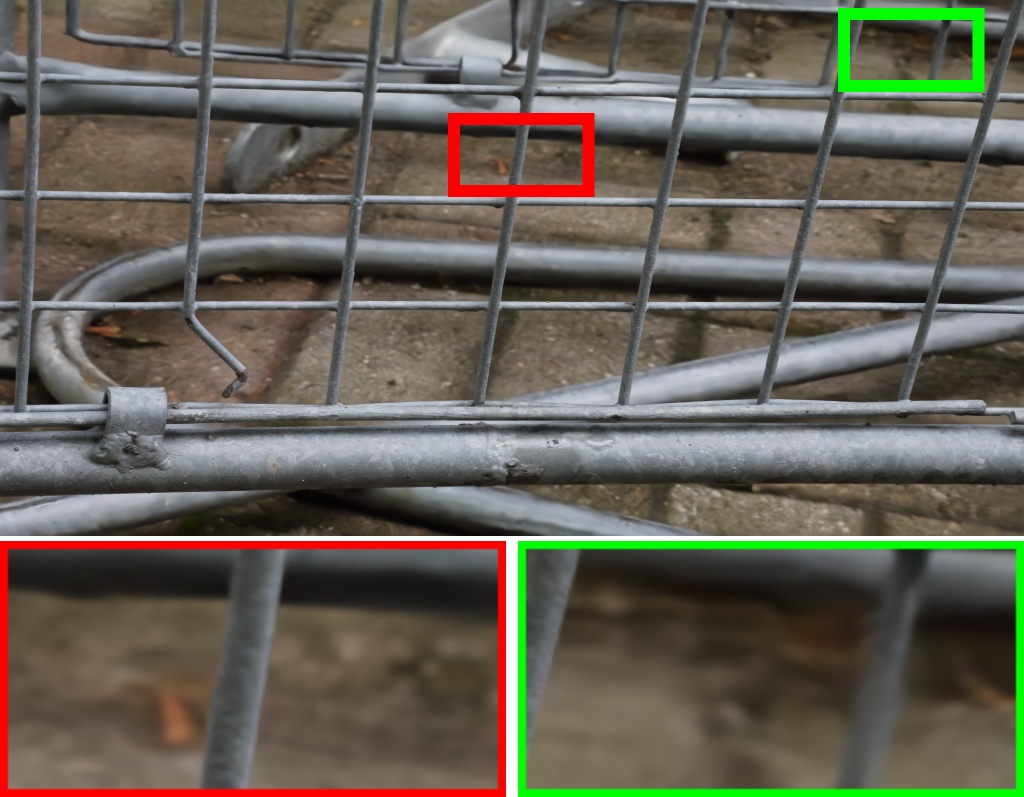}&
								\includegraphics[width=0.155\textwidth]{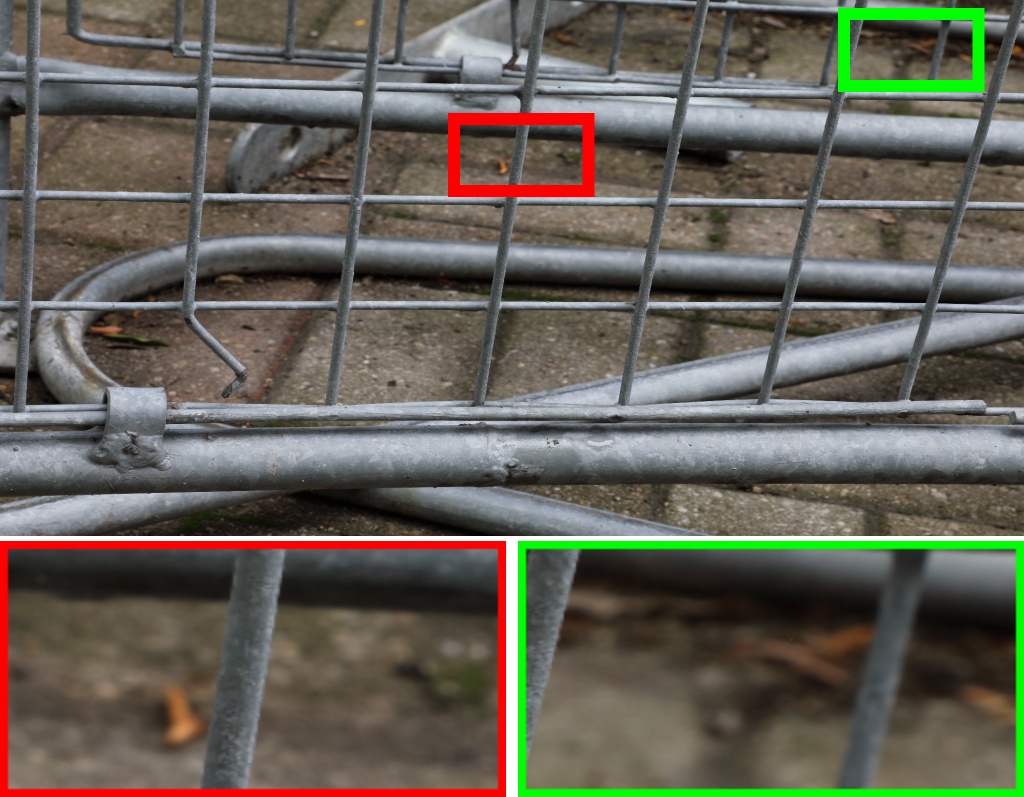}
								\\
								\includegraphics[width=0.155\textwidth]{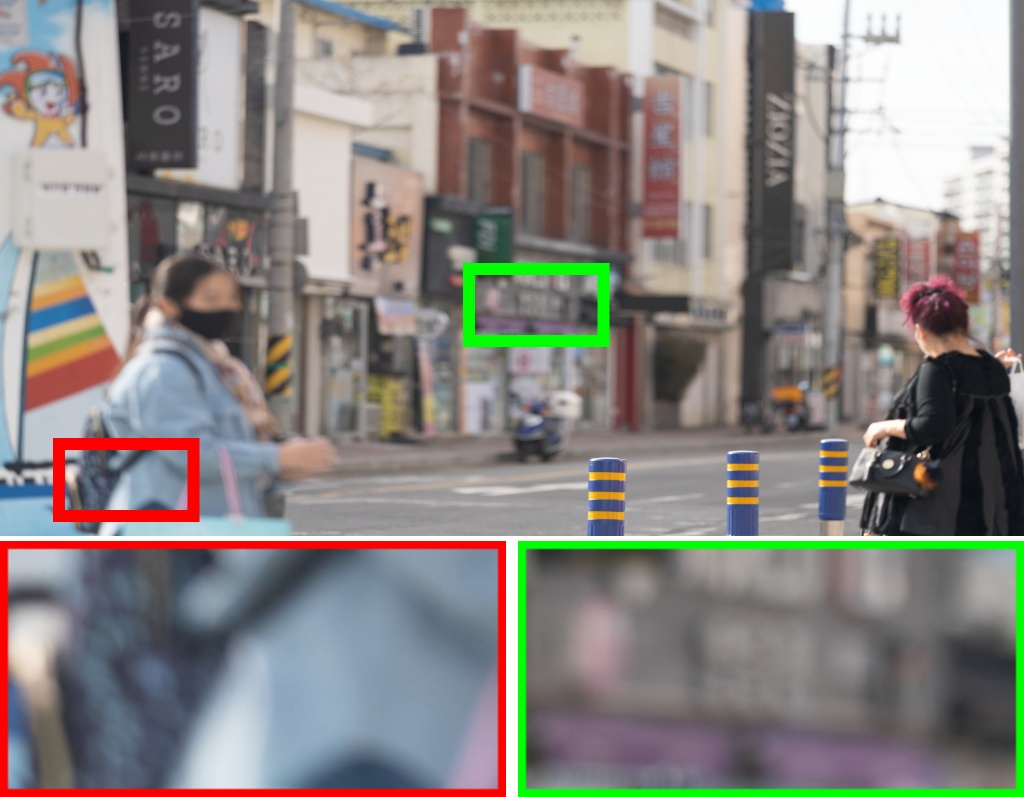}&
								\includegraphics[width=0.155\textwidth]{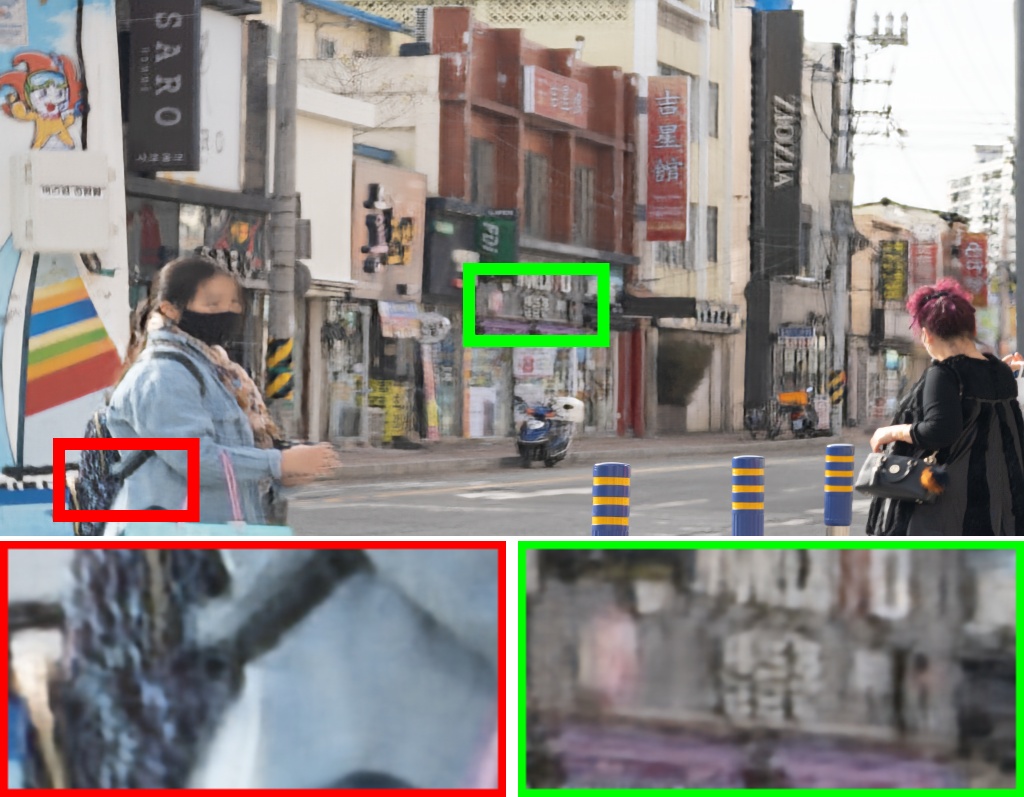}&
								\includegraphics[width=0.155\textwidth]{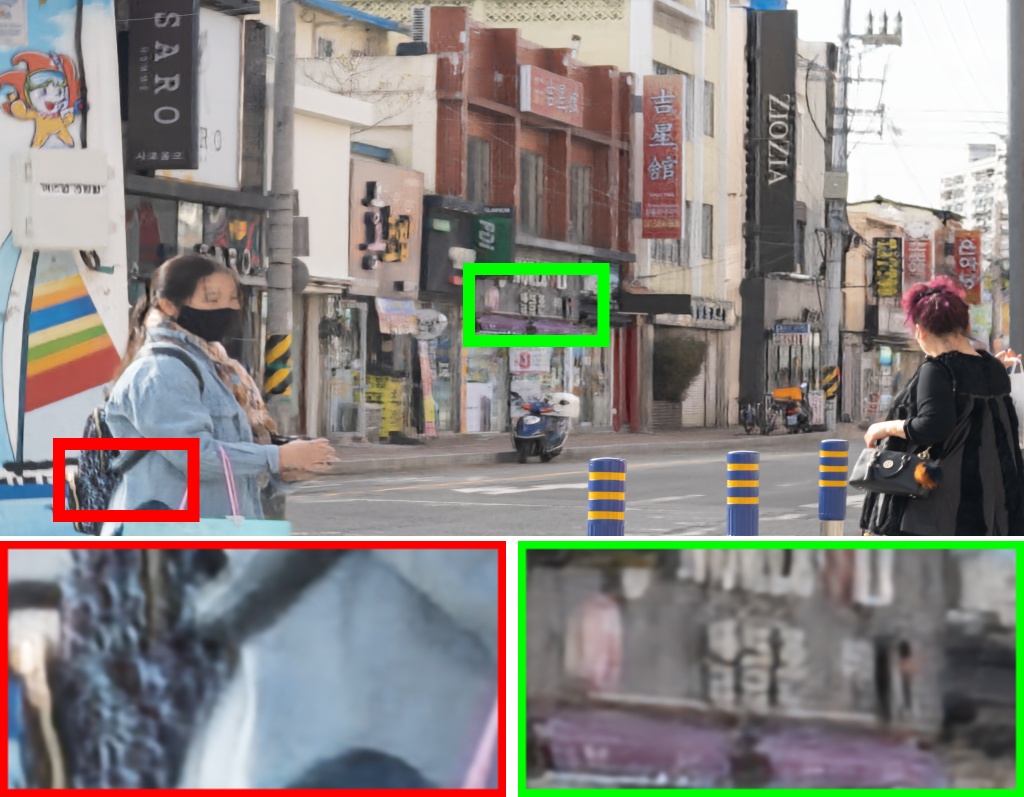}&
								\includegraphics[width=0.155\textwidth]{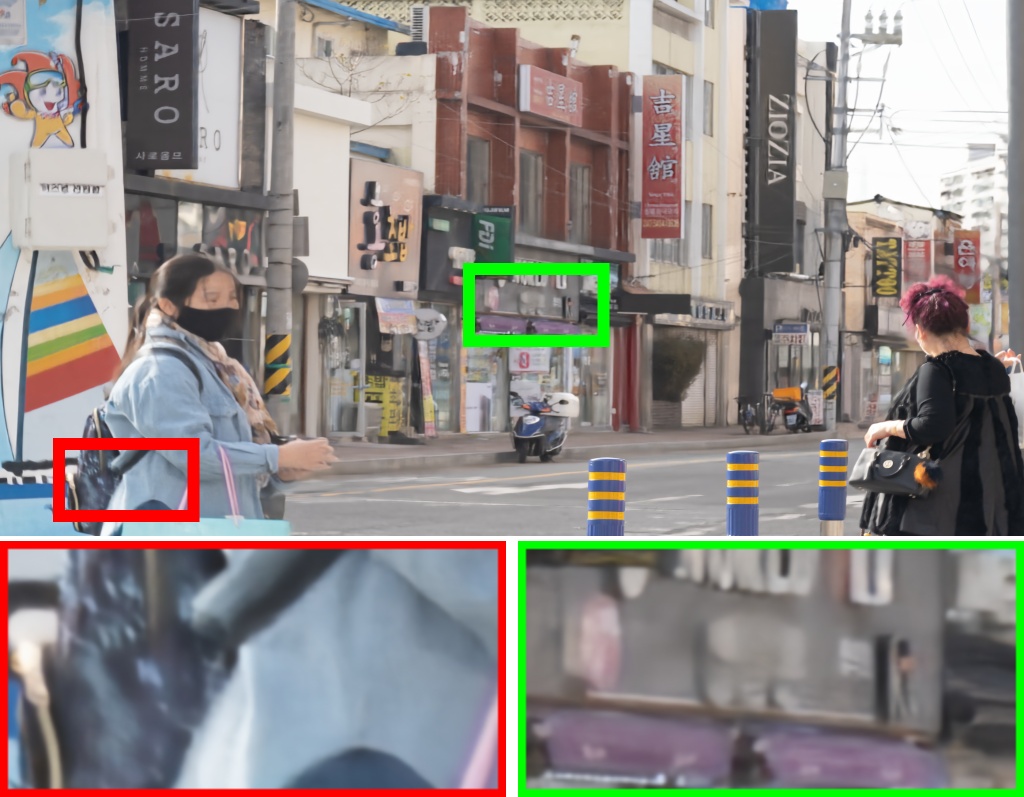}&
								\includegraphics[width=0.155\textwidth]{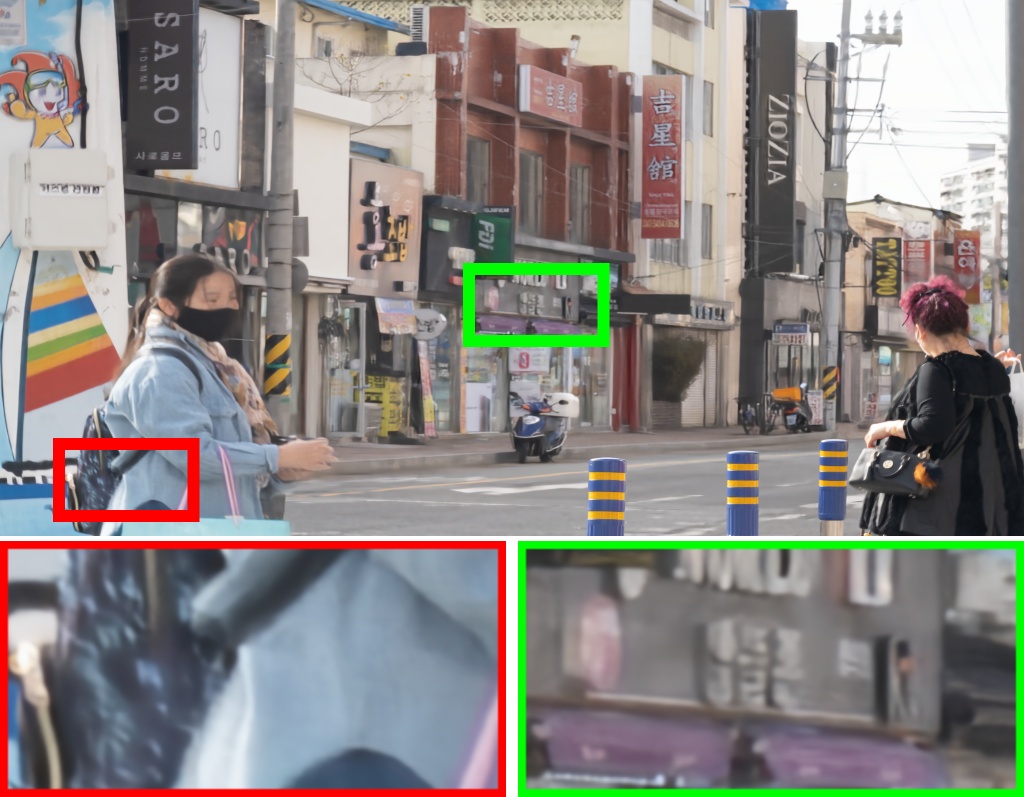}&
								\includegraphics[width=0.155\textwidth]{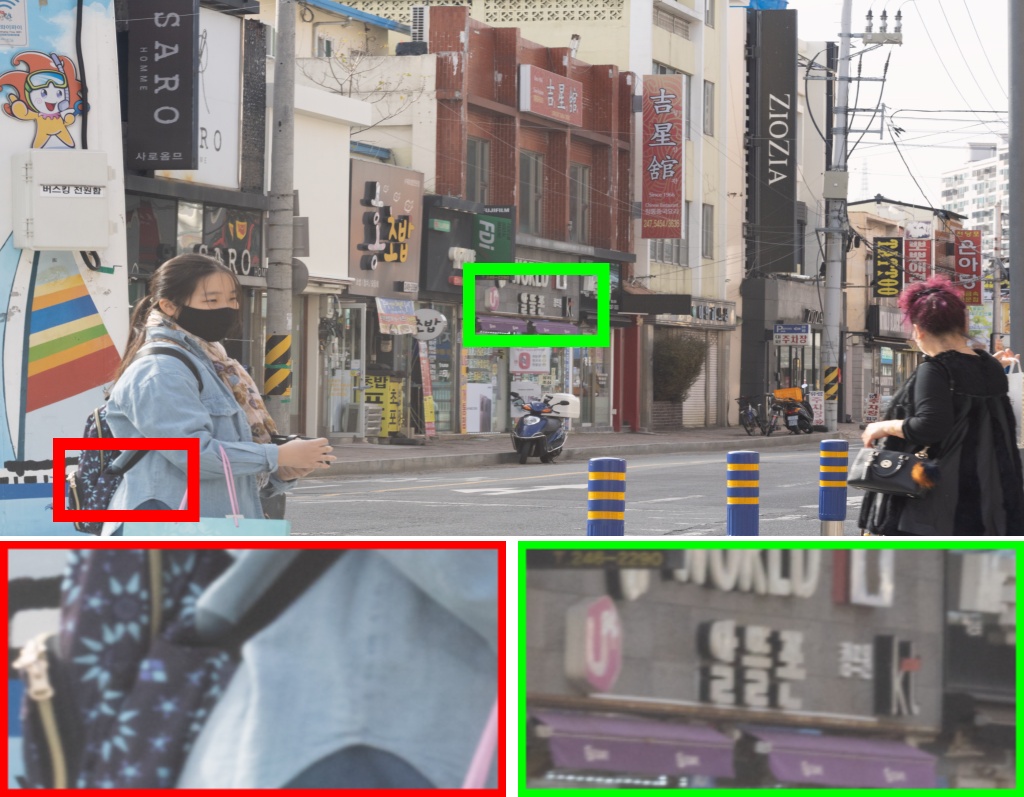}
								\\
								\includegraphics[width=0.155\textwidth]{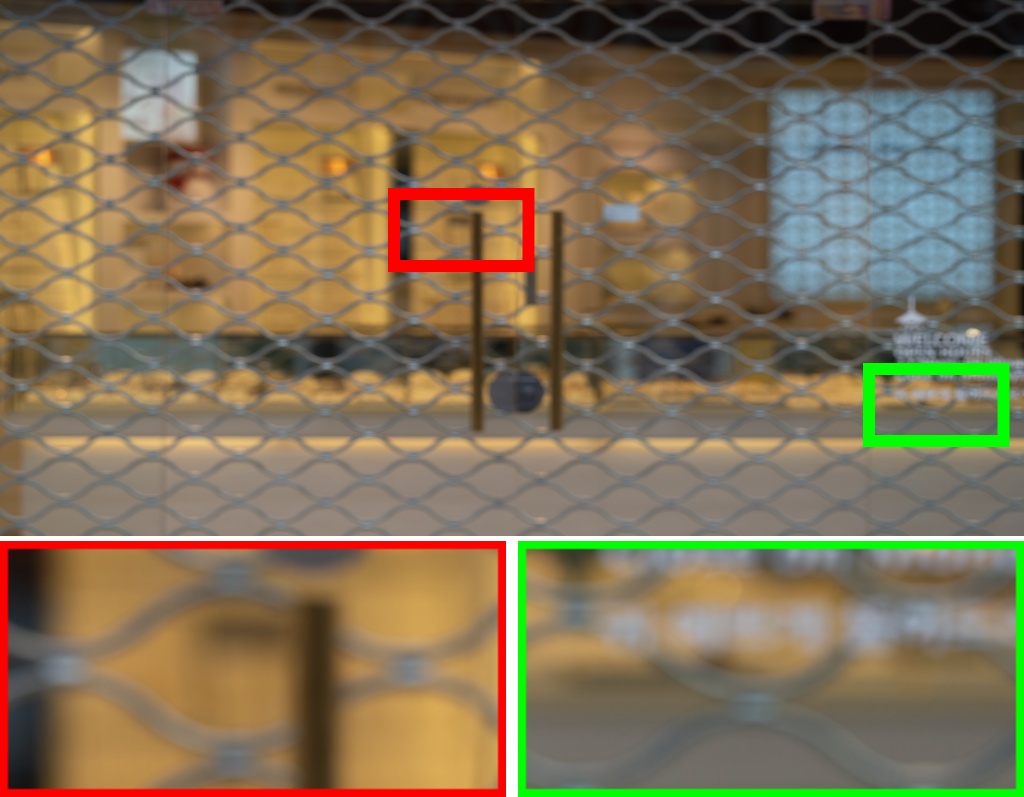}&
								\includegraphics[width=0.155\textwidth]{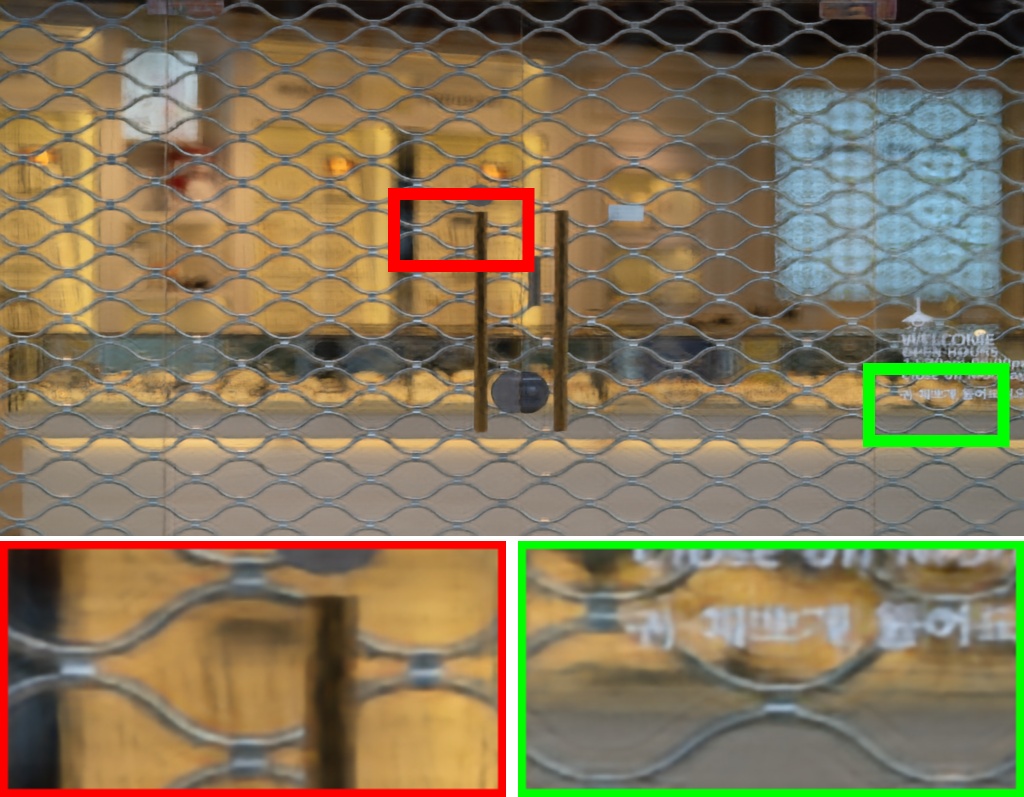}&
								\includegraphics[width=0.155\textwidth]{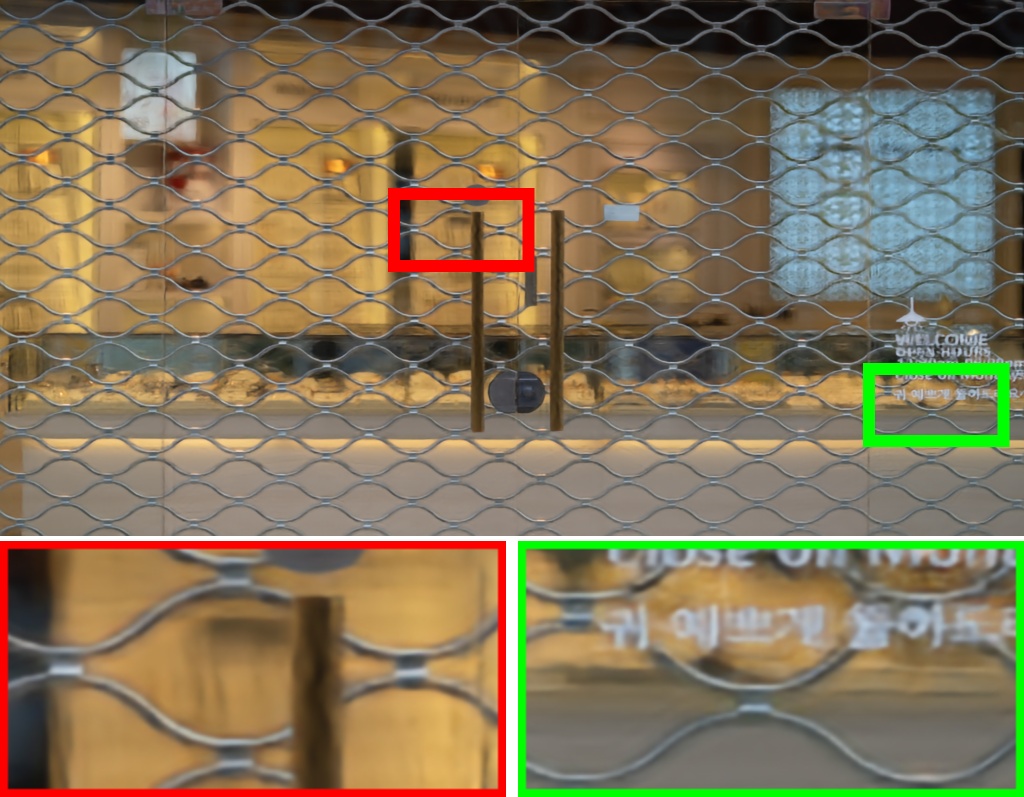}&
								\includegraphics[width=0.155\textwidth]{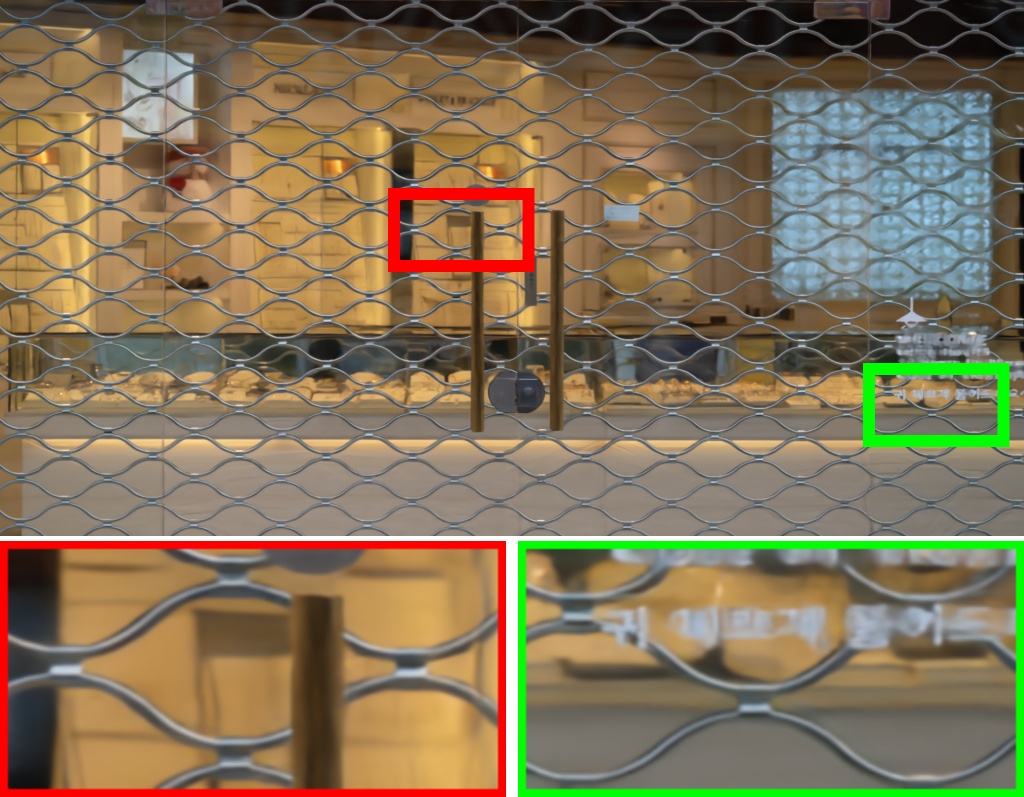}&
								\includegraphics[width=0.155\textwidth]{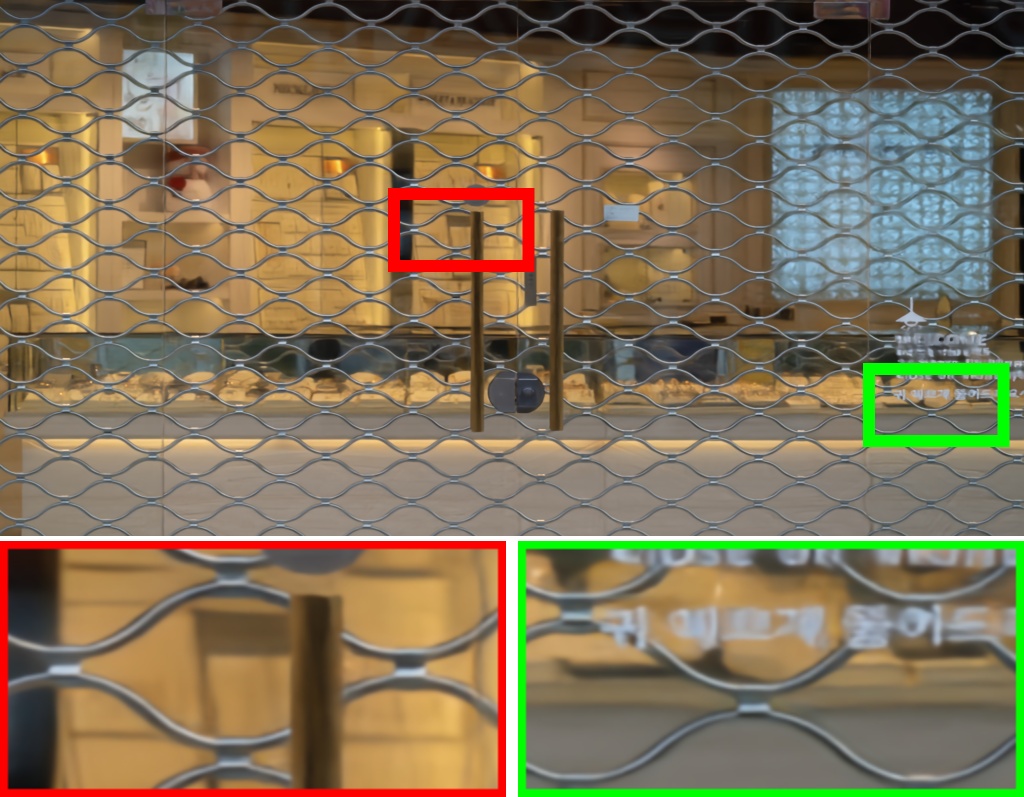}&
								\includegraphics[width=0.155\textwidth]{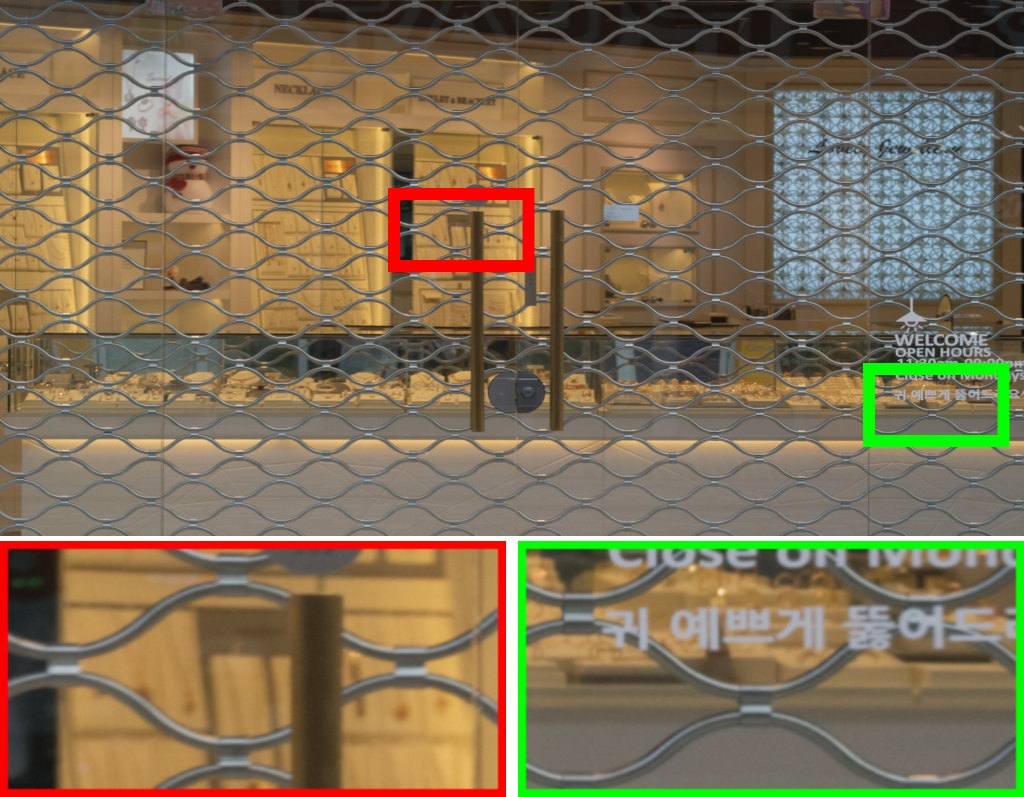}
								\\
								\includegraphics[width=0.155\textwidth]{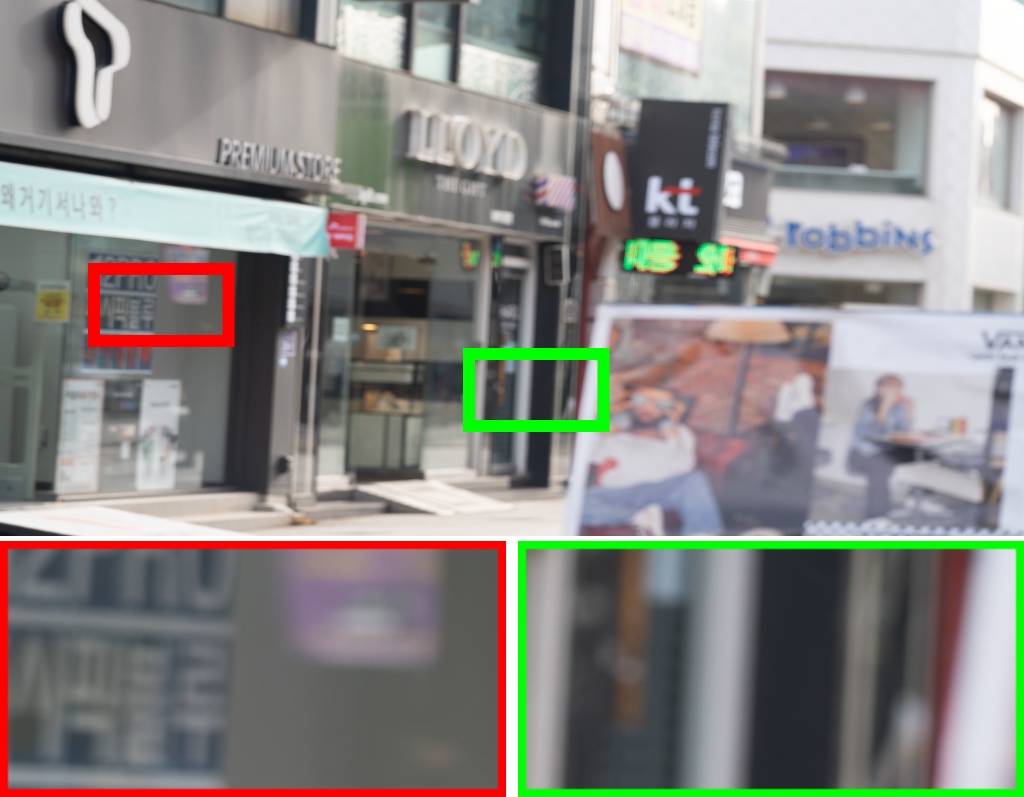}&
								\includegraphics[width=0.155\textwidth]{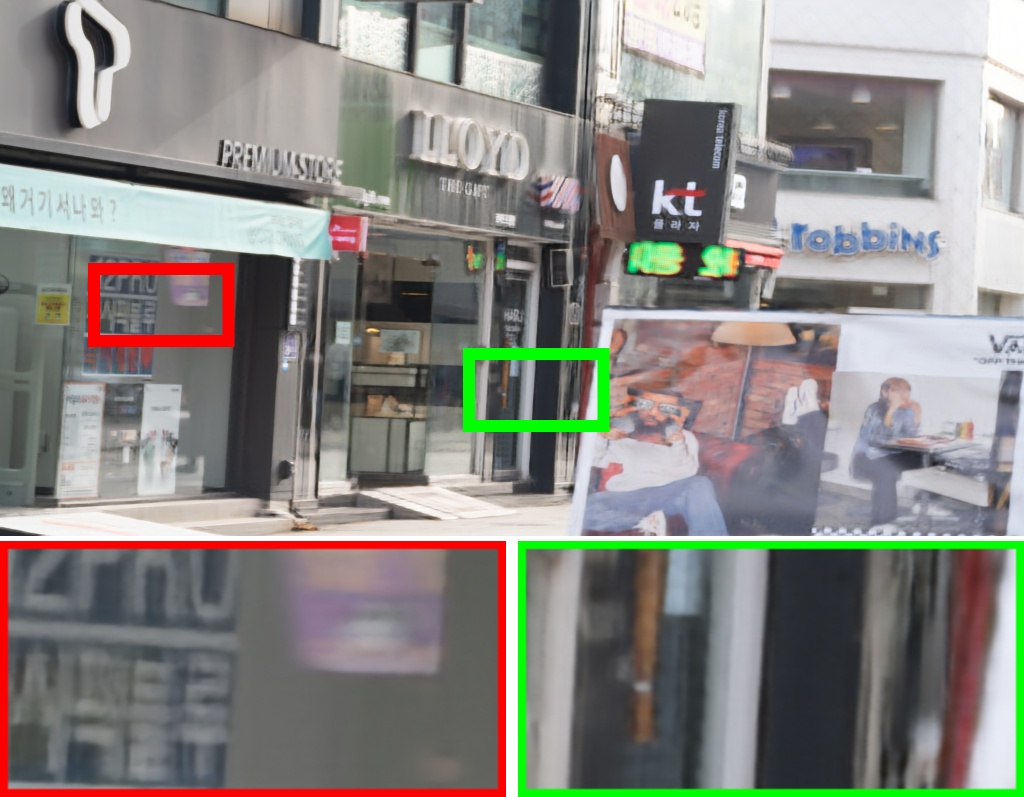}&
								\includegraphics[width=0.155\textwidth]{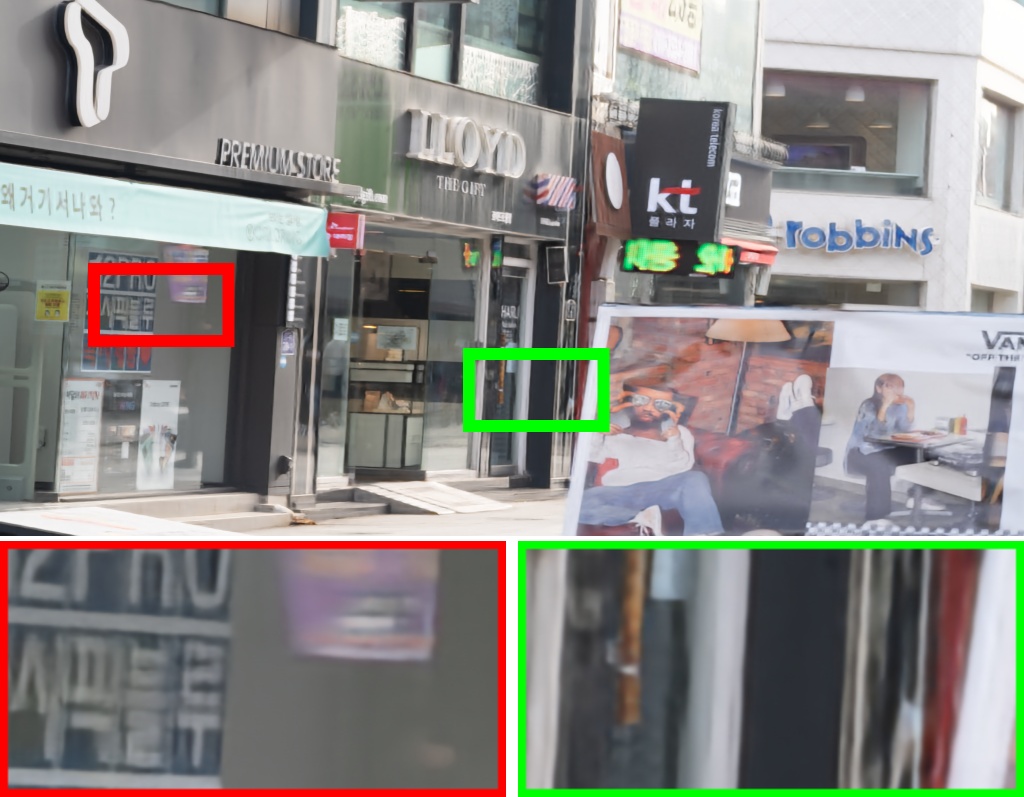}&
								\includegraphics[width=0.155\textwidth]{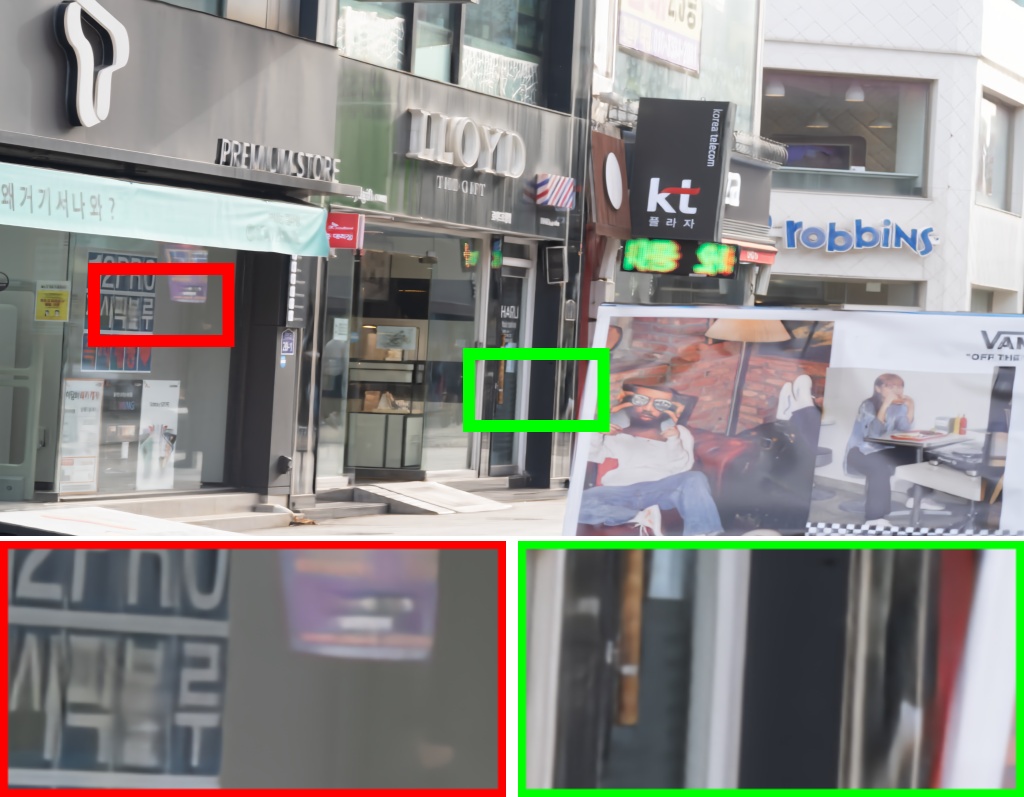}&
								\includegraphics[width=0.155\textwidth]{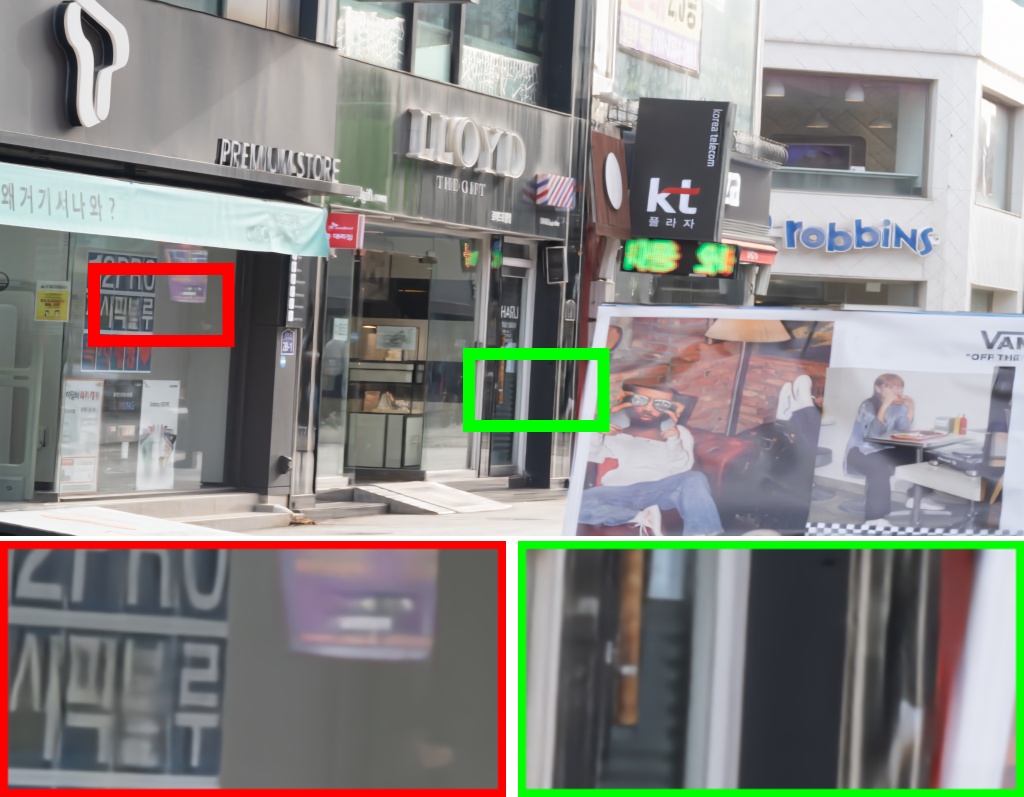}&
								\includegraphics[width=0.155\textwidth]{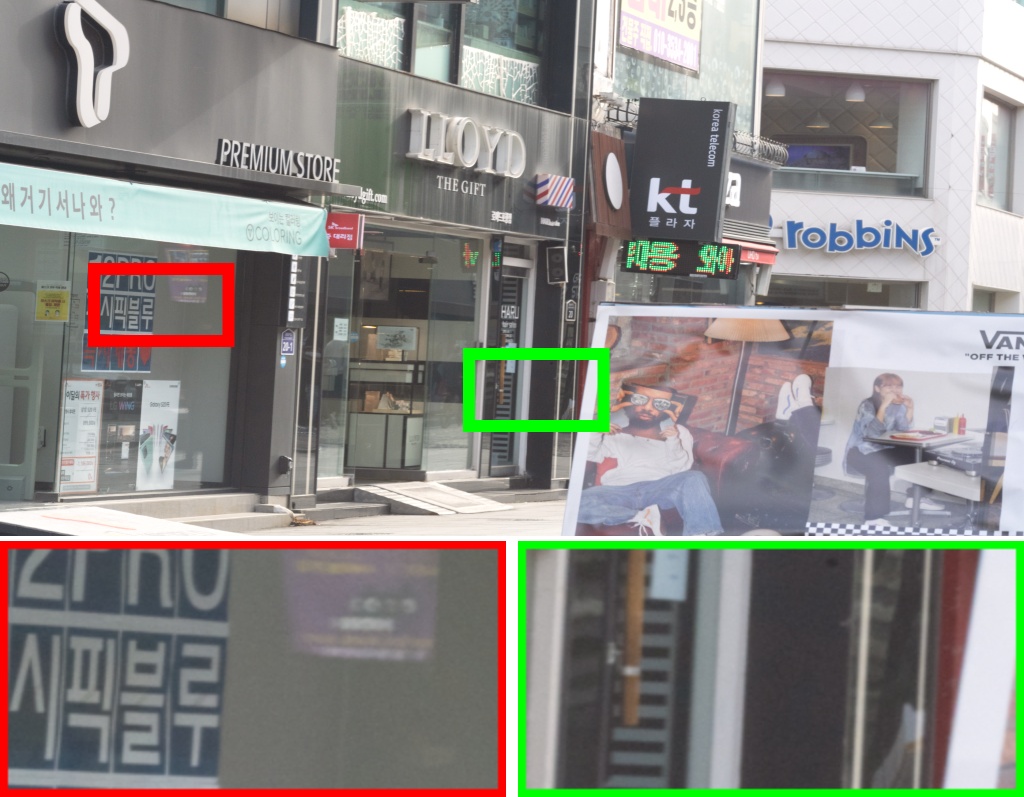}
								\\
								Input&
								Son et al.~\cite{son2021single}&
								IFAN~\cite{lee2021iterative}& 
								% GRL$_{S}$-B\cite{li2023efficient}&
								Restormer\cite{zamir2022restormer}&
								$\text{Restormer}^{\dagger}$&
								GT
								\\
							\end{tabular}
							% \end{adjustbox}
						
					\end{tabular}
					\caption{\small Visual comparison of competing methods on the DPDD dataset and the RealDOF dataset. The first three rows are from the DPDD dataset, and the last three rows are from the RealDOF dataset. Our method demonstrates better visual results in terms of textures and structures.}
					\label{fig:RealDOF_fig}
				\end{figure*}
				\subsection{Evaluation on DED Dataset}
\begin{figure}[!t]
					\small
					\centering
					\setlength{\abovecaptionskip}{5pt} 
					\setlength{\belowcaptionskip}{0pt}
					\setlength{\tabcolsep}{3pt}
					\footnotesize
					\begin{tabular}{cccc}
						% \hspace{-4mm}
						\includegraphics[width=0.11\textwidth]{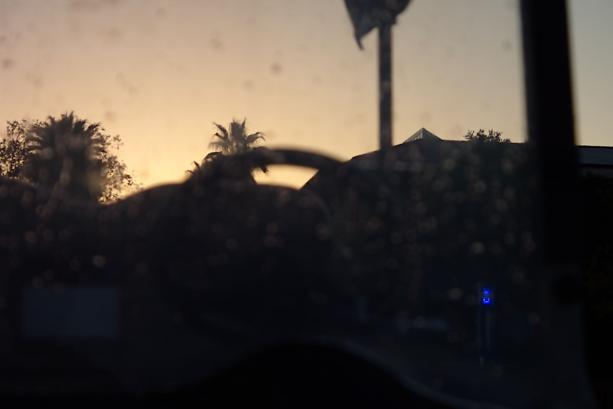}&
						\includegraphics[width=0.11\textwidth]{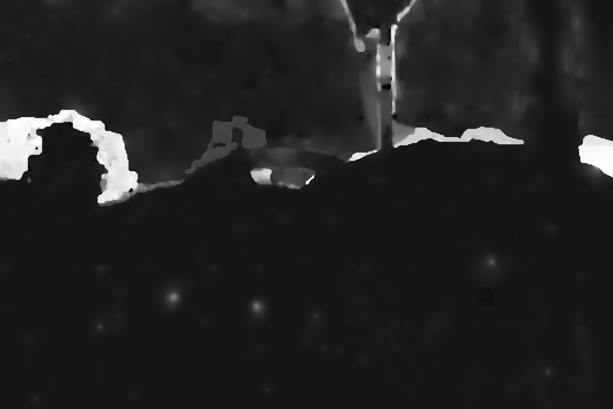}&
						\includegraphics[width=0.11\textwidth]{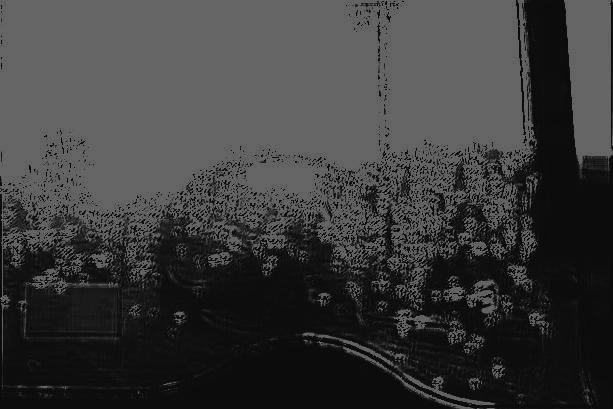}&
						\includegraphics[width=0.11\textwidth]{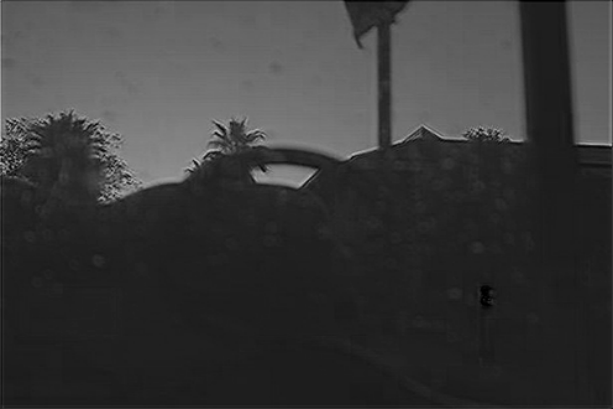}
						\\
						
						\includegraphics[width=0.11\textwidth]{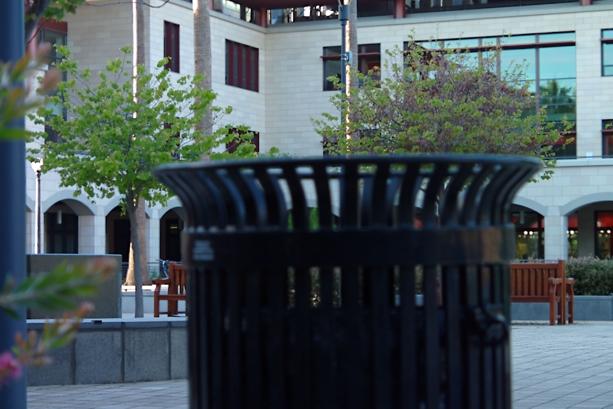}&               \includegraphics[width=0.11\textwidth]{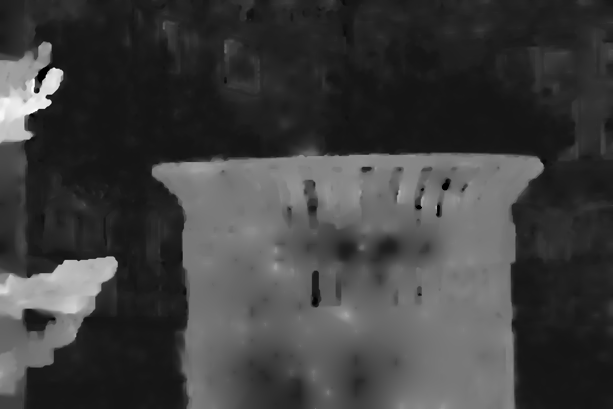}&
						\includegraphics[width=0.11\textwidth]{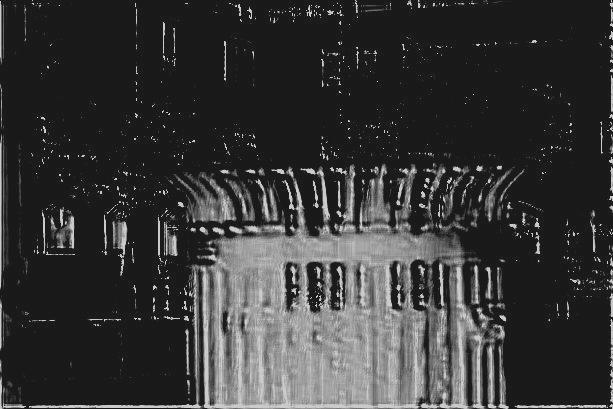}&
						\includegraphics[width=0.11\textwidth]{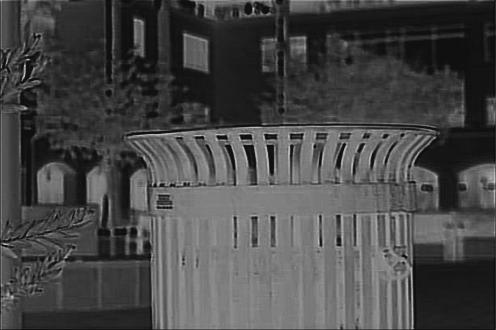}
						\\
						Image&
						Defocus map& 
						DID-ANet\cite{ma2021defocus}&
						Ours
						\\
					\end{tabular}
					\caption{\small Visual comparison of defocus blur maps. 
						Since defocus blur map cannot be truly captured using a camera, it in this case is estimated from light field data captured by a Lytro camera. 
						The defocus blur map estimated by Ours better reflects spatially variant blur amounts than that predicted by DID-ANet \cite{ma2021defocus}. 
						Note that our estimated defocus blur
						maps are with $H \times W \times M $ dimension with $M=35$, and we reduced
						their dimensionality to $H \times W \times 1$ by adopting principal component analysis.
					}
					\label{fig:ded}
				\end{figure}

				The DED dataset~\cite{ma2021defocus} provides training triplets, while the testing set only releases blurred images without their corresponding ground-truth, making it not possible to directly obtain the evaluation results in \cite{ma2021defocus}.  
				Additionally, the pre-trained model of DID-ANet~\cite{ma2021defocus} released by the authors is corrupted\footnote{\url{https://github.com/xytmhy/DID-ANet-Defocus-Deblurring}}.
				Therefore, for testing on the DED dataset~\cite{ma2021defocus}, we randonly re-split the training set of the DED dataset into training and testing sets in a ratio 8:2, and DID-ANet~\cite{ma2021defocus} is trained by adopting their default training settings.  
				Moreover, we trained Restormer and our model for comparison. 
				The re-splitted dataset has also been released by us\footnote{\url{https://github.com/ssscrystal/Reblurring-guided-JDRL}}.
				The results are presented in Table \ref{tab:DED}. 
				Since DED dataset is captured by a Lytro camera, the training triplets are well aligned, it is reasonable that our method only achieves minor improvements than Restormer. 
				The performance gain is mainly from the design of our baseline deblurring model. 
				We also visualized our defocus blur maps in Fig. \ref{fig:ded}. 
				We note that ground-truth defocus maps in DED dataset are also not truly captured, and instead it is estimated by from the light field images by Lytro camera.  
				Since our defocus blur maps are with $M=35$ dimension, we reduced their dimensionality from 35 to 1 by adopting principal component analysis.
				We can see that defocus map by our method can better reflect the spatially variant blur amount than that by DID-ANet.

				\begin{table}[t]%
					% \footnotesize%
					% \arrayrulewidth0.5pt
					\centering
					\caption{\small{Quantitative comparison on DED dataset \cite{ma2021defocus}. 
							We note that the images from DED are generated from light field data captured by a Lytro camera, and are well aligned.}}
					\label{tab:DED}
					\footnotesize
					% \resizebox{\textwidth}{!}{
						\setlength{\tabcolsep}{4pt}
						\begin{tabular}{c|ccccc}
							\toprule
							Method  & PSNR$\uparrow$  & SSIM$\uparrow$ &\ LPIPS$\downarrow$  &FID$\downarrow$&DISTS$\downarrow$\\
							\midrule
							DID-ANet~\cite{ma2021defocus} &  29.59 & 0.860 & 0.186 & 72.97 & 0.193\\
							Restormer\cite{zamir2022restormer} & 31.29 & 0.890 & 0.121 &58.74 & 0.141\\
							\midrule
							$\text{Restormer}^{\dagger}$ &  \textbf{31.37} & \textbf{0.894} & \textbf{0.112} &\textbf{57.26} &\textbf{0.132} \\	
							\bottomrule
							\hline
						\end{tabular}
					\end{table}

					\begin{figure*}[!t]
						\small
						\centering
						\setlength{\abovecaptionskip}{5pt} 
						\setlength{\belowcaptionskip}{0pt}
						\setlength{\tabcolsep}{3pt}
						\footnotesize
						\begin{tabular}{cccc}
							\includegraphics[width=0.22\textwidth]{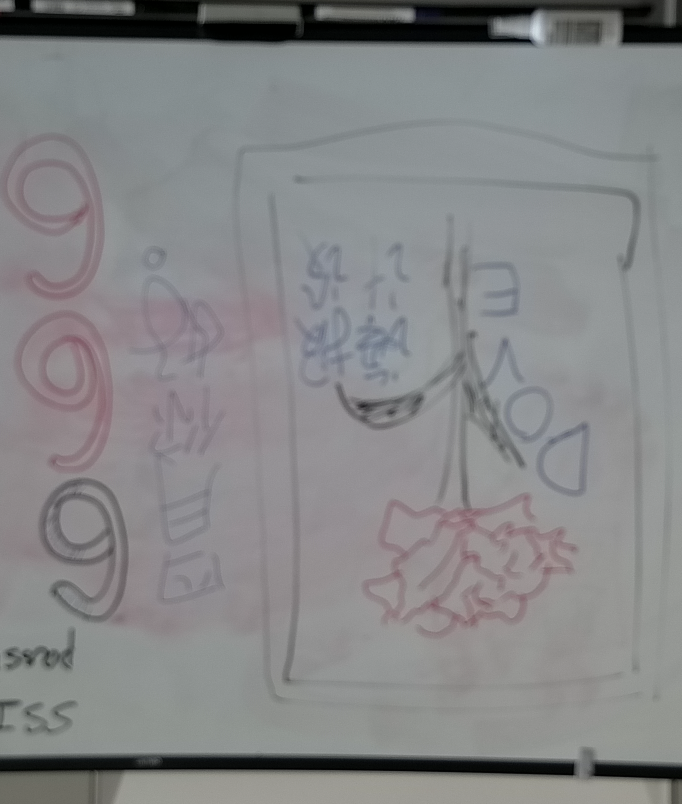}&               \includegraphics[width=0.22\textwidth]{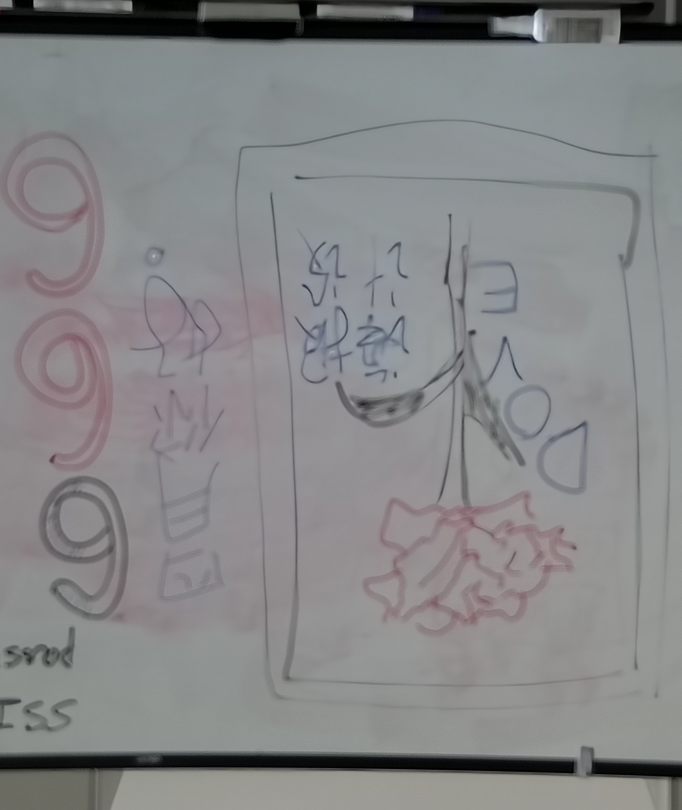}&
							\includegraphics[width=0.22\textwidth]{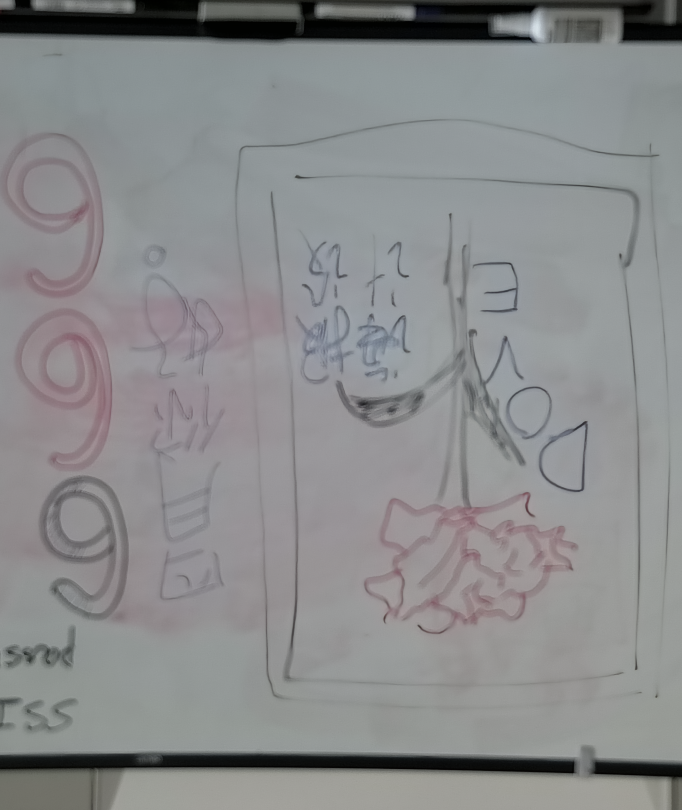}&
							\includegraphics[width=0.22\textwidth]{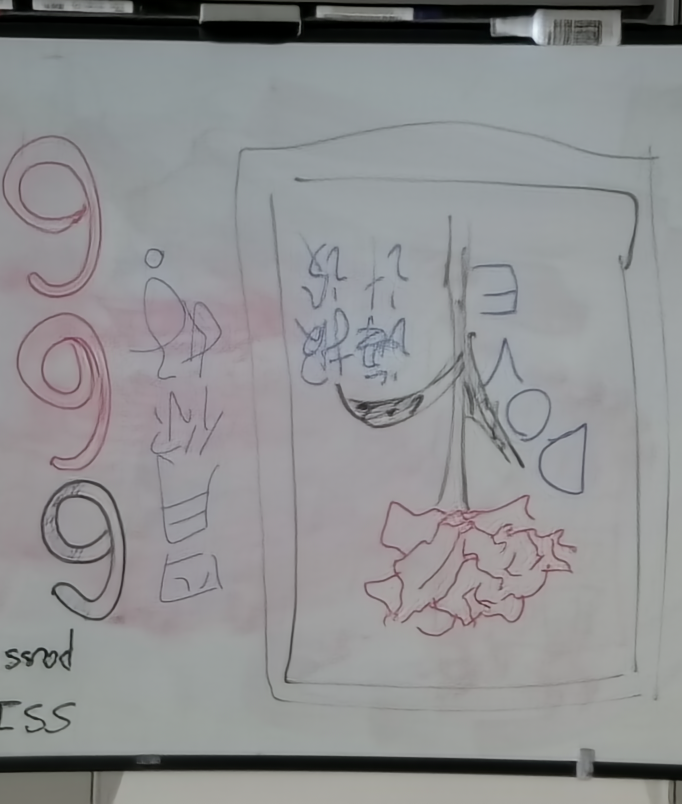}
							\\
							Blurry image&
							Restormer from Table \ref{tab:DED}& 
							Restormer from Table \ref{tab:DPDD_tab}&
							Ours
							\\
						\end{tabular}
						\caption{An image with mild defocus blur from our SDD dataset is handled by two models trained on DPDD and DED datasets respectively, \emph{i.e}., Restormer from Table \ref{tab:DPDD_tab} and Restormer from Table \ref{tab:DED}.  The limited generalization ability to new sensor emphasizes the necessity of an effective learning framework that supports misaligned training pairs.
						}
						\label{fig:domain}
					\end{figure*}
					\subsection{Generalization Evaluation}
					\noindent{\textbf{Evaluation of Pre-trained models on Different Dataset.}}
					This section begins with an evaluation of the performance of models pre-trained on the DPDD~\cite{abuolaim2020defocus} and DED datasets\cite{ma2021defocus}, \emph{i.e}., Restormer from Table \ref{tab:DPDD_tab} and Restormer from Table \ref{tab:DED}, when applied to images from our SDD dataset. 
					Although defocus blur in the input blurry image is not severe as depicted in Figure \ref{fig:domain}, both Restormer models trained on DPDD and DED datasets are limited in removing defocus blur, due to the domain gaps between different sensors for capturing images. 
					The limited generalization ability of trained models to new sensors emphasizes the necessity for a rapid data collection and training framework to enhance model adaptability and performance across diverse sensor types.
					
					\noindent{\textbf{Evaluation of Generalization Ability.}}
					Furthermore, we follow existing works to adopt RealDOF dataset for evaluating generalization ability. We used the trained model from the DPDD dataset~\cite{abuolaim2020defocus} for evaluation. As shown in Fig.~\ref{intro}, the RealDOF dataset also exhibits a certain degree of misalignment. Therefore, the strategy we designed for misaligned datasets also demonstrates effectiveness on the RealDOF dataset. Our evaluation results were calculated between the input images and the ground-truth images after warping, as presented in Table~\ref{tab:realDOF_tab}.
					Although trained on the DPDD dataset, Our learning framework shows robust generalization to other datasets.
					The experimental results also indicate that the domain gap between RealDOF and DPDD~\cite{abuolaim2020defocus} is smaller compared to the gap between DPDD~\cite{abuolaim2020defocus} and SDD.
					Through our framework, training with loosely aligned data pairs not only reduces data collection costs but also achieves better visual performance, providing an efficient and low-cost deblurring strategy for rapid adaptation to new sensors.

					{
						\noindent{\textbf{Cross-dataset Evaluation.}}
						Moreover, we perform cross-dataset evaluation. Within our reblurring-based learning framework, we utilize the UNet architecture to train models on three datasets, \emph{i.e.,} SDD, DPDD \cite{abuolaim2020defocus}, and DED \cite{ma2021defocus}. As shown in Table~\ref{tab:realDOF_dataset}, the model trained on DPDD \cite{abuolaim2020defocus} achieves the best performance, followed by our SDD-trained model, while the DED-trained model exhibited the lowest performance. 
						We attribute the observed experimental results primarily to the varying scales and data distribution of the training datasets. Specifically, DPDD \cite{abuolaim2020defocus} comprises 7,000 training pairs, DED \cite{ma2021defocus} contains 1,112 pairs, and our SDD includes 4,830 pairs.
						We note that the primary contribution of our work lies in providing an approach for effective deployment on target devices by learning a device-specific image defocus deblurring model, while relaxing the strict requirement for perfect alignment in training pairs.  
						Thus, our SDD serves as a testbed for addressing misalignment issues in training pairs and validating the effectiveness of the deblurring models and training strategies, without requiring generalization across different cameras. Consequently, it is acceptable that the model trained on our SDD (captured using a HUAWEI camera) does not achieve the best metrics on RealDOF (captured using a Google Pixel camera).}

					\begin{table}[t]%
						% \footnotesize%
						% \arrayrulewidth0.5pt
						\centering
						\caption{\small{Quantitative comparison on RealDOF dataset \cite{lee2021iterative}.}}
						\label{tab:realDOF_tab}
						\footnotesize
						% \resizebox{\textwidth}{!}{
							\setlength{\tabcolsep}{4pt}
							\begin{tabular}{c|ccccc}
								\toprule
								% 		& \multicolumn{4}{|c}{Deformed GT} \\
								% 		\midrule
								Method  &  \ PSNR$\uparrow$& \ SSIM$\uparrow$&\ LPIPS$\downarrow$ &FID$\downarrow$&DISTS$\downarrow$\\
								% \midrule
								% JNB~\cite{Shi_2015_CVPR} &  21.02 & 0.651  & 0.590 \\
								% EBDB~\cite{karaali2017edge} & 21.37 & 0.659  & 0.579 \\
								% DMENet~\cite{Lee_2019_CVPR} &  21.38 & 0.658  & 0.581 \\
								\midrule
								DPDNet$_{S}$~\cite{abuolaim2020defocus} & 22.66 & 0.702  & 0.397 &73.07 &0.213\\	
								Son et al.\cite{son2021single} & 24.35 & 0.756  & 0.308 & 64.57 & 0.187\\
								IFAN \cite{lee2021iterative}& 25.39 & 0.794  & 0.264 &43.39 &0.162\\
								% 			\midrule
								% UNet & 23.51 & 0.729 & 0.0456 & 0.368 \\
								MPRNet~\cite{zhang2019deep} & 24.66 & 0.771  & 0.313 &52.96 &0.186\\
								Restormer\cite{zamir2022restormer} & 25.80 & 0.815 & 0.249 &40.40 &0.153 \\
								Loformer\cite{mao2024loformer} & 24.72 & 0.780  & 0.306 &47.99 &0.164\\
								% GRL$_{S}$-B\cite{li2023efficient}&25.06&0.797&0.261\\
								
								\midrule
								$\text{Restormer}^{\dagger}$  & \textbf{25.98} &\textbf{0.819} & \textbf{0.220} &\textbf{39.69} &\textbf{0.151} \\		
								\bottomrule
								\hline
							\end{tabular}
						\end{table}
						
						{
							\begin{table}[t]%
								% \footnotesize%
								% \arrayrulewidth0.5pt
								\centering
								\caption{{\small{Evaluation on RealDOF \cite{lee2021iterative} by training UNet models on different datasets.}}}
								\label{tab:realDOF_dataset}
								\footnotesize
								% \resizebox{\textwidth}{!}{
									\setlength{\tabcolsep}{4pt}
									\begin{tabular}{c|ccccc }
										\toprule
										% 		& \multicolumn{4}{|c}{Deformed GT} \\
										% 		\midrule
										Training Data  &  \ PSNR$\uparrow$ \  & \ SSIM$\uparrow$\   &\ LPIPS$\downarrow$\ &FID$\downarrow$\ &DISTS$\downarrow$\  \\
										% \hline
										\midrule
										DED\cite{ma2021defocus} & 21.81 & 0.672 & 0.414 &79.20 & 0.221\\
										DPDD\cite{abuolaim2020defocus}  & 22.28 & 0.687 & 0.401 &  74.27 & 0.216 \\		
										SDD  & 22.03 & 0.680  & 0.406 & 75.41 & 0.219 \\
										\bottomrule
										\hline
									\end{tabular}
								\end{table}

								\subsection{Ablation Study}
								To systematically validate the effectiveness of our proposed deblurring framework, we conduct comprehensive ablation studies from two perspectives: (1) the fusion model and (2) the reblurring-guided learning components. All experiments are performed on the SDD dataset using the UNet architecture as the backbone, with quantitative comparisons against variants that systematically remove key modules. These studies aim to isolate the contributions of each component and verify the necessity of our design choices.
								\subsubsection{Fusion Model Effectiveness Analysis}
								To validate the superiority of our deformable cross-attention fusion model, we provide experimental comparison against these methods, \emph{i.e.,} DConv \cite{dai2017deformable}, LDConv \cite{zhang2024ldconv}, and DAT \cite{xia2022vision}.  
								\begin{table}[t]%
									% \footnotesize%
									% \arrayrulewidth0.5pt
									\centering
									\caption{{\small{Comparison of different deformable modules for fusing defocus blur map and blurry image. }}}
									\label{tab:deformable}
									\footnotesize
									% \resizebox{\textwidth}{!}{
										\setlength{\tabcolsep}{5pt}
										\begin{tabular}{c|ccccc}
											\toprule
											% 		& \multicolumn{4}{|c}{Deformed GT} \\
											% 		\midrule
											Method  &  \ PSNR$\uparrow$ \  & \ SSIM$\uparrow$\   &\ LPIPS$\downarrow$\ &FID$\downarrow$\ &DISTS$\downarrow$\ \\
											% \hline
											\midrule
											DConv & 26.01 & 0.787 & 0.303 &57.68 & 0.183\\
											LDConv & 25.89 & 0.774  & 0.305 & 58.14 & 0.184\\
											Ours  & 26.20 & 0.796 & 0.298 &  55.49 & 0.180\\		
											\bottomrule
											\hline
										\end{tabular}
									\end{table}
									
									\begin{itemize}
										\item \textbf{DConv} learns per-position offsets within a fixed $k \times k$ grid, adapting locally to geometric variations in single-modality features. While we also employ learned offsets, we apply them in the Transformer attention domain, using query-driven reference points rather than a fixed convolution grid. This allows for more flexible, spatially-global alignment during fusion of blurred images and defocus maps.
										
										\item \textbf{LDConv} extends deformable convolution by enabling arbitrary sampling patterns and linear scaling of filter parameters, offering parameter-efficient but flexible sampling. Unlike LDConv’s convolutional sampling and interpolation, our method leverages Transformer cross-attention to fuse features between modalities using deformable sampling guided by cross-modality queries—providing a more expressive and context-aware fusion mechanism.
										
										\item \textbf{DAT} learns shared offset groups to shift key/value positions from a uniform reference grid, leading to data-dependent, sparse attention. We similarly employ query-conditioned deformable attention, but specifically tailor it to spatial alignment across two modalities (blurry image \& defocus map). Our reference points are explicitly optimized to correct misalignment in defocus estimation, rather than general semantic aggregation.
									\end{itemize}
									As shown in Table~\ref{tab:deformable}, DConv and LDConv exhibit inferior performance compared to Ours, due to their limited capability in capturing long-range correspondence. 
									Overall, deformable attention is likely to outperform deformable convolution, especially in scenarios with significant spatial misalignment, owing to its broader perception field.
									In Table~\ref{tab:deformable}, the results of DAT are omitted, because of its excessive GPU memory consumption, such as running out of memory on an 80G A100 GPU. The primary computational burden stems from the integration of local and deformable attention mechanisms, where local attention may not benefit resolving spatial misalignment when fusing defocus blur maps with blurry images. 
									
									\subsubsection{Framework Components Ablation}
								}
								% To conduct an efficient ablation study, we employ the UNet as our deblurring model to primarily validate the effectiveness of our reblurring-guided learning framework. 
								To systematically evaluate the contributions of our reblurring-guided learning framework, we design ablation variants based on the UNet deblurring model, all trained on the SDD dataset:
								The effectiveness of the deblurring model involving pseudo defocus blur map for supervision has already been demonstrated in Table \ref{tab:SDD more metrics}, where Ours with Eq. \eqref{eq:jdrl} outperforms Ours with Eq. \eqref{eq:jdrl-simple}. 
								Our ablation studies are performed on the SDD dataset, and six variants are designed to analyze their contributions. 
								Variant \#1 represents a vanilla UNet trained using $\mathcal{L}_1$ loss.
								Variants \#2 and \#3 correspond to Ours with Eq. \eqref{eq:jdrl-simple} by excluding the bi-directional optical flow deforming process and calibration mask, respectively. 
								Variants \#4, \#5, and \#6 aim to demonstrate the effectiveness of components within our reblurring module. Specifically, they variants of  Ours with Eq. \eqref{eq:jdrl-simple} by removing the reblurring module, isotropic kernels module $\mathcal{R}_{kpn}$ and weight prediction module $\mathcal{R}_{wpn}$ from, respectively.
								
								% \begin{table}[t]%\footnotesize%\arrayrulewidth0.5pt
									%     \centering
									%     \setlength{\abovecaptionskip}{0pt} 
									%     \setlength{\belowcaptionskip}{0pt}
									%     \caption{\small{Ablation study on SDD dataset.}}
									%     \label{tab:ablation}
									%     \footnotesize
									%     \begin{tabular}{l|c}
										%         \toprule
										%         Variant &  \ PSNR/SSIM/LPIPS\\
										%         \midrule
										%         \#1 w/ $\mathcal{L}_1$ loss   & 24.62/0.758/0.344 \\
										%         \#2 w/o Cycle Deformation      &  24.80/0.758/0.347 \\
										%         \#3 w/o Calibration Mask      & 25.78/0.780/0.309 \\
										%         \midrule
										%         \#4 w/o reblurring module      & 25.58/0.777/0.311  \\
										%         \#5 w/o $\mathcal{R}_{kpn}$     & 25.68/0.778/0.311 \\
										%         \#6 w/o $\mathcal{R}_{wpn}$      & 25.74/0.778/0.332 \\
										%         % \#7 w/o $\mathcal{P}$      & 25.82/0.783/0.0320/0.305 \\
										%         \midrule
										%         \ \ \ \ \ Ours (Eq. \eqref{eq:jdrl-simple})      &25.94/0.791/0.289\\		
										%         \bottomrule
										%         \hline
										%     \end{tabular}
									% \end{table}
								\begin{table}[ht]%\footnotesize%\arrayrulewidth0.5pt
									\centering
									\setlength{\abovecaptionskip}{0pt} 
									\setlength{\belowcaptionskip}{0pt}
									\caption{\small{Ablation study on SDD dataset.}}
									\label{tab:ablation}
									\footnotesize
									\begin{tabular}{l|c}
										\toprule
										Variant &  \ PSNR/SSIM/LPIPS\\
										\midrule
										\#1 w/ $\mathcal{L}_1$ loss   & 24.62/0.758/0.344 \\
										\#2 w/ Warp Operation           & 25.51/0.776/0.322\\
										\#3 w/o Cycle Deformation      &  24.80/0.758/0.347 \\
										\#4 w/o Calibration Mask      & 25.78/0.780/0.309 \\
										\midrule
										\#5 w/o reblurring module      & 25.58/0.777/0.311  \\
										\#6 w/o $\mathcal{R}_{kpn}$     & 25.68/0.778/0.311 \\
										\#7 w/o $\mathcal{R}_{wpn}$      & 25.74/0.778/0.332 \\
										\midrule
										\quad Ours (Eq. \eqref{eq:jdrl-simple})      &25.82/0.783/0.305\\
										\quad Ours (Eq. \eqref{eq:jdrl})     &26.20/0.796/0.298\\
										\bottomrule
										\hline
									\end{tabular}
								\end{table}
								The quantitative results are presented in Table~\ref{tab:ablation}. Observations from experiments \#1, \#2 and \#3 demonstrate that the absence of the deformation process significantly diminishes network performance. Moreover, the inclusion of the calibration mask appears to moderately enhance performance, as depicted in experiment \#4. 
								Experiment \#5 clearly indicates that the lack of reblurring module substantially impacts overall network performance. 
								The results of experiments \#6 and \#7 validate the advantages of isotropic blur kernels prediction module $\mathcal{R}_{kpn}$ and weight prediction module $\mathcal{R}_{wpn}$ within reblurring module for the reblurring process.

								\section{Conclusion}\label{sec5}
								In this paper, we propose a reblurring-guided learning framework, designed specifically to tackle the significant challenge of misalignment in training pairs for single image defocus deblurring. 
								The proposed method is distinctively composed of a deblurring module that integrates prior knowledge through a lightweight prediction model and a bi-directional optical flow-based deformation technique.
								This enables the framework to adeptly accommodate spatial misalignments between training pairs.
								Furthermore, the spatially variant reblurring module plays a pivotal role in reblurring the deblurred output to achieve spatial alignment with the original blurry image, utilizing predicted isotropic blur kernels and generating weighting maps for this purpose.
								Moreover, we also establish a new single image defocus deblurring dataset to evaluate our method and benefit future research. Extensive results on SDD, DPDD, DED and RealDOF datasets validate the effectiveness of our method in comparison to state-of-the-art methods.

								
								
								
								% \end{appendices}
							
							% %%===========================================================================================%%
							% %% If you are submitting to one of the Nature Portfolio journals, using the eJP submission   %%
							% %% system, please include the references within the manuscript file itself. You may do this  %%
							% %% by copying the reference list from your .bbl file, paste it into the main manuscript .tex %%
							% %% file, and delete the associated \verb+\bibliography+ commands.                            %%
							% %%===========================================================================================%%
							\bibliographystyle{unsrt}
							% {\footnotesize
								\bibliography{sn-bibliography}
								% }
							% \bibliographystyle{apacite}
							% %% if required, the content of .bbl file can be included here once bbl is generated
							% \input sn-article.bbl

						\end{document}